\documentclass[10pt]{article}
\usepackage{amsmath, amssymb,amsthm, fullpage, color, relsize, enumerate, enumitem, hyperref, mathtools}
\usepackage{latexsym}
\usepackage{xparse}
\usepackage{array}
\usepackage{xstring}
\usepackage{indentfirst}
\usepackage{authblk}
\usepackage{bm,dsfont} 
\usepackage{graphicx,adjustbox}
\usepackage{chngcntr}
\usepackage{import}
\usepackage{comment}
\usepackage{amsfonts}
\usepackage{lipsum}
\usepackage[T2A, T1]{fontenc}
\usepackage[utf8]{inputenc}
\usepackage{mathtools}  
\usepackage{multirow,multicol}
\usepackage[noadjust]{cite}
\usepackage{subfig}
\usepackage{caption}
\usepackage[russian, english]{babel}
\numberwithin{equation}{section}
\usepackage{authblk}
\usepackage[toc,page]{appendix}
\usepackage{pgfplots}
\pgfplotsset{compat=newest} 
\usepackage{tikzscale}
\pgfplotsset{plot coordinates/math parser=false}

\DeclareMathOperator{\tr}{tr}

\title{Sequential change-point detection in high-dimensional Gaussian graphical models\footnote{GM is partially supported by NSF grants DMS-1545277, DMS-1632730 and NIH grant 1R01-GM1140201A1, and YA by NSF DMS-1513040.} \blfootnote{Email addresses: \textit{ \href{mailto:hkeshava@umn.edu}{hkeshava@umn.edu}, \href{mailto:gmichail@ufl.edu}{gmichail@ufl.edu}, \href{mailto:yvesa@umich.edu}{yvesa@umich.edu}}}}

\date{\vspace{-5ex}}
\author[1]{Hossein Keshavarz}
\author[2]{George Michailidis}
\author[3]{Yves Atchad\'{e}}
\affil[1]{Institute for Mathematics and its Applications, University of Minnesota}
\affil[2]{Department of Statistics \& UF Informatics Institute, University of Florida}
\affil[3]{Department of Statistics, University of Michigan}

\theoremstyle{definition}
\newtheorem{defn}{Definition}[section]

\newtheorem{prop}{Proposition}[section]

\newtheorem{lem}{Lemma}[section]
\newtheorem{thm}{Theorem}[section]
\newtheorem{cor}{Corollary}[section]
\newtheorem{rem}{Remark}[section]
\newtheorem{assu}{Assumption}[section]

\newtheorem{clawithinpf}{Claim}
\makeatletter
\@addtoreset{clawithinpf}{proof}
\makeatother

\long\def\comment#1{}

\makeatletter
\newcommand{\thickhline}{%
	\noalign {\ifnum 0=`}\fi \hrule height 1pt
	\futurelet \reserved@a \@xhline
}
\newcolumntype{"}{@{\hskip\tabcolsep\vrule width 1pt\hskip\tabcolsep}}
\makeatother

\makeatletter
\def\amsbb{\use@mathgroup \M@U \symAMSb}
\makeatother

\makeatletter
\g@addto@macro\@floatboxreset\centering
\makeatother

\makeatletter
\newcommand*{\rom}[1]{\expandafter\@slowromancap\romannumeral #1@}
\makeatother

\newcommand{\suchthat}{\mathop{\mathrm{s.t.}}}

\newcommand\blfootnote[1]{%
	\begingroup
	\renewcommand\thefootnote{}\footnote{#1}%
	\addtocounter{footnote}{-1}%
	\endgroup
}

\newcommand{\bb}[1]{\mathbb{#1}} 
\newcommand{\bbM}[1]{\mathds{#1}} 
\newcommand{\cc}[1]{\mathcal{#1}} 
 
\newcommand{\zero}{\boldsymbol{0}}

\newcommand{\sign}{\mathop{\mathrm{sign}}}
\newcommand{\FA}{\mathop{\mathrm{FA}}}
\newcommand{\MD}{\mathop{\mathrm{MD}}}
\newcommand{\BC}{\mathop{\mathrm{BC}}}
\newcommand{\AC}{\mathop{\mathrm{AC}}}
\newcommand{\diag}{\mathop{\mathrm{diag}}}

\newcommand{\supp}{\mathop{\mathrm{supp}}}

\newcommand{\cov}{\mathop{\mathrm{cov}}} 
\newcommand{\corr}{\mathop{\mathrm{corr}}} 
\newcommand{\var}{\mathop{\mathrm{var}}} 
\newcommand{\sd}{\mathop{\mathrm{std}}}

\newcommand{\argmin}{\operatornamewithlimits{arg\min}}

\newcommand{\brac}[1]{\left[#1\right]}
\newcommand{\set}[1]{\left\{#1\right\}}
\newcommand{\abs}[1]{\left\lvert #1 \right\rvert}
\newcommand{\paren}[1]{\left(#1\right)}
\newcommand{\parbra}[1]{\left(#1\right]}
\newcommand{\brapar}[1]{\left[#1\right)}
\newcommand{\Bigparen}[1]{\Big(#1\Big)}
\newcommand{\Bigbrac}[1]{\Big[#1\Big]}

\newcommand{\InnerProd}[2]{\langle #1,#2 \rangle}

\newcommand{\cp}[1]{\overset{#1}{\rightarrow}}    
\newcommand{\eqd}{\overset{d}{=}}  
\newcommand{\RelNum}[2]{\overset{#1}{#2}}

\newcommand{\TimeVarQuant}[3]{
	\IfEqCase{#3}{
		{sub}{{#1}_{#2}}
		{sup}{{#1}^{#2}}
	}
}

\newcommand{\LpNorm}[2]{\left\| #1\right\|_{\ell_{#2}}}

\newcommand{\norm}[2]{\left\|#1\right\|_{#2}}
\newcommand{\OpNorm}[3]{\left\| #1\right\|_{#2\rightarrow#3}}
\newcommand{\RNum}[1]{\uppercase\expandafter{\romannumeral #1\relax}}

\providecommand{\keywords}[1]{\textit{Key words.} #1}

\begin{document}
\maketitle

\begin{abstract}
High dimensional piecewise stationary graphical models represent a versatile class for modelling time varying networks arising in diverse application areas, including biology, economics, and social sciences. There has been recent work in offline detection and estimation of regime changes in the topology of sparse graphical models. However, the online setting remains largely unexplored, despite its high relevance to applications in sensor networks and other engineering monitoring systems, as well as financial markets. To that end, this work introduces a novel scalable online algorithm for detecting an unknown number of abrupt changes in the inverse covariance matrix of sparse Gaussian graphical models with small delay. The proposed algorithm is based upon monitoring the conditional log-likelihood of all nodes in the network and can be extended to a large class of continuous and discrete graphical models. We also investigate asymptotic properties of our procedure under certain mild regularity conditions on the graph size, sparsity level, number of samples, and pre- and post-changes in the topology of the network. Numerical works on both synthetic and real data illustrate the good performance of the proposed methodology both in terms of computational and statistical efficiency across numerous experimental settings.
\end{abstract}


\section{Introduction}\label{Intro}

Recent technological advances in data mining have revolutionized the collection of complex and high-resolution financial, biological and social data \cite{wu2014data}. Characterizing and understanding the relationships amongst a large number of variables poses novel methodological and technical challenges. Probabilistic graphical models capture the conditional dependence structure between variables of interest \cite{wainwright2008graphical} and thus have become a standard tool for the aforementioned task. Further, Gaussian processes provide powerful models in many applications, which when combined with the computational advantages of estimating Gaussian Graphical Models (GGM) has rendered them very popular in empirical work (see e.g., \cite{bessler2003structure, jones2005covariance}).

An undirected probabilistic graphical model comprises of $p$ nodes (variables of interest $\boldsymbol{X}=\brac{X_1,\ldots,X_p}^\top$) $\set{V_1,\ldots,V_p}$, and whose edges represent conditional dependence relationships amongst them; specifically, there is an edge between $V_i$ and $V_j$ if they are conditionally dependent given all other variables (nodes). In GGMs $\boldsymbol{X}$ is a zero mean Gaussian random vector with covariance matrix $\Sigma$ and in the precision matrix $\Omega\coloneqq\Sigma^{-1}$, an edge connects $V_i$ and $V_j$ if and only if $\Omega_{ij} \ne 0$. In other words, the conditional dependency structure and topology of a GGM can be uniquely encoded by $\Omega$. Consequently, a rich literature has been developed around estimating $\Omega$ in time invariant GGMs given $n$ i.i.d zero mean observations $\boldsymbol{X}_1,\ldots, \boldsymbol{X}_n\in\bb{R}^p$ from the network. In case,
where $n>p$ the Maximum Likelihood estimator of $\Omega$ corresponds to the inverse of the empirical covariance matrix of $\Sigma$, while in the case of $n<p$ one needs to impose additional structural assumptions, such as sparsity, on underlying $\Omega$ (see \cite{buhlmann2011statistics} and references therein).

Despite voluminous research on estimating stationary graphical models, in various scenarios the underlying dependency structure dynamically evolves over time. Next, we discuss several real-world problems centring around temporally evolving high-dimensional data with underlying network structure.
\begin{enumerate}[label = (\alph*),leftmargin=*]
\item Temporal fluctuations in functional connectivity (FC), which is referred to as dynamic-FC, has recently received a lot of attention in resting-state blood-oxygen level-dependent functional magnetic resonance imaging (rs-fMRI) \cite{hutchison2013resting,hindriks2016can}. The dynamic-FC is commonly investigated using covariance matrix of FC over consecutive windowed segments.
\item The similarity pattern of streaming measurements in a large sensor network can be subject to abrupt changes due to anomalous behaviour of an unknown subset of nodes. For instance, Kalitsis et al. \cite{kallitsis2016adaptive} studied the online detection of false data injection attacks in wide-area smart grid networks.

\item Stock return time series occasionally exhibit radical changes associated
with stochastic switching between high and low volatility regimes, financial crises or changes in government policy. Despite a rich literature on regime-switching low-dimensional time series models \cite{bai1998estimating, bai2003computation}, there remains an incomplete understanding of this idea for high dimensional financial network data. Note that there are a few recent studies \cite{atchade2017scalable,barnett2016change} on detecting multiple sudden changes in the dependency structure of S\&P 500 stocks returns for the period of $1982-2000$ and $2000-2016$  as a result of the stock market crash of October $1987$ (known as \emph{Black Monday}), beginning of the \emph{great recession} on January $2008$, and the bankruptcy declaration day of \emph{Lehman Brothers Holdings Inc.} on September $15$, $2008$.
\end{enumerate} 

Learning temporally evolving dependency structures across a large number of variables requires specification of the mechanism that drives the underlying dynamics. A simple, analytically tractable and widely applicable mechanism is given by assuming piecewise stationary dynamics subject to unknown break points. This versatile model, which is known as the \emph{change-point} model, has been extensively studied for low dimensional time series during the last few decades; see for example, \cite{basseville1993detection,horvath2014extensions,aue2009break,keshavarz2017optimal} and references therein. However, there is significantly less work towards understanding the algorithmic and theoretical aspects of change-point estimation for high-dimensional time series data. 

Change-point detection algorithms are classified into two groups: \emph{offline} and \emph{sequential (online)}. Given the entire data set beforehand, the objective of offline algorithms is to spot the abrupt changes by scanning through the available data. On the other hand in the sequential framework, detection and collection of new samples run concurrently and the goal is to find sudden changes with the smallest delay after they occur. Given independent observations in the high dimensional paradigm, \cite{li2012two,cai2013two} designed offline two sample-tests for identifying a single change-points in the covariance matrix. In \cite{kolar2010estimating,kolar2012estimating,gibberd2017multiple} maximizing a regularized \emph{pseudo-likelihood} with \emph{fused-lasso} penalty has been proposed for estimating multiple structural breaks in the inverse covariance matrix of high dimensional sparse GGMs. Roy et al. \cite{roy2016change} introduced a two-step algorithm for estimating a single abrupt change in the parameters of high dimensional sparse Markov random fields. Given an estimate of sudden change, the parameters before and after the change-point are separately estimated by maximizing an $\ell_1$ penalized pseudo-likelihood function. A brute force search on a coarse grid is necessary for updating the location of change-point in each step. Solely focusing on GGMs, Atchad\'{e} and Bybee \cite{atchade2017scalable} proposed an approximate majorize-minimize (MM) algorithm for reducing the computational cost of brute force search. Note that the $\ell_1$ penalized loss maximization algorithms in \cite{roy2016change,atchade2017scalable} are capable of estimating a single change-point and extension to the case of multiple jumps requires binary segmentation. Note that despite its relatively low computational cost, the binary segmentation is a greedy procedure that is not guaranteed to maximize the log-likelihood function. Finally, Chen et al. \cite{chen2015graph} proposed a non-parametric sequential algorithm for detecting sudden changes in the similarity graph of dependent random variables. 

Motivated by applications in sensor networks and financial markets, we propose a novel {\em sequential algorithm} for detecting sudden changes in sparse inverse covariance structure of high-dimensional GGMs and investigate its theoretical and numerical properties. To the best of our knowledge, this constitutes the first attempt for addressing the problem of online detection of abrupt changes in high-dimensional and sparse graphical models. The sequential change-point detection in an inverse covariance (precision) matrix inevitably associates with two technical challenges. First unlike the covariance matrix, the elements (or eigenvalues) of the precision matrix can not be easily characterized by the data, particularly in the \textquotedblleft{\emph{large $p$, small $n$}}\textquotedblright{} framework. Another challenge, which is a distinctive feature of high dimensional sequential detection, arises when the main concern is spotting sudden shifts with reasonable delay. In contrast to offline high-dimensional change-point detection, the number of \emph{post-change} observations is considerably smaller than $p$. Therefore unlike penalized likelihood-based abrupt-change estimators in \cite{atchade2017scalable,gibberd2017multiple,kolar2010estimating,kolar2012estimating,roy2016change}, the detection procedure should be decoupled from estimating the post-change dependence structure of GGM.

\paragraph{Outline of online detection strategy.} We conclude this section by presenting a concise and high-level introduction to the proposed sequential detection algorithm. Let $G_t = \paren{V,E_t}$ denote a zero-mean GGM with $p$ vertices and a time-varying precision matrix $\Omega^{\paren{t}}$. For each $t$, we observe a single realization of $G_t$, which is represented by $\boldsymbol{X}_t = \brac{X_{t,1},\ldots,X_{t,p}}^\top$. We conduct a statistical test to determine whether a jump occurred at time $t$, i.e. $\Omega^{\paren{t}}\ne \Omega^{\paren{t+1}}$. As detecting small changes can be challenging for complicated objects such as $\Omega^{\paren{t}}$ (especially for large $p$), our study is hinged on two moderate and prevalent restrictive conditions.
\begin{enumerate}[label = (\alph*),leftmargin=*]
\item (\emph{Sparsity}) For regulating the amount of conditional dependence in $G_t$, $\Omega^{\paren{t}}$ is assumed to be a sparse matrix, for each $t$. Some type of sparsity assumption often holds in the aforementioned applied problems.
\item (\emph{Detection delay}) Many sequential decision rules rely on utilizing post-change features of the process. We adopt a similar approach by allowing to observe $w$ samples ahead before making a decision. Namely, we raise an alarm at $t$ by using two sources of information, $w$ post-change observations ($\boldsymbol{X}_{t+1},\ldots,\boldsymbol{X}_{t+w}$) and pre-change features such as an $\ell_1$-regularized estimate of $\Omega^{\paren{t}}$. In previous work on offline change-point learning in sparse graphical models, the estimated jump is of order $\log p$ distant from the true location of a change-point (e.g., Theorem $8$ of \cite{atchade2017scalable} or Theorem $1$ in \cite{roy2016change}). So roughly speaking, $w$ is an online variant of the corresponding abrupt-change estimation error in offline approach.
\end{enumerate}

Comparing the pre- and post-change conditional log-likelihood of all nodes is at the heart of our proposed algorithm. We particularly show that $-2\log\bb{P}\paren{X_{ts}\mid X_{ts'}:\;s'\ne s; \Omega^{\paren{t}}}+\log\Omega^{\paren{t}}_{ss}-\log\paren{2\pi}$ has a $\chi^2_1$ density. However if a change occurs at $t$ lasted for at least $w$ time points, i.e., $\Omega^{\paren{t+w}} = \ldots = \Omega^{\paren{t+1}} \ne \Omega^{\paren{t}}$, then 
\begin{equation*}
\Pi^{\paren{t,w}}_s\coloneqq \set{-2\log\bb{P}\paren{X_{t+r,s}\mid X_{t+r,s'}:\;s'\ne s; \Omega^{\paren{t}}}+\log\Omega^{\paren{t}}_{ss}-\log\paren{2\pi}}^w_{r=1},
\end{equation*}
is a set of i.i.d. random variables that are not centered around one. Indeed there are some nodes for which
\begin{equation*}
\beta^{\paren{t,w}}_s\coloneqq \bb{E}\brac{\frac{-2}{w}\sum_{r=1}^w \log\bb{P}\paren{X_{t+r,s}\mid X_{t+r,s'}:\;s'\ne s; \Omega^{\paren{t}}}+\log\paren{\frac{\Omega^{\paren{t}}_{ss}}{2\pi}}\Big\rvert \mbox{There is an abrupt-change at } t} \ne 1.
\end{equation*}
Therefore for a non-negative convex function $f$ such that $f\paren{x}>0$ if and only if $x\ne 1$ (its exact formulation will be introduced in Section \ref{Section2.1}), after proper normalization $\sum_{s=1}^p f(\beta^{\paren{t,w}}_s)$ concentrates around zero, if no jump occurs at time $t$. Indeed we utilize a suitable convex barrier $f$ for designing a decision rule in order to magnify any deviation from pre-change pseudo-likelihood function. The exact formulation of this idea is postponed to Section \ref{DtctAlgo}. It is also noteworthy to mention that since the core building blocks of our procedure is based upon comparing pre and post change pseudo-likelihood of data, we believe that it can be extended to more generic Markov random fields. 

We study the performance of our algorithm in two settings: fully known and unknown $\Omega^{\paren{t}}$. Although the case of a fully known $\Omega^{\paren{t}}$ is not realistic, it provides insight in understanding our contribution, as well as the intrinsic complexity of the underlying detection problem. When the pre-change precision matrix is unknown, we require the change-points to be far from each other and the boundary point $t=0$. In this case, we first consistently estimate $\Omega^{\paren{1}}$ using any well-studied offline estimation algorithm such as QUIC \cite{hsieh2014quic} in the burn-in period ($t=1,\ldots,n_0$ for some large enough $n_0$). Note that the first abrupt-change is supposed to appear after $n_0$. After having an adequate quality estimate of $\Omega^{\paren{1}}$ in hand, we concurrently run the detection test and update our pre-change precision matrix in a sequential fashion. Our asymptotic analysis reveals that when both $p$ and sample size grow in such a way that we can consistently estimate pre-change precision matrix of GGM, the detection rate does not differ from the ideal case of a fully-known pre-change dependence structure. In other words, the detection power of our online algorithm is robust against small enough estimation errors, which is highly desirable for real applications.

The remainder of the paper is organized as follows:
Section \ref{Section2.1} rigorously formulates the online change-point problem as a hypothesis testing procedure and introduce the required statistical ingredients for understanding the subsequent sections. Section \ref{DtctAlgo} is devoted to presenting the proposed detection algorithm for both fully known and unknown pre-change attributes of the GGM. Section \ref{Section3} is reserved for investigating the behaviour of proposed decision rule under both null (no-change at $t$) and alternative (sudden change at $t$) hypotheses. In Section \ref{Section4}, we study asymptotic properties of our algorithm in the case of unknown pre-change precision matrix (combination of online detection and estimation). Section \ref{SimulSec} assesses the performance of proposed algorithm by numerical experiments on synthetic and real data. Section \ref{Discussion} serves as the conclusion and discusses future directions. We prove the main results of the paper in Section \ref{Proofs}. Lastly, Appendices \ref{AppendixA} and \ref{AppendixB} contain auxiliary technicalities which are essential in Section \ref{Proofs}.

\paragraph{Notation.} $\bbM{1}\paren{\cdot}$, $\wedge$ and $\vee$ successively refer to indicator function, minimum and maximum operators. We use $I_m$, $\zero_m$ and $\bbM{1}_m$ respectively denote the $m$-by-$m$ identity matrix, all zeros column vector of length $m$, and all ones column vector of length $m$. $S^{p\times p}_{++}$ stands for the space of strictly positive definite $p$ by $p$ matrices. For two matrices of the same size $M$ and $M'$, $\InnerProd{M}{M'}{}\coloneqq \sum_{i,j} M_{ij}M'_{ij}$ denotes their usual inner product. For $M\in\bb{R}^{p\times p}$ and $B_1,B_2\subset\set{1,\ldots,p}$, $M_{B_1,B_2} = \brac{M_{ij}: i\in B_1, j\in B_2}$ denote the sub-matrix of $M$ associated to $\paren{B_1,B_2}$-block. $\diag\paren{M}$ refers to the main diagonal entries of $M$. We use the following norms on $M\in\bb{R}^{m\times n}$. $\OpNorm{M}{2}{2}$ represents the usual operator norm (largest singular value of $M$). For any $1\leq p\leq\infty$, $\LpNorm{M}{p}$ stands for element-wise $\ell_p$-norm of vectorized $M$. For any $x > 0$, $\Gamma\paren{x}$ denotes the Gamma function at $x$ and $\psi^{\paren{r}}$ stands for the \emph{poly-gamma function} of order $r$, which is defined by 
\begin{equation*}
\psi^{\paren{r}}\paren{x} = \frac{d^{r+1}\log \Gamma\paren{x}}{dx^{r+1}},\quad\forall\;x>0 \mbox{ and }r=0,1,\ldots.
\end{equation*}
For two non-negative sequences $\set{a_m}^{\infty}_{m=1}$ and $\set{b_m}^{\infty}_{m=1}$, we write $a_m \lesssim b_m$ if there exists a bounded positive scalar $C_{\max}$ (depending on model parameters) such that $\limsup_{m\rightarrow\infty} a_m/b_m \leq C_{\max}$. Moreover, $a_m \asymp b_m$ refers to the case that $a_m \lesssim b_m$ and $a_m \gtrsim b_m$. For a non-negative deterministic $\set{a_m}^{\infty}_{m=1}$ and random sequence $\set{b_m}^{\infty}_{m=1}$, we write $b_n = \cc{O}_{\bb{P}}\paren{a_n}$, if $\bb{P}\paren{b_m\leq C_{\max} a_m} \rightarrow 1$, as $m\rightarrow\infty$, for some bounded positive scalar $C_{\max}$ (which may depend on model parameters). $D_{KL}\paren{\bb{P}_1 \;\| \;\bb{P}_2}$ represents the \emph{Kullback–Leibler} (KL) divergence between two distributions $\bb{P}_1$ and $\bb{P}_2$. Lastly for a binary test statistic $\Xi$, the false alarm and miss-detection probabilities are respectively defined by
\begin{equation*}
\boldsymbol{P}_{\FA}\paren{\Xi}\coloneqq \bb{P}\paren{\Xi = 1 \mid \bb{H}_0},\quad\mbox{and}\quad \boldsymbol{P}_{\MD}\paren{\Xi}\coloneqq \bb{P}\paren{\Xi = 0 \mid \bb{H}_1}.
\end{equation*} 

\section{Problem formulation and detection algorithm}\label{Methodology}

We start by providing a rigorous formulation of the problem at hand and then introduce the
detection strategy initially for the case of a known precision matrix before the change point which provides insight into the technical aspects of the problem, followed by the real world setting of an unknown precision matrix that needs to be estimated in an online fashion from the available data.

\subsection{Background and setup}\label{Section2.1}

Let $G_t = \paren{V,E_t},\;t \in\set{1,\ldots,T}$ be a time-varying undirected zero mean GGM with (fixed) node set $V$ of size $p$, where $T$ denotes the number of observed independent samples. For any $t\in\set{1,\ldots,T}$, we observe a single realization of $G_t$ represented by $\boldsymbol{X}_t = \brac{X_{t,1},\ldots, X_{t,p} }^\top$. In particular $\boldsymbol{X}_t$ is a $p$-variate Gaussian vector with density function 
\begin{equation*}
g\paren{\boldsymbol{x};\TimeVarQuant{\Omega}{\paren{t}}{sup}} = \paren{2\pi}^{-p/2} \sqrt{\det \TimeVarQuant{\Omega}{\paren{t}}{sup}} \exp\paren{ -\frac{\boldsymbol{x}^\top \TimeVarQuant{\Omega}{\paren{t}}{sup} \boldsymbol{x}}{2} },\quad\forall\; \boldsymbol{x}\in\bb{R}^p,
\end{equation*}
where $\TimeVarQuant{\Omega}{\paren{t}}{sup}$ denotes the symmetric positive definite precision matrix of $\boldsymbol{X}_t$. Note that $E_t$ is fully identifiable from the non-zero off-diagonal elements of $\TimeVarQuant{\Omega}{\paren{t}}{sup}$;
namely
\begin{equation*}
\TimeVarQuant{E}{t}{sub} = \set{ \paren{\alpha,\alpha'}\in V \times V:\; \alpha\ne \alpha',\; \TimeVarQuant{\Omega}{\paren{t}}{sup}_{\alpha,\alpha'} \ne 0 }.
\end{equation*}
Adopting a piecewise constant model for $\TimeVarQuant{\Omega}{\paren{t}}{sup}$ is a popular approach to modelling multiple abrupt changes in the dependency structure of $G_t$. Specifically, assume that there is a set $\cc{D}^\star\subset \set{1,\ldots,T}$, sorted in ascending order and with $t^\star_0 = 1$, such that
\begin{equation}\label{PiecewiseConstModel}
\TimeVarQuant{\Omega}{\paren{t}}{sup} = \sum_{j=1}^{T} \Omega_j \bb{I}\paren{t^\star_j \leq t < t^\star_{j+1}},\quad \forall\;t\in\set{1,\ldots,T}. 
\end{equation} 
In Eq. \eqref{PiecewiseConstModel} $\cc{D}^\star$ stands for the collection of unknown change-points between $1$ and $T$. As a consequence $\set{\boldsymbol{X}_t:\; t^\star_j \leq t < t^\star_{j+1}}$ are independent and identically distributed samples drawn from $g\paren{\cdot;\Omega_j}$.

Detecting jumps in $\TimeVarQuant{\Omega}{\paren{t}}{sup}$ is equivalent to solving $T$ separate hypothesis testing problems formulated by
\begin{equation}\label{HypTest}
\bb{H}_{0,t}: t\notin\cc{D}^\star\;\; \mbox{vs.} \;\; 
\bb{H}_{1,t}: t\in\cc{D}^\star,\quad \forall\;t\in\set{1,\ldots,T}.
\end{equation}
In offline setting, we observe the entire set $\set{\boldsymbol{X}_t}^T_{t=1}$ prior to test any hypothesis in \eqref{HypTest}. Simply put, a binary decision function $\Xi_t\paren{\boldsymbol{X}_1,\ldots,\boldsymbol{X}_T}\in\set{0,1}$ is designed for disambiguating $\bb{H}_{0,t}$ from $\bb{H}_{1,t}$ with false alarm rate below some $\pi_0\in\paren{0,1}$, for any $t\in\set{1,\ldots,T}$. On the other hand in the online regime, $\Xi_t$ solely depends on $\set{\boldsymbol{X}_i}^{t+w}_{i=1}$, for some pre-specified delay $w$. Henceforth, we focus exclusively on the online setting.

For better understanding the rational behind our proposed detection algorithm, we make a succinct overview of two key statistical concepts: KL-divergence and conditional log-likelihood of multivariate Gaussian vectors. We first introduce an alternative formulation of the KL-divergence between underlying distributions under $\bb{H}_{0,t}$ and $\bb{H}_{1,t}$. From a statistical viewpoint, the false alarm of $\Xi_t$, regardless of its complexity, is not negligible when the KL-divergence between $\bb{P}\paren{\boldsymbol{X}_1,\ldots,\boldsymbol{X}_{t+w} \mid \bb{H}_{0,t} }$ and $\bb{P}\paren{\boldsymbol{X}_1,\ldots,\boldsymbol{X}_{t+w} \mid \bb{H}_{1,t} }$ is small. For ease of presentation, we consider a simple scenario. Suppose that the inverse covariance matrix of $G_t$ switches from $\Omega_1$ to $\Omega_2$ at time $t$ and it remains in the new regime until $t+w$, i.e. $\cc{D}^\star\cap \set{t,\ldots,t+w} = \set{t}$. For notational convenience let $\Psi \coloneqq \Omega^{-1/2}_2 \Omega_1 \Omega^{-1/2}_2$ and define $f:\paren{0,\infty}\mapsto\brapar{0,\infty}$ by
\begin{equation}\label{f}
f\paren{x} \coloneqq x - 1 - \log x,\quad\forall\;x>0,
\end{equation}
Since $\set{\boldsymbol{X}_{r}}^{t+w}_{r=1}$  independent and zero-mean random vectors under both $\bb{H}_{0,t}$ and $\bb{H}_{1,t}$, the KL-divergence between $\bb{P}\paren{\boldsymbol{X}_1,\ldots,\boldsymbol{X}_{t+w} \mid \bb{H}_{1,t} }$ and $\bb{P}\paren{\boldsymbol{X}_1,\ldots,\boldsymbol{X}_{t+w} \mid \bb{H}_{0,t} }$, which is denoted by $\bb{D}$, can be written as
\begin{eqnarray}\label{KL-DivGauss}
\bb{D} &=&  \sum_{1\leq r\leq w} D_{KL}\Bigparen{ \bb{P}\paren{\boldsymbol{X}_{t+r}\mid \bb{H}_{1,t} } \;\|\; \bb{P}\paren{\boldsymbol{X}_{t+r}\mid \bb{H}_{0,t} } } = wD_{KL}\Bigparen{ \bb{P}\paren{\boldsymbol{X}_{t+1}\mid \bb{H}_{1,t} } \;\|\; \bb{P}\paren{\boldsymbol{X}_{t+1}\mid \bb{H}_{0,t} } } \nonumber\\
&=&  w \brac{ \tr\paren{\Omega^{-1}_2\Omega_1 }- \log\paren{\frac{\det \Omega_1}{\det \Omega_2}} - p } = w\Bigparen{ \tr\paren{\Psi}- \log\det \Psi - p } = w\sum_{j=1}^{p} f\Bigparen{ \lambda_j\paren{\Psi} }.
\end{eqnarray}
The conditional log-likelihood of a GGM is another key component in the proposed detection algorithm. It is known that for a Gaussian random vector $\boldsymbol{X} = \brac{X_1,\ldots,X_p}^\top \sim \cc{N}\paren{\zero_p, \Sigma = \Omega^{-1}}$, the log-likelihood function of $X_s$ given $X_{-s}$ satisfies the following identity for any $s\in\set{1,\ldots,p}$.
\begin{eqnarray}\label{Z_s}
Z_s &\coloneqq& -2\log \bb{P}\paren{ X_s|X_{-s} } - \log\paren{\frac{2\pi}{\Omega_{ss}}} = \Omega_{ss}\paren{X_s + \sum_{t\ne s} \frac{X_t\Omega_{st} }{\Omega_{ss}}}^2 = \Omega_{ss}\paren{\frac{\sum_{t=1}^{p} X_t\Omega_{st}}{\Omega_{ss}}}^2\nonumber\\ &=&\frac{\InnerProd{\boldsymbol{X}}{\Omega_{s,:}}^2}{\Omega_{ss}}.
\end{eqnarray}
Note that $Z_s$ represents the data dependent part of the conditional negative log-likelihood. Moreover,
\begin{equation*}
\var\paren{\InnerProd{\boldsymbol{X} }{\Omega_{:,s}}} =\Omega_{s,:} \cov\paren{\boldsymbol{X}} \Omega_{:,s} = \Omega_{s,:}\Omega^{-1}\Omega_{:,s} = \paren{\Omega \Omega^{-1} \Omega}_{ss} = \Omega_{ss}.
\end{equation*}
That is, $\set{Z_s:s=1\ldots,p}$ forms a class of dependent $\chi^2_1$ random variables.

\subsection{Detection algorithm for a known pre-change precision matrix}\label{DtctAlgo}

Next, we introduce a novel online algorithm for detecting abrupt changes in $\Omega^{\paren{t}}$ with delay $w$; namely, determine whether $\paren{t+1} \in \cc{D}^\star$, given $\set{\boldsymbol{X}_1,\ldots, \boldsymbol{X}_t, \ldots, \boldsymbol{X}_{t+w} }$. To gain insights in to the nature of the problem, we first consider the oracle framework in which the pre-change precision matrix $\Omega^{\paren{t}}$ is fully known, that allows us to solely focus on the detection procedure without having to estimate the pre-change parameters. The setting of an unknown $\Omega^{\paren{t}}$ is addressed in Section \ref{Section2.3}.  

As we discussed in Section \ref{Intro}, the proposed test statistic is motivated by employing a network based pseudo-likelihood function. For any $s\in\set{1,\ldots,p}$, define
\begin{equation}\label{Y^tw_s}
Y^{\paren{t,w}}_{s} \coloneqq \frac{1}{w\Omega^{\paren{t}}_{ss} } \sum_{r = 1}^{w}\InnerProd{\boldsymbol{X}_{t+r} }{\Omega^{\paren{t}}_{:,s}}^2.
\end{equation}
Based on Eq. \eqref{Z_s}, $Y^{\paren{t,w}}_{s}$ is indeed the empirical average of the (shifted) negative conditional log-likelihood of node $V_s$ given all the other nodes in $G_t$. When dealing with GGMs, $Y^{\paren{t,w}}_{s}$ is as a linear combination of $w$ independent quadratic forms of Gaussian random variables. Next, we investigate the distribution of $Y^{\paren{t,w}}_{s}$ under $\bb{H}_{0,t}$. In this case $\Omega^{\paren{t}} = \Omega^{\paren{t+1}} = \ldots = \Omega^{\paren{t+w}}$. Thus, for any $r\in\set{1,\ldots,w}$, $\InnerProd{\boldsymbol{X}_{t+r} }{\Omega^{\paren{t}}_{:,s}}$ is a centered Gaussian random variable whose variance is given by
\begin{equation}\label{Y^tw_sDistH0}
\var\paren{\InnerProd{\boldsymbol{X}_{t+r} }{\Omega^{\paren{t}}_{:,s}}} =\Omega^{\paren{t}}_{s,:} \cov\paren{X_{t+r}} \Omega^{\paren{t}}_{:,s} = \Omega^{\paren{t}}_{s,:}\brac{\Omega^{\paren{t}}}^{-1}\Omega^{\paren{t}}_{:,s} = \paren{\Omega^{\paren{t}} \brac{\Omega^{\paren{t}}}^{-1} \Omega^{\paren{t}}}_{ss} = \Omega^{\paren{t}}_{ss}.
\end{equation}
Identity \eqref{Y^tw_sDistH0} reveals that when $\bb{H}_{0,t}$ holds, then
\begin{equation*}
\frac{\InnerProd{\boldsymbol{X}_{t+r} }{\Omega^{\paren{t}}_{:,s}}}{ \sqrt{\Omega^{\paren{t}}_{ss}} } \sim \cc{N}\paren{0,1} \;\;\Longrightarrow\;\; wY^{\paren{t,w}}_{s} \sim \chi^2_w,\quad\forall\;s\in\set{1,\ldots,p}.
\end{equation*}
According to the strong law of large numbers, $Y^{\paren{t,w}}_{s}$ concentrates around one with high probability for all $s = 1,\ldots, p$, as $w$ increases. In contrast, under $\bb{H}_{1,t}$, whenever $\Omega^{\paren{t}} \ne  \Omega^{\paren{t+1}} = \ldots = \Omega^{\paren{t+w}}$, the expected value of $Y^{\paren{t,w}}_{s}$ is given by
\begin{equation}\label{Y^tw_sDistH1}
\bb{E}\paren{Y^{\paren{t,w}}_{s} \mid \bb{H}_{1,t} } = \frac{1}{w} \sum_{r=1}^{w}\frac{\paren{\Omega^{\paren{t}} \brac{\Omega^{\paren{t+1}}}^{-1} \Omega^{\paren{t}}}_{ss} }{\Omega^{\paren{t}}_{ss}} = \frac{\paren{\Omega^{\paren{t}} \brac{\Omega^{\paren{t+1}}}^{-1} \Omega^{\paren{t}}}_{ss} }{\Omega^{\paren{t}}_{ss}}.
\end{equation}

\begin{rem}
Define $\Psi^{\paren{t}}\in S^{p\times p}_{++}$ by $\Psi^{\paren{t}} \coloneqq \brac{\Omega^{\paren{t}}}^{1/2} \brac{\Omega^{\paren{t+1}}}^{-1} \brac{\Omega^{\paren{t}}}^{1/2}$. The formulation of KL-divergence between multi-variate Gaussian densities in Eq. \eqref{KL-DivGauss} shows that distinguishing $\bb{H}_{0,t}$ and $\bb{H}_{1,t}$ is not possible when $\Psi^{\paren{t}} = I_p$. On the other hand, careful comparison between Eq. \eqref{Y^tw_sDistH0} and \eqref{Y^tw_sDistH1} reveals a slightly stronger condition for distinguishing $\bb{H}_{0,t}$ from $\bb{H}_{1,t}$. Strictly speaking conditional log-likelihood terms can not differentiate null and alternative hypotheses, when $\diag\paren{ \Psi^{\paren{t}} } = \bbM{1}_p$. In words, a test based on the pseudo-likelihood function rules out the possibility of detecting abrupt changes at $t+1$ when
\begin{equation*}
\Psi^{\paren{t}} \ne I_p,\;\;\mbox{and}\;\; \Psi^{\paren{t}}_{ss} = 1,\quad \forall\;s = 1,\ldots, p.
\end{equation*}
Although this may seem a statistical drawback of using the pseudo-likelihood function for detection purposes, we believe that such situations do not arise in many practical scenarios involving high-dimensional streaming network data. Therefore for scalability purposes, it is still of great interest to design online change-point detection algorithms based on the pseudo-likelihood function. Further, such an algorithm (unlike likelihood-based detection algorithms) can be also applicable to more general Markov random field models. 
\end{rem}

Next, we introduce our online detection algorithm. Recall function $f$ from Eq. \eqref{f} and consider the following test statistic.
\begin{equation*}
T_t = \frac{ \sum_{s=1}^{p} f\paren{ Y^{\paren{t,w}}_{s} } - \bb{E} \brac{ \sum_{s=1}^{p} f\paren{ Y^{\paren{t,w}}_{s} } \Big\rvert \bb{H}_{0,t} } }{ \sd\brac{\sum_{s=1}^{p} f\paren{ Y^{\paren{t,w}}_{s} } \Big\rvert \bb{H}_{0,t} }}.
\end{equation*}
For a pre-specified $\pi_0\in\paren{0,1}$ (denoting the desirable false alarm rate) and a pre-determined critical value $\zeta_{\pi_0}$, we design the following binary decision function 
\begin{equation}\label{Test}
\Xi_t = \bbM{1}\paren{ T_t \geq \zeta_{\pi_0} },
\end{equation}
for selecting between $\bb{H}_{0,t}$ and $\bb{H}_{1,t}$. Notice that we raise an abrupt change flag, whenever $\Xi_t = 1$. Namely, $\Omega^{\paren{t}} = \Omega^{\paren{t+1}} = \ldots = \Omega^{\paren{t+w}}$, as long as $T_t$ is strictly less than $\zeta_{\pi_0}$.

Before looking closely into technical aspects of the proposed algorithm, we concisely explain the rationale behind our approach. We argued that if the dependence structure of $G_t$ does not vary at $t+1$, all random variables $Y^{\paren{t,w}}_{s},\;s=1,\ldots,p$ lie in a neighborhood of one. We also opted a non-negative strongly convex barrier function $f$, whose single root is at $x = 1$. Hence, $f(Y^{\paren{t,w}}_{s})$ lies around zero, for any $s$. In contrast, when the network undergoes an abrupt change, $\bb{E}f(Y^{\paren{t,w}}_{s})$ is strictly positive for a some nodes $s\in\set{1,\ldots,p}$. As a result, $T_t$ exhibits relatively larger values under the alternative hypothesis. Finally, due to the convexity of $f$, the deviation between $\bb{H}_{0,t}$ and $\bb{H}_{1,t}$ is more pronounced in $T_t$ for stronger changes in $\Omega^{\paren{t}}$. 

The first question that needs to be addressed is how to simply standardize $\sum_{s=1}^{p} f(Y^{\paren{t,w}}_{s})$ under $\bb{H}_{0,t}$. 

\begin{figure}
	\centering
	\input{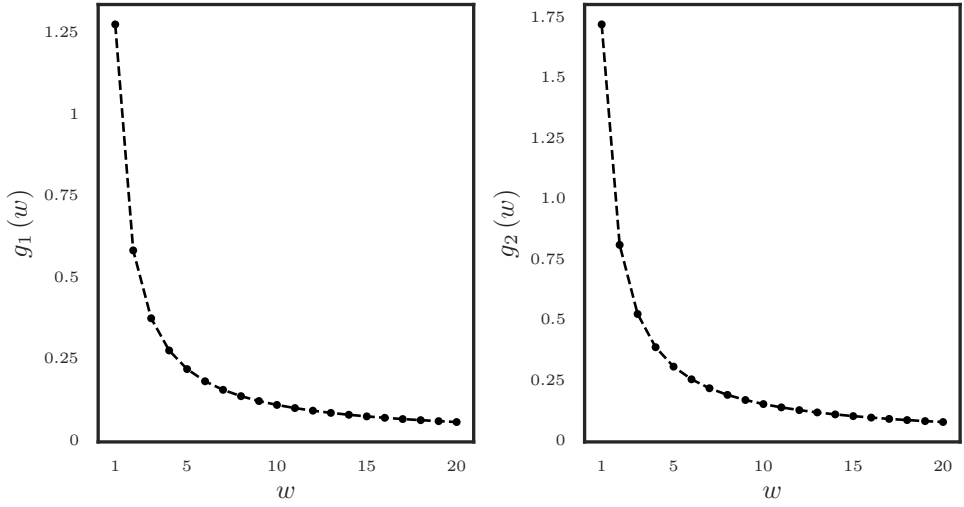} 
	\caption{black dots in the left and right panels respectively exhibit $g_1\paren{w}$ and $g_2\paren{w}$ for $w\in\set{1,\ldots,20}$.}
	\label{Fig:Fig1}
\end{figure}

\begin{rem}\label{Rem2.2}
One can easily justify that under $\bb{H}_{0,t}$, $\InnerProd{\boldsymbol{X}_{t+r} }{\Omega^{\paren{t}}_{:,s}}/\sqrt{\Omega^{\paren{t}}_{ss}},\;r = 1,\ldots,w$, form a set of i.i.d. standard Gaussian random variables, for any $s\in\set{1,\ldots,p}$. Thus, $wY^{\paren{t,w}}_{s}$ has a chi-square density with $w$ degrees of freedom. Lemma \ref{Lemma1App} states that
\begin{eqnarray}\label{Ef}
g_1\paren{w}&\coloneqq& \bb{E} \brac{ f\paren{ Y^{\paren{t,w}}_{s} } \Big\rvert \bb{H}_{0,t} } = \log\paren{\frac{w}{2}} - \psi^{\paren{0}}\paren{\frac{w}{2}}, \nonumber\\
g_2\paren{w}&\coloneqq& \sd\brac{ f\paren{ Y^{\paren{t,w}}_{s} } \Big\rvert \bb{H}_{0,t} } = \sqrt{\psi^{\paren{1}}\paren{\frac{w}{2}}-\frac{2}{w}},\quad \forall\;s = 1,\ldots,p.
\end{eqnarray}
So $T_t$ can be rewritten in the following form.
\begin{equation*}
T_t = \frac{ \sum_{s=1}^{p} f\paren{ Y^{\paren{t,w}}_{s} } - pg_1\paren{w} }{ \sd\brac{\sum_{s=1}^{p} f\paren{ Y^{\paren{t,w}}_{s} } \Big\rvert \bb{H}_{0,t} } }.
\end{equation*}
$g_1\paren{\cdot}$ and $g_2\paren{\cdot}$ are respectively exhibited in the left and right  panels of Figure \ref{Fig:Fig1}. It is obvious from Figure \ref{Fig:Fig1} that both $g_1\paren{w}$ and $g_2\paren{w}$ converge to zero, as $w\rightarrow\infty$. In particular, it is known that\footnote{We refer the reader to $5.11.2$ of \url{https://dlmf.nist.gov/5.11} and $5.15.8$ of \url{https://dlmf.nist.gov/5.15}}
\begin{equation*}
\lim\limits_{w\rightarrow\infty} wg_1\paren{w} = 1, \quad\mbox{and} \quad \lim\limits_{w\rightarrow\infty} wg_2\paren{w}=\sqrt{2}.
\end{equation*}
\end{rem}

Next, we evaluate the denominator of $T_t$. To do so, we first define the partial correlation matrix of $G_t$, denoted by $R^{\paren{t}}$.
\begin{equation*}
R^{\paren{t}} \coloneqq \brac{ \frac{\Omega^{\paren{t}}_{s_1,s_2}}{ \sqrt{ \Omega^{\paren{t}}_{s_1,s_1} \Omega^{\paren{t}}_{s_2,s_2} } } }^p_{s_1,s_2=1} \in \bb{R}^{p\times p}.
\end{equation*}
Straightforward calculations show that for standard Gaussian random variables $Z_1$ and $Z_2$ with correlation $r$, $\cov\paren{Z^2_1, Z^2_2} = 2r^2$. This fact together with the linearity property of the covariance function imply that
\begin{equation}\label{CovY}
\cov\paren{ Y^{\paren{t,w}}_{s_1}, Y^{\paren{t,w}}_{s_2} \mid \bb{H}_{0,t} } = \frac{2}{w} \paren{R^{\paren{t}}_{s_1,s_2}}^2,\quad\forall\; s_1,s_2\in\set{1\ldots,p}.
\end{equation}
Simply put, $R^{\paren{t}}$ encodes the dependence between conditional log-likelihood functions in $G_t$. Now employing the second identity in Eq. \eqref{Ef} yields
\begin{equation*}
\var\brac{\sum_{s=1}^{p} f\paren{ Y^{\paren{t,w}}_{s} } \Big\rvert \bb{H}_{0,t} } = g^2_2\paren{w} \sum_{s_1,s_2=1}^{p} \corr\brac{ f\paren{ Y^{\paren{t,w}}_{s_1} }, f\paren{ Y^{\paren{t,w}}_{s_2} } \Big\rvert \bb{H}_{0,t} }.
\end{equation*}
Finding a closed form expression for the correlation between $f(Y^{\paren{t,w}}_{s_1})$ and $f(Y^{\paren{t,w}}_{s_2})$, under $\bb{H}_{0,t}$, demands algebraically cumbersome manipulations. However, it is not too difficult to see that it only depends on $R^{\paren{t}}_{s_1,s_2}$ and $w$. Namely, there exists $h_w:\brac{-1,1}\mapsto\brac{-1,1}$ such that
\begin{equation}\label{h_w}
\corr\brac{ f\paren{ Y^{\paren{t,w}}_{s_1} }, f\paren{ Y^{\paren{t,w}}_{s_2} } \Big\rvert \bb{H}_{0,t} } = h_w\paren{ R^{\paren{t}}_{s_1,s_2} }.
\end{equation}
\begin{rem}
We employ numerical techniques for approximating $h_w$. Figure \ref{Fig:Fig2} displays $h_w$ for different values of $w$. The plot in the right panel in Figure \ref{Fig:Fig2} depicts that $h_w\paren{r}\approx r^4$ for $w\geq 10$. In Lemma \ref{Lemma3App}, we rigorously substantiate this observation by introducing a uniform upper bound on $\abs{h_w\paren{r} - r^4}$. Specifically, we show that 
\begin{itemize}
\item $h_w$ passes the origin and $h_w\paren{1} = h_w\paren{-1} = 1$.
\item As $w\rightarrow\infty$, $\max_{r\in\brac{-1,1}} \abs{h_w\paren{r} - r^4} = \cc{O}\paren{w^{-1}}$.
\end{itemize}
Approximating $h_w\paren{r}$ with $r^4$ offers the following numerically convenient proxy for $T_t$ for large enough $w$. 
\begin{equation}\label{ApproxTt}
T_t = \frac{ \sum_{s=1}^{p} \brac{f\paren{ Y^{\paren{t,w}}_{s} } - g_1\paren{w} } }{ g_2\paren{w} \sqrt{\sum_{s_1,s_2=1}^{p} h_w\paren{ R^{\paren{t}}_{s_1,s_2} } } } \approx \frac{ \sum_{s=1}^{p} \brac{f\paren{ Y^{\paren{t,w}}_{s} } - g_1\paren{w} } }{ g_2\paren{w} \LpNorm{ R^{\paren{t}} }{4}^2 }.
\end{equation}
\end{rem}
\begin{figure}
\centering
\input{hw.pgf} 
\caption{$h_w\paren{x}$ versus $x$ and $x^4\sign\paren{x}$ for $w = 1, 5, 10$ and $20$.}
\label{Fig:Fig2}
\end{figure}
In summary, when $\Omega^{\paren{t}}$ is fully known, we propose the following sequential test for detecting a sudden change at time $t+1$.
\begin{equation}\label{Xi_t}
\Xi_t = \bbM{1}\paren{ \frac{ \sum_{s=1}^{p} \brac{f\paren{ Y^{\paren{t,w}}_{s} } - g_1\paren{w} } }{ g_2\paren{w} \sqrt{\sum_{s_1,s_2=1}^{p} h_w\paren{ R^{\paren{t}}_{s_1,s_2} } } } \geq \zeta_{\pi_0} }
\end{equation}

\subsection{Detection algorithm: Unknown pre-change precision matrix}\label{Section2.3}

The precision matrix is usually unknown before the change-point and needs to be estimated from the data. In this case, $T_t$ is approximated by plugging into Eq. \eqref{ApproxTt} a positive definite estimate of $\Omega^{\paren{t}}$. Specifically, let $\hat{\Omega}^{\paren{t}}$ be an estimate of $\Omega^{\paren{t}}$. Then, the partial correlation matrix $R^{\paren{t}}$ can be estimated by
\begin{equation*}
\hat{R}^{\paren{t}} = \brac{\frac{\hat{\Omega}^{\paren{t}}_{ij} }{ \sqrt{\hat{\Omega}^{\paren{t}}_{ii}\hat{\Omega}^{\paren{t}}_{jj}} }}^p_{i,j=1}.
\end{equation*}
Moreover for any $s\in\set{1,\ldots,p}$, we can estimate $Y^{\paren{t,w}}_s$ (recall it from Eq. \eqref{Y^tw_s}) in the following way.
\begin{equation*}
\hat{Y}^{\paren{t,w}}_s \coloneqq \frac{1}{w\hat{\Omega}^{\paren{t}}_{ss} } \sum_{r = 1}^{w}\InnerProd{\boldsymbol{X}_{t+r} }{\hat{\Omega}^{\paren{t}}_{:,s}}^2.
\end{equation*}
When we have access to $\hat{R}^{\paren{t}}$ and $\set{\hat{Y}^{\paren{t,w}}_s}^p_{s=1}$, we propose to estimate $T_t$ and $\Xi_t$ by
\begin{equation}\label{THat_t}
\hat{T}_t = \frac{ \sum_{s=1}^{p} \brac{f\paren{ \hat{Y}^{\paren{t,w}}_{s} } - g_1\paren{w} } }{ g_2\paren{w} \sqrt{\sum_{s_1,s_2=1}^{p} h_w\paren{ \hat{R}^{\paren{t}}_{s_1,s_2} } } },\quad \hat{\Xi}_t = \bbM{1}\paren{\hat{T}_t \geq \zeta_{\pi_0}}.
\end{equation}
As seen in Eq. \eqref{THat_t}, the key to approximating $T_t$ is the availability of a good estimate of $\Omega^{\paren{t}}$, which is a challenging task, especially in settings where consecutive change-points are close to each other. Thus, adequate separation in time between two consecutive change-points should be present for obtaining a good approximation of the proposed test statistic. The following assumption formalizes this notion. Recall that $t^\star_j$ denotes the location of $j$-th change-point and for notational consistency, we choose $t^\star_0 = 1$.

\begin{assu}\label{Assu2.1}
For each $j\in\bb{N}$, there exists a large enough $n_j\in\bb{N}$ (depending on $p, w$ and sparsity pattern of network between $j$-th and $\paren{j-1}$-th abrupt change) such that 
\begin{equation*}
\abs{t^\star_j-t^\star_{j-1}} > n_j,\;\forall\; j\geq 1.
\end{equation*}
\end{assu}

Note that Assumption \ref{Assu2.1} generalizes the boundary condition for the offline single change-point estimation problem (see Assumption $2$ in \cite{roy2016change}). For the time being, selection of $n_j$ is postponed for later sections. We refer to the first $n_j$ samples after $t^\star_j$ as the \emph{burn-in period}. We also assume that there exists a fixed, bounded window size $n_0$ such that $n_j = n_0$ for any $j\geq 1$.

We first provide the intuition behind the algorithm. Detecting each change-point goes through two phases: \emph{warm-up} and \emph{detection-estimation cycle}. For simplicity, we only focus on detecting the first abrupt change (located at $t^\star_1$). Note that for each $t\leq t^\star_1$, $\Omega^{\paren{t}} = \Omega^{\paren{1}}$.
\begin{enumerate}[label = (\alph*),leftmargin=*]
	\item (warm-up) We estimate $\Omega^{\paren{1}}$, which is denoted by $\hat{\Omega}^{\paren{1}}$, using $\boldsymbol{X}_1,\ldots,\boldsymbol{X}_{n_0}$.
	\item For any $t>n_0$, as long as $\hat{\Xi}_t = 0$, we use $\boldsymbol{X}_t$ for updating $\hat{\Omega}^{\paren{1}}$. In contrast if $\hat{\Xi}_t = 1$ (an abrupt change at $t$), then we wait for $\boldsymbol{X}_{t+1},\ldots,\boldsymbol{X}_{t+n_0}$ for estimating the post-change inverse covariance matrix.
\end{enumerate}

Next, we describe our approach to updating the estimated pre-change precision matrix in phase $\paren{b}$. Again for simplicity, we only focus on updating $\Omega^{\paren{1}}$.

\paragraph{Updating $\Omega^{\paren{1}}$.} A mini-batch procedure is used for updating $\Omega^{\paren{1}}$, wherein we first obtain $B$ (a predetermined block size) new samples and subsequently a new estimate at time $t = n+kB$ ($k\in\bb{N}$) by employing $\boldsymbol{X}_1,\ldots,\boldsymbol{X}_{n_0+kB}$; the parameter $k$ tracks the number of size-$B$ batches before the first abrupt change. Throughout this paper, we employ the CLIME algorithm \cite{cai2011constrained} or alternatively the QUIC estimator \cite{hsieh2014quic} for estimating $\Omega^{\paren{1}}$, since both enjoy desirable theoretical and numerical properties. The detailed pseudocode of this proposed procedure is presented in \emph{Algorithm $1$}.

\begin{center}\label{Algo1}
	\begin{tabular}{ >{\centering\arraybackslash}m{6in} >{\centering\arraybackslash}m{0.5in} }
		\thickhline
		\vspace{1mm}
		\begingroup
		\fontsize{12pt}{12pt}\selectfont
		\textbf{Algorithm 1} Online detection with batch update of pre-change precision matrix 
		\endgroup\\		
		\hline
		\multicolumn{1}{l}{\textbf{Input:} $n_0, w, B, \zeta_{\pi_0}$ and tuning parameter $\tau$}\\
		\multicolumn{1}{l}{\textbf{Initialization} Set $\hat{\cc{D}} = \emptyset$ and $b = 0$. Given $\boldsymbol{X}_1,\ldots,\boldsymbol{X}_{n_0}$, obtain $\Omega^{\paren{1}}$ by CLIME or QUIC with tuning}\\
		\multicolumn{1}{l}{parameter $\tau$. Also set $\hat{t}_{last} = 0$, where $\hat{t}_{last}$ denotes the estimated location of the last change-point.}\\
		\multicolumn{1}{l}{\textbf{Iterate} For $t = \paren{n_0+1},\ldots,T$} \\
		\begin{itemize}
			\item[] Set $\Xi_t = \bbM{1}\paren{\hat{T}_t \geq \zeta_{\pi_0} }$.
			\item[] If $\Xi_t = 0$ (no change-point)
			
			\begin{itemize}
				\item[] $b \leftarrow b+1$ and $t \leftarrow t+1$.
				\item[] If $b = B$ (Update pre-change precision matrix)
				\begin{itemize}
					\item[] Obtain $\hat{\Omega}^{\paren{t}}$ using the CLIME or QUIC methods to $\boldsymbol{X}_{1+\hat{t}_{last}},\ldots, \boldsymbol{X}_{t-1},\boldsymbol{X}_{t}$.
					\item[] $b\leftarrow 0$.
				\end{itemize}
				\item[] Else
				\begin{itemize}
					\item[] $\hat{\Omega}^{\paren{t}} = \hat{\Omega}^{\paren{t-1}}$
				\end{itemize}
			\end{itemize}
			
			\item[] Else
			\begin{itemize}
				\item[] $\hat{t}_{last} = t$ and $\hat{\cc{D}} \leftarrow \hat{\cc{D}} \cup \set{{\hat{t}_{last}}}$.
				\item[] Given $\boldsymbol{X}_{t},\ldots, \boldsymbol{X}_{t+n_0-1}$, estimate post-change precision matrix using CLIME or QUIC methods.
				\item[] $t \leftarrow t+n_0$ and $b\leftarrow 0$.
			\end{itemize}
			
		\end{itemize}\\
		
		\multicolumn{1}{l}{\textbf{Output: $\hat{\cc{D}}$}}\\
		\hline
	\end{tabular}
\end{center}

\comment{\item In the online framework, we update pre-change precision matrix after receiving a new observation using \emph{regularized dual averaging (RDA)} procedure \cite{xiao2010dual} with a choice regularization parameter $\tau$. Strictly speaking, until detecting the first change-point $\hat{\Omega}^{\paren{t}}$ is updated by
\begin{equation}\label{RDAUpdate}
\hat{\Omega}^{\paren{t+1}} = \frac{t\hat{\Omega}^{\paren{t}} + \phi_{t+1} }{t+1} \;\;\mbox{with}\;\;\phi_{t+1} = \argmin_{\phi}\brac{ \InnerProd{\sum_{r=1}^t \frac{ \paren{\boldsymbol{X}_r\boldsymbol{X}^\top_r - \phi^{-1}_r}}{t}}{\phi}+\tau_1\LpNorm{\phi}{1} + \frac{\tau_2}{2\sqrt{t}}\LpNorm{\phi}{2}^2 }.
\end{equation}
In Eq.\eqref{RDAUpdate}, the term $\paren{\boldsymbol{X}_r\boldsymbol{X}^\top_r - \phi^{-1}_r}$ is the gradient of $-2\log$-likelihood of $\boldsymbol{X}_r$ with inverse covariance matrix $\phi_r$. Note that $\phi_{t+1}$ can be equivalently rewritten in the following way. 
\begin{equation*}
\phi_{t+1} = \argmin_{\phi} \brac{ \frac{1}{2}\LpNorm{\phi + \frac{1}{\tau_2\sqrt{t}}\sum_{r=1}^t  \paren{\boldsymbol{X}_r\boldsymbol{X}^\top_r - \phi^{-1}_r} }{2}^2 + \frac{\tau_1\sqrt{t}}{\tau_2}\LpNorm{\phi}{1} }.
\end{equation*}
Therefore $\phi_{t+1}$ has a closed-form solution given by (see Eq. $10$ of \cite{xiao2010dual})
\begin{equation*}
\phi_{t+1} = \brac{ -\frac{\sqrt{t}}{\tau_2}\paren{ \paren{\bar{g}_t}_{ij}-\tau_1\sign(\paren{\bar{g}_t}_{ij})  } \bbM{1}\paren{ \abs{\paren{\bar{g}_t}_{ij}\geq \tau_1} } }^p_{i,j=1}, \quad\mbox{where}\quad \bar{g}_t = \frac{1}{t}\sum_{r=1}^t  \paren{\boldsymbol{X}_r\boldsymbol{X}^\top_r - \phi^{-1}_r}.
\end{equation*}
We finally refer the reader to \emph{Algorithm $2$} for a complete pseudocode.

\begin{center}\label{Algo2}
	\begin{tabular}{ >{\centering\arraybackslash}m{6in} >{\centering\arraybackslash}m{0.5in} }
		\thickhline
		\vspace{1mm}
		\begingroup
		\fontsize{12pt}{12pt}\selectfont
		\textbf{Algorithm 2} Online detection with online update of pre-change precision matrix 
		\endgroup\\		
		\hline
		\multicolumn{1}{l}{\textbf{Input:} $n, w, \zeta_{\pi_0}$ and tuning parameter $\tau$}\\
		\multicolumn{1}{l}{\textbf{Initialization} Set $\hat{\cc{D}} = \emptyset$. Given $\boldsymbol{X}_1,\ldots,\boldsymbol{X}_{n}$, estimate $\Omega^{\paren{1}}$ using RDA algorithm with tuning}\\
		\multicolumn{1}{l}{parameter $\tau$. Also set $\hat{t}_{last} = 0$, where $\hat{t}_{last}$ denotes the estimated location of the last change-point.}\\
		\multicolumn{1}{l}{\textbf{Iterate} For $t = \paren{n+1},\ldots,T$} \\
		\begin{itemize}
			\item[] Set $\Xi_t = \bbM{1}\paren{\hat{T}_t \geq \zeta_{\pi_0} }$.
			\item[] If $\Xi_t = 0$ (no change-point)
			
			\begin{itemize}
				\item[] Update $\hat{\Omega}^{\paren{t}}$ by Eq. \eqref{RDAUpdate} 
			\end{itemize}
			
			\item[] Else
			\begin{itemize}
				\item[] $\hat{t}_{last} = t$ and $\hat{\cc{D}} \leftarrow \hat{\cc{D}} \cup \set{{\hat{t}_{last}}}$.
				\item[] $t \leftarrow t+n$ and $b\leftarrow 0$.
				\item[] Given $\boldsymbol{X}_{t},\boldsymbol{X}_{t-1},\ldots, \boldsymbol{X}_{t-n+1}$, estimate post-change precision matrix using CLIME of QUIC methods.
			\end{itemize}	
		\end{itemize}\\
		\multicolumn{1}{l}{\textbf{Output: $\hat{\cc{D}}$}}\\
		\hline
	\end{tabular}
\end{center}
\end{enumerate}}

\section{Asymptotic analysis of $\Xi_t$ for a fully known $\Omega^{\paren{t}}$}\label{Section3}

Next, we establish large-sample properties of the proposed test $\Xi_t$ introduced in Eq. \eqref{Xi_t}, under both the null and alternative hypotheses. The section addresses the following three issues:
\begin{itemize}
\item Choosing the critical value of the test, $\zeta_{\pi_0}$, for a fixed false alarm probability $\pi_0\in\paren{0,1}$. 
\item Investigating the false alarm and detection power of $\Xi_t$ for large sparse graphs.
\item Introducing the key concepts for the asymptotic analysis of the algorithm in the realistic case of an unknown pre-change precision matrix.
\end{itemize}
We present the obtained results in two sub-sections. Section \ref{Section3.1} studies the distribution of $T_t$ (recall its formulation from Eq. \eqref{ApproxTt}) under the null hypothesis $\bb{H}_{0,t}$ and introduces a simple way of choosing the critical value $\zeta_{\pi_0}$. In Section \ref{Section3.2}, we investigate the statistical power of $T_t$ for detecting small changes in the structure of $\Omega^{\paren{t}}$. We start by defining the set of well-behaved precision matrices.

\begin{defn}\label{WellPosedMatrixSet}
Let $M$ and $\alpha_{\min}$ be two bounded and strictly positive scalars. Define $\cc{C}^{p\times p}_{++}\paren{\alpha_{\min}, M}$ by
\begin{equation*}
\cc{C}^{p\times p}_{++}\paren{\alpha_{\min}, M} = \set{A\in S^{p\times p}_{++}:\; \OpNorm{A}{1}{1}\leq M,\; \lambda_{\min}\paren{A}\geq \alpha_{\min} }.
\end{equation*}
\end{defn}

For theoretical purposes, throughout this section both pre- and post-change precision matrices are assumed to belong to $\cc{C}^{p\times p}_{++}\paren{\alpha_{\min}, M}$ for some bounded scalars $M$ and $\alpha_{\min}$. Namely, although an abrupt change may affect the topology of the network, the post-change precision matrix is still well-behaved, exhibiting bounded condition number and sparse edge set.

\subsection{Distribution of $T_t$ under $\bb{H}_{0,t}$}\label{Section3.1}

Precise evaluation of null distribution is extremely challenging, due to complex nature of $T_t$. Therefore, for approximating $\zeta_{\pi_0}$, especially for large networks, we inevitably focus on finding the asymptotic null distribution of $T_t$. Let $d_{\max}$ and $\bar{d}$ respectively denote the maximum and average degree of $G_t$. Namely 
\begin{equation*}
d_{\max} \coloneqq \max_{s_1=1,\ldots,p} \abs{\set{ s_2: \Omega^{\paren{t}}_{s_1,s_2} \ne 0 }},\quad\mbox{and}\quad \bar{d} \coloneqq \frac{1}{p}\sum_{s_1=1}^{p} \abs{\set{ s_2: \Omega_{s_1,s_2} \ne 0 }}.
\end{equation*}
For brevity, the dependence of $p$ and $t$ in $d_{\max}$ and $\bar{d}$ are dropped. Throughout the remainder of the paper, \emph{asymptotic regime} refers to the scenario that $p$, and possibly $d_{\max}$ and $\bar{d}$, tend to infinity.

\begin{assu}\label{AssuCLT}
The following conditions hold in the asymptotic regime.
\begin{enumerate}[label = (\alph*),leftmargin=*]
\item $\Omega^{\paren{t}}\in \cc{C}^{p\times p}_{++}\paren{\alpha_{\min}, M}$ for some fixed, bounded and strictly positive scalars $M$ and $\alpha_{\min}$.
\item $\bar{d}d_{\max}$ grows \emph{slower} than $\sqrt{p}$, i.e., $\limsup_{p\rightarrow\infty}\frac{d_{\max}\bar{d}}{\sqrt{p}} = 0$.
\end{enumerate}
\end{assu}

Next, we present the first main result of this section.

\begin{thm}\label{CLTforT}
Suppose that there is no change-point between $t$ and $t+w$, i.e. $\Omega^{\paren{t}} = \Omega^{\paren{t+1}} = \ldots = \Omega^{\paren{t+w}}$. If $\Omega^{\paren{t}}$ satisfies Assumption \ref{AssuCLT}, then 
\begin{equation*}
T_t \cp{d} \cc{N}\paren{0,1}.
\end{equation*}
\end{thm}

According to Theorem \ref{CLTforT}, $T_t$ converges in distribution to a standard Gaussian random variable under certain asymptotic regularity conditions. Thus, $\boldsymbol{P}_{FA}\paren{\Xi_t}$ is guaranteed to remain below $\pi_0$, if we choose 
\begin{equation*}
\zeta_{\pi_0} = Q_{\pi_0}
\end{equation*}
where $Q_x$ stands for the inverse Gaussian $Q$-function at $x$, i.e. $\int_{Q_x}^{\infty} \paren{2\pi}^{-\frac{1}{2}} \exp\paren{-u^2/2} du = x$.

A close look at $T_t$ in Eq. \eqref{ApproxTt} reveals that $T_t$ is a standardized linear combination of non-Gaussian components $\set{f(Y^{\paren{t,w}}_s):\;s=1,\ldots,p}$. Further, as discussed in Section \ref{DtctAlgo} (see Eq. \eqref{CovY}), the random variables $Y^{\paren{t,w}}_{s_1}$ and $Y^{\paren{t,w}}_{s_2}$ are independent if there is no edge between $s_1$ and $s_2$ at time $t$, i.e. $\Omega^{\paren{t}}_{s_1,s_2} = 0$. Therefore, $T_t$ can be viewed as a linear combination of components in a sparse, stationary non-Gaussian random field. Utilizing \emph{Stein's method} (Lemma $2$ of \cite{bolthausen1982central}) and adopting the proof of Theorem $3.3.1$ of \cite{guyon1995random} to the case of growing $d_{\max}$ and $\bar{d}$, we establish a central limit theorem for $T_t$.

\begin{figure}
\centering
\scalebox{0.7}{\input{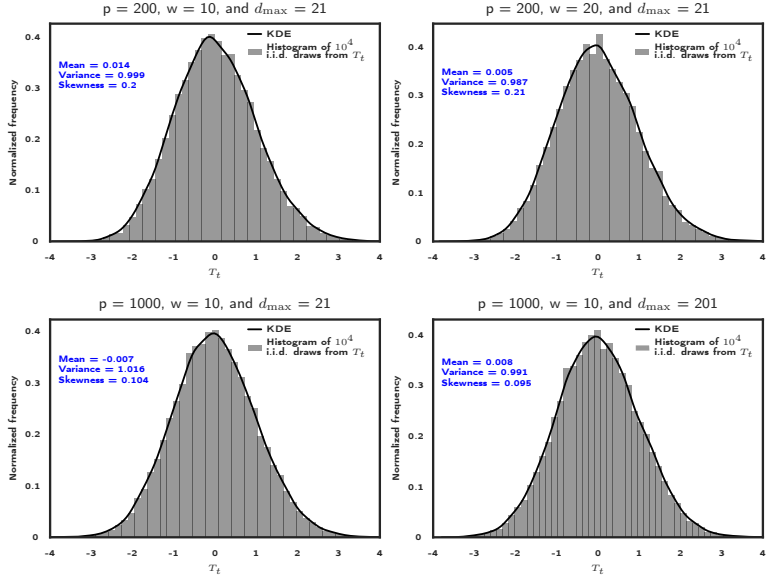}}
\caption{Histogram and kernel density estimator of $T_t$ for different scenarios regarding $\paren{p,d_{\max},w}$. The condition number of $\Omega^{\paren{t}}$ equals to $2.05, 2.02, 2.08,$ and $1.24$ from left to right and top to bottom, respectively.}
\label{Fig:Fig3}
\end{figure}

Figure \ref{Fig:Fig3} depicts the histogram of $T_t$ and its kernel density estimate for different values of $p, w$ and $d_{\max}$ for $10^4$ independent replicates. In each case, $\Omega^{\paren{t}}$ is constructed in the following way: We first generate a matrix $U\in\bb{R}^{p\times p}$ with $d_{\max}$ non-zero entries in each row. Non-zero elements of $U$ are independently generated from a uniform distribution on $\paren{-1,1}$. We then choose the symmetric positive definite $\Omega^{\paren{t}}$ by
\begin{align*}
&\Omega^{\paren{t}} \leftarrow U + U^\top + 1.5d_{\max}I_p,\\
&\Omega^{\paren{t}} \leftarrow \frac{\Omega^{\paren{t}}}{\lambda_{\min}\paren{\Omega^{\paren{t}}}}.
\end{align*}
Thus, in each case $\bar{d} = d_{\max}$ and $\alpha_{\min} = 1$. It is apparent from Figure \ref{Fig:Fig3} that $T_t$ has approximately a standard Gaussian density for all four cases. It is also obvious that the skewness in the histograms decays as $p$ increases.

\begin{rem}
The condition $\paren{b}$ in Assumption \ref{AssuCLT} restricts the growth rate of $\bar{d}d_{\max}$. For instance, it is asymptotically violated for \emph{star network} topologies, i.e. a tree with a single hub node and $p-1$ leaves of length one. One can easily verify that for star networks $d_{\max} = p$ and $\bar{d} = p^{-1}\paren{2\paren{p-1}+p}\approx 3$. Therefore,
\begin{equation*}
\frac{d_{\max}\bar{d}}{\sqrt{p}} \asymp \sqrt{p}\rightarrow \infty.
\end{equation*}
Such graphs are prevalent in centralized sensor or computer networks. Note that $\bar{d} = \cc{O}\paren{1}$ in many applied problems, even for star networks. So roughly speaking, Assumption \ref{AssuCLT} holds whenever $d_{\max} = o\paren{\sqrt{p}}$, as $p\rightarrow\infty$. Namely, $\hat{\Omega}^{\paren{t}}$ can have a finite number of \emph{local hub nodes}, whose degrees grow slower than $\sqrt{p}$. Extensive numerical work presented in Section \ref{SimulSec} aims to understand whether condition $\paren{b}$ in Assumption \ref{AssuCLT} is an artifact of our asymptotic analysis or reflects a shortcoming of the proposed algorithm.
\end{rem}

\comment{\begin{figure}
\centering
\scalebox{0.7}{\input{histKDE_Tt1.pgf}}
\caption{Histogram and kernel density estimator of $T_t$ for different scenarios regarding $\paren{p,d_{\max},w}$. The condition number of $\Omega^{\paren{t}}$ equals to $3.56$ in all plots.}
\label{Fig:Fig4}
\end{figure}}

\subsection{Distribution of $T_t$ under $\bb{H}_{1,t}$}\label{Section3.2}

Next, we study the detection power of $\Xi_t$, under the asymptotic regime of Assumption \ref{AssuCLT}.

\begin{assu}\label{Assum3.2}
The following conditions are imposed on the change-point location and post-change precision matrix.
\begin{enumerate}[label = (\alph*),leftmargin=*]
\item $\Omega^{\paren{t}}\ne \Omega^{\paren{t+1}}$ and $\Omega^{\paren{t+1}} = \ldots, = \Omega^{\paren{t+w}}$.
\item $\Omega^{\paren{t+1}} \in \cc{C}^{p\times p}_{++}\paren{\alpha_{\min}, M}$ for some $M,\alpha_{\min}\in\paren{0,\infty}$.
\end{enumerate}
\end{assu}

For brevity, define $\Delta = \brac{\Delta_s:s=1,\ldots,p}^\top\in\bb{R}^p$ by
\begin{equation}\label{Delta}
\Delta_s \coloneqq \frac{ \brac{\Omega^{\paren{t}}\paren{\Omega^{\paren{t+1}}}^{-1}\Omega^{\paren{t}}}_{ss} }{ \Omega^{\paren{t}}_{ss} } - 1,\quad\forall\;s=1,\ldots,p,
\end{equation}
with $\Delta$ encoding the (relative) amount of sudden change in the network. For instance, if $\Omega^{\paren{t}} = \Omega^{\paren{t+1}}$, then all entries of $\Delta$ are zero and $\LpNorm{\Delta}{2}$ is small if the dependence structure of GGM experiences a weak change at time $t$. Further, define $\bar{\Psi}_p$ by
\begin{equation*}
\bar{\Psi}_p \coloneqq \frac{1}{p}\sum_{s=1}^p f\paren{1+\Delta_s}.
\end{equation*}
Note that $\bar{\Psi}_p$ is well-defined as $\min_{s=1,\ldots,p} \Delta_s > -1$. $\bar{\Psi}_p$ roughly quantifies the average relative change at time $t$. We now present the key result of this section.

\begin{thm}\label{DtctRateT}
Suppose that Assumptions \ref{AssuCLT} and \ref{Assum3.2} hold. Further, assume that $\pi_0,\pi_1\leq \frac{1}{2}$. Consider $\Xi_t$ with the critical value $\zeta_{\pi_0} = Q_{\pi_0}$. Then, $\Xi_t$ satisfies the following condition in the asymptotic regime 
\begin{equation*}
\boldsymbol{P}_{\FA}\paren{\Xi_t}\leq \pi_0,\quad\mbox{and}\quad\boldsymbol{P}_{\MD}\paren{\Xi_t}\leq \pi_1,
\end{equation*}
as long as
\begin{equation}\label{Psi_pBound}
\bar{\Psi}_p\geq \frac{4}{w}\sqrt{\frac{1}{p}\brac{\log\paren{\frac{1}{2\pi_0}} + \log\paren{\frac{1}{2\pi_1}}}}.
\end{equation} 
\end{thm}

\begin{rem}
Theorem \ref{DtctRateT} confirms that detection of changes in the dependence structure of a GGM becomes relatively easier as $p$ grows. The intuition behind Eq. \eqref{Psi_pBound} is that in large GGMs, borrowing information across different edges improves the detection power. Another intuitive aspect of the asymptotic result in Eq. \eqref{Psi_pBound} is the capacity of recognizing weaker change-points for larger delay $w$. Increasing $w$ provides more information about the post-change dependence structure of the GGM. We also draw the reader's attention to the different asymptotic behavior of $p$ and $w$ in Eq. \eqref{Psi_pBound}. Recall the formulation of $\Xi_t$ from Eq. \eqref{Xi_t}. We discussed in Remark \ref{Rem2.2} that the standard deviation of $f(Y^{\paren{t,w}}_s),\;s=1,\ldots,p,$ decays at rate $w^{-1}$ (instead of $w^{-1/2}$), as $w$ grows. Simply put, the proposed convex barrier function $f$ plays a critical role in $T_t$, as it introduces a test statistic with very small variance under both the null and alternative hypotheses. Therefore, the detection capability of $\Xi_t$ rapidly increases with $w$.
\end{rem}

\begin{rem}\label{CPScenarios}
In the offline setting with a single change point located at $t^\star$, it has been shown in \cite{atchade2017scalable,roy2016change} that $t^\star$ can be estimated with order $\log p$ accuracy, i.e. $\abs{\hat{t}-t^\star} = \cc{O}\paren{\log p}$. In the online framework, $w$ (detection delay) plays the role of the estimation error in the change-point location. So in practical scenarios, one can choose $w = \cc{O}\paren{\log p}$. For this case the detection rate \eqref{Psi_pBound} can be rewritten as follows
\begin{equation*}
\bar{\Psi}_p\geq 4\sqrt{\frac{-\log\paren{2\pi_0}-\log\paren{2\pi_1}}{p\log^2 p}}.
\end{equation*}
\end{rem}

Next, we explore different settings in light of Theorem \ref{DtctRateT}..
\begin{enumerate}[label = (\alph*),leftmargin=*]
\item \emph{Uniform change} in $\Omega^{\paren{t}}$: Assume that there is some $\beta\geq 1$ such that $-\Omega^{\paren{t+1}} + \beta^{-1}\Omega^{\paren{t}}$ is strictly positive definite. So the change-point affects all the eigenvalues of $\Omega^{\paren{t}}$. Namely, the network is subject to a dense abrupt change in the spectral domain. In this case for all $s\in\set{1,\ldots,p}$, $\Delta_s$ admits
\begin{equation*}
1+\Delta_s = \frac{ \brac{\Omega^{\paren{t}}\paren{\Omega^{\paren{t+1}}}^{-1}\Omega^{\paren{t}}}_{ss} }{ \Omega^{\paren{t}}_{ss} } \geq \frac{ \brac{\Omega^{\paren{t}}\paren{\beta^{-1}\Omega^{\paren{t}}}^{-1}\Omega^{\paren{t}}}_{ss} }{ \Omega^{\paren{t}}_{ss} } = \beta.
\end{equation*}
Hence $\bar{\Psi}_p \geq f\paren{\beta}$ and so the condition \eqref{Psi_pBound} holds, whenever
\begin{equation*}
f\paren{\beta}\geq 4\sqrt{\frac{-\log\paren{2\pi_0}-\log\paren{2\pi_1}}{pw^2}}.
\end{equation*}
\item \emph{Rank-$r$ change} in $\Omega^{\paren{t}}$: Let $\Omega^{\paren{t}} = U\Lambda U^\top$ be the eigen-decomposition of $\Omega^{\paren{t}}$, i.e., $UU^\top = U^\top U = I_p$ and $\Lambda$ is a diagonal matrix with $\diag\paren{\Lambda} = \brac{\lambda_1,\ldots,\lambda_p}$. For simplicity, also assume that $\diag\paren{\Omega^{\paren{t}}} = \bbM{1}_p$. We assume that only top $r$ eigenvalues of $\Omega^{\paren{t}}$ are impacted by the abrupt change at $t$. Particularly, $\Omega^{\paren{t+1}}$ can be decomposed in the following way.
\begin{equation*}
\Omega^{\paren{t+1}} = U \begin{bmatrix}
\lambda_1\paren{1+\beta_1} & & \\
& \ddots & \\
& & \lambda_p\paren{1+\beta_p}
\end{bmatrix} U^\top\quad \suchthat\quad\beta_{r+1} = \ldots = \beta_p = 0.
\end{equation*}
We also assume that $\min_{i=1,\ldots,r}\beta_i\geq \beta_{\min}$ for some strictly positive $\beta_{\min}$. Roughly speaking $\beta_{\min}$ can not be so large as $\Omega^{\paren{t+1}}\in\cc{C}^{p\times p}_{++}\paren{\alpha_{\min}, M}$. Due to the  convexity of $f$, $\bar{\Psi}_p$ can be controlled from below by $f\paren{\sum_{s=1}^p \paren{1+\Delta_s}/p}$. Lastly, we obtain a sharp upper bound on $\sum_{s=1}^p \paren{1+\Delta_s}$. Observe that
\begin{eqnarray*}
\frac{1}{p}\sum_{s=1}^p \paren{1+\Delta_s} &=& \sum_{s=1}^p \frac{ \brac{\Omega^{\paren{t}}\paren{\Omega^{\paren{t+1}}}^{-1}\Omega^{\paren{t}}}_{ss} }{p \Omega^{\paren{t}}_{ss} } = \frac{1}{p}\tr\paren{ \Omega^{\paren{t}}\paren{\Omega^{\paren{t+1}}}^{-1}\Omega^{\paren{t}} } = \sum_{s=1}^p \frac{\lambda^2_s}{p\lambda_s \paren{1+\beta_s} }\\
&=& \sum_{s=1}^p \frac{\lambda_s}{p \paren{1+\beta_s} }  = \frac{1}{p}\sum_{s=1}^p \lambda_s - \frac{1}{p}\sum_{s=1}^r \frac{\lambda_s\beta_s}{1+\beta_s} \RelNum{\paren{a}}{=} 1 - \frac{1}{p}\sum_{s=1}^r \frac{\lambda_s\beta_s}{1+\beta_s} \\
&\leq& 1 - \frac{\beta_{\min}}{1+\beta_{\min}} \frac{\sum_{s=1}^r \lambda_s}{p}.
\end{eqnarray*}
Notice that identity $\paren{a}$ is implied from the fact that all diagonal entries of $\Omega^{\paren{t}}$ equal to one. So,
\begin{equation*}
\bar{\Psi}_p \geq f\paren{\sum_{s=1}^p \frac{1+\Delta_s}{p}} \geq f\paren{1 - \frac{\beta_{\min}}{1+\beta_{\min}} \frac{\sum_{s=1}^r \lambda_s}{p}}.
\end{equation*}
In summary, the condition \eqref{Psi_pBound} holds, as long as
\begin{equation*}
f\paren{1 - \frac{\beta_{\min}}{1+\beta_{\min}} \frac{\sum_{s=1}^r \lambda_s}{p}}\geq 4\sqrt{\frac{-\log\paren{2\pi_0}-\log\paren{2\pi_1}}{pw^2}}.
\end{equation*}
It is noteworthy that for small enough $\beta_{\min}$, $\frac{\beta_{\min}}{1+\beta_{\min}} \frac{\sum_{s=1}^r \lambda_s}{p}$ has the same asymptotic behavior as $ \frac{\beta_{\min}r}{p}$.
\item \emph{Small relative change}: In this case, $\Omega^{\paren{t+1}} = \Omega^{\paren{t}} + \Theta$ for a positive semi-definite matrix $\Theta$ satisfying
\begin{equation*}
\OpNorm{\paren{\Omega^{\paren{t}}}^{-\frac{1}{2}}\Theta\paren{\Omega^{\paren{t}}}^{-\frac{1}{2}}}{2}{2} \leq \xi < 1,
\end{equation*}
for a $\xi\in\brapar{0,1}$. One can verify the following fact for any positive semi-definite $A$ with $\OpNorm{A}{2}{2} \leq \xi$.
\begin{equation*}
\paren{I-\paren{\xi+1}^{-1}A} - \paren{I_p+A}^{-1} \in S^{p\times p}_{++}.
\end{equation*}
Using this fact we control $1+\Delta_s$ from above, $\forall\;s=1,\ldots, p$. Choose $A= \paren{\Omega^{\paren{t}}}^{-\frac{1}{2}}\Theta\paren{\Omega^{\paren{t}}}^{-\frac{1}{2}}$, then
\begin{eqnarray*}
1+\Delta_s &=& \frac{ \brac{\Omega^{\paren{t}}\paren{\Omega^{\paren{t}} + \Theta}^{-1}\Omega^{\paren{t}}}_{ss} }{\Omega^{\paren{t}}_{ss} } = \frac{ \brac{\paren{\Omega^{\paren{t}}}^{\frac{1}{2}}\paren{I_p+A}^{-1}\paren{\Omega^{\paren{t}}}^{\frac{1}{2}}}_{ss} }{\Omega^{\paren{t}}_{ss} } \leq 1 - \frac{ \brac{\paren{\Omega^{\paren{t}}}^{\frac{1}{2}}A\paren{\Omega^{\paren{t}}}^{\frac{1}{2}}}_{ss} }{\paren{1+\xi}\Omega^{\paren{t}}_{ss} }\\
&=&1-\frac{\Theta_{ss}}{\paren{1+\xi}\Omega^{\paren{t}}_{ss}} < 1.
\end{eqnarray*}
Since $f$ is a decreasing, convex function at $\parbra{0,1}$, we have
\begin{equation*}
\bar{\Psi}_p = \frac{1}{p} \sum_{s=1}^p f\paren{1+\Delta_s} \geq f\paren{\sum_{s=1}^p \frac{1+\Delta_s}{p}} \geq f\paren{1 - \frac{1}{\paren{1+\xi}p}\sum^p_{s=1} \frac{\Theta_{ss}}{\Omega^{\paren{t}}_{ss}} }.
\end{equation*}
Therefore $\Xi_t$ has the desirable properties, if
\begin{equation*}
f\paren{1 - \frac{1}{\paren{1+\xi}p}\sum^p_{s=1} \frac{\Theta_{ss}}{\Omega^{\paren{t}}_{ss}} }\geq 4\sqrt{\frac{-\log\paren{2\pi_0}-\log\paren{2\pi_1}}{pw^2}}.
\end{equation*}
Lastly note that $p^{-1}\sum^p_{s=1} \frac{\Theta_{ss}}{\Omega^{\paren{t}}_{ss}}$ encodes the mean relative change in the conditional variance of all nodes.
\item \emph{Localized change}: In this example, we consider a setting that the change-point only affects a single node and the edges connected to it. Let $\Theta \coloneqq \Omega^{\paren{t+1}} -\Omega^{\paren{t}}$ and suppose that $\diag\paren{\Omega^{\paren{t}}} = \bbM{1}_p$. The abrupt change being trapped in some $s\in\set{1,\ldots,p}$ and its neighbors means that
\begin{equation*}
\Theta = ve^\top_s + e_sv^\top = \brac{v \; e_s}\begin{bmatrix}
e^\top_s\\
v^\top\\
\end{bmatrix},
\end{equation*}
for $s$-th coordinate unit vector $e_s\in\bb{R}^p$ and a $v\in\bb{R}^p$ such that for any $u\in\set{1,\ldots,p}$, $v_u = 0$ if $\Omega^{\paren{t}}_{su} = 0$. For simplicity define $\beta\coloneqq \LpNorm{v}{2}$, $U \coloneqq \brac{v \; e_s}\in\bb{R}^{p\times 2}$, and $V \coloneqq \brac{e_s \; v} \in\bb{R}^{p\times 2}$. As parts $\paren{a}-\paren{c}$, our goal is to study the asymptotic behavior of $\bar{\Psi}_p$ for growing $p$. We only consider the case that $\beta = \cc{O}\paren{1}$ when $p\rightarrow\infty$. Using \emph{Woodbury}'s matrix identity yields
\begin{equation*}
\Delta_u =  \frac{ \brac{\Omega^{\paren{t}}\paren{\Omega^{\paren{t}} +UV^\top}^{-1}\Omega^{\paren{t}}}_{uu} }{\Omega^{\paren{t}}_{uu} } - 1= \brac{U \paren{I_2+V^\top\paren{\Omega^{\paren{t}}}^{-1}U}^{-1}V^\top}_{uu} ,\quad\forall\;u=1,\ldots,p.
\end{equation*}
Observe that if $\Omega^{\paren{t}}_{su} = 0$, then the $u$-th row of $U$ and $V$ are filled with zeros. So $\Delta_u = 0$ and $f\paren{1+\Delta_u} = 0$. On the other hand for neighbors of $s$, we have
\begin{equation*}
\abs{\Delta_u} \leq \frac{\LpNorm{U_{u,:}}{2}\LpNorm{V_{u,:}}{2}}{\lambda_{\min} \paren{I_2+V^\top\paren{\Omega^{\paren{t}}}^{-1}U}} = \frac{\LpNorm{U_{u,:}}{2}^2}{\lambda_{\min} \paren{I_2+V^\top\paren{\Omega^{\paren{t}}}^{-1}U}}
\end{equation*}
Since the smallest eigenvalue of $\Omega^{\paren{t+1}} = \Omega^{\paren{t}} + UV^\top$ is greater than $\alpha_{\min}>0$ and its largest eigenvalue is less than $M$, one can find a bounded positive scalar $C_{\min}$ depending on $M$ and $\alpha_{\min}$, such that
\begin{equation*}
\lambda_{\min} \paren{I_2+V^\top\paren{\Omega^{\paren{t}}}^{-1}U}\geq \sqrt{C_{\min}} \quad\Longrightarrow \quad \abs{\Delta_u} \leq \frac{\LpNorm{U_{u,:}}{2}^2}{\sqrt{C_{\min}}}.
\end{equation*}
Thus, as $p\rightarrow\infty$, all $\Delta_u$ terms remain in a small neighborhood of zero, since $\LpNorm{U}{2}^2 = 1+\beta^2 = \cc{O}\paren{1}$. By applying a Second order \emph{Taylor} expansion of $f$ near one, we get
\begin{equation*}
f\paren{1+\Delta_u} \leq C_{\max}\Delta^2_u,\quad\forall\;u=1,\ldots,p,
\end{equation*}
for some bounded positive scalar $C_{\max}$. Thus, some straightforward algebraic derivations lead to
\begin{equation}\label{Psi_pRate}
\bar{\Psi}_p  = \frac{1}{p} \sum^p_{u=1} f\paren{1+\Delta_u} \leq\frac{C_{\max}}{p}\sum^p_{u=1} \Delta^2_u \leq \frac{C_{\max}}{pC_{\min}}\sum^p_{s=1} \LpNorm{U_{u,:}}{2}^4 \asymp \frac{1}{p} \sum^p_{s=1} \LpNorm{U_{u,:}}{2}^4 \leq \frac{\LpNorm{U}{2}^4}{p} \asymp \frac{1}{p}
\end{equation}
Comparing Eq. \eqref{Psi_pRate} with the sufficient detectability condition in Eq. \eqref{Psi_pBound} reveals that the considered localized change \emph{can not be detected} with small error.
\end{enumerate} 

We conclude this section by relaxing condition $\paren{b}$ in Assumption \ref{Assum3.2}. Choose $w_+\in\bb{N}$ strictly smaller than $w$. The following result introduces a sufficient condition on $\bar{\Psi}_p$ for detecting the change point at time $t_{-}\coloneqq t-\paren{w-w_+}$. Notice that the next $w$ samples at time $t_{-}$ satisfy 
\begin{equation}\label{RelaxCond}
\Omega^{\paren{t_{-}}} = \ldots = \Omega^{\paren{t}} \ne \Omega^{\paren{t+1}} = \ldots = \Omega^{\paren{t+w_+}}.
\end{equation}
In words, we observe $w_+$ post-change samples, which is equivalent to having a detection delay of $w_+$.

\begin{thm}\label{DtctRateTRelaxed}
Suppose that Assumption \ref{AssuCLT}, constraint $\paren{a}$ in Assumption \ref{Assum3.2}, and condition \eqref{RelaxCond} hold. Choose $\pi_0,\pi_1\leq \frac{1}{2}$ and consider $\Xi_{t_{-}}$ with the critical value $\zeta_{\pi_0} = Q_{\pi_0}$. Then, asymptotically
\begin{equation*}
\boldsymbol{P}_{\FA}\paren{\Xi_{t_{-}}}\leq \pi_0,\quad\mbox{and}\quad\boldsymbol{P}_{\MD}\paren{\Xi_{t_{-}}}\leq \pi_1,
\end{equation*}
whenever
\begin{equation*}
\bar{\Psi}_p\geq \frac{4}{w_+}\sqrt{\frac{1}{p}\brac{\log\paren{\frac{1}{2\pi_0}} + \log\paren{\frac{1}{2\pi_1}}}}.
\end{equation*} 
\end{thm}

Due to space considerations, the proof of Theorem \ref{DtctRateTRelaxed} is omitted since it follows along the lines of the proof of Theorem \ref{DtctRateT}, with slightly more involved calculations. 

\section{Asymptotic analysis of $\Xi_t$: unknown $\Omega^{\paren{t}}$}\label{Section4}

As discussed in Section \ref{Section2.3}, the pre-change precision matrix needs to be estimated from the observed samples. Our developed theoretical approach for studying $\hat{\Xi}_t$ (and its associated test statistic $\hat{T}_t$) relies on a large-sample sharp bound on the estimation error of the precision matrix. Specifically, we need operator and \emph{Frobenius} norm convergence rates in terms of $p, d_{\max}$ of the network, and sample size. Because of space constraints, the non-asymptotic analysis of estimating sparse precision matrices is beyond the scope of this paper. Consequently, we only employ off-the-shelf theoretical results for this task. For instance, we heavily use Theorems $1$ and $4$ of \cite{cai2011constrained} for studying Algorithm \ref{Algo1} with the CLIME estimator. 

In Section \ref{Section2.3}, we qualitatively stated that separation of consecutive change-points is needed to guarantee the consistency of Algorithm \ref{Algo1}. The following condition formalizes this notion.

\begin{assu}\label{Assu4.1}
There exists a bounded large enough scalar $C$ (depending on $\alpha_{\min}$ and $M$) such that
\begin{equation*}
\abs{t^{\star}_j-t^\star_{j-1}} \geq N\coloneqq Cpd_{\max}\log^2 p \paren{w\vee \log p}, \quad\forall\;j\geq 1.
\end{equation*}
\end{assu}

Next, we establish the null distribution of $\hat{T}_t$ in Algorithm \ref{Algo1} (with the CLIME estimator). 

\begin{thm}\label{CLTforTHat}
Suppose that there is no change-point between $t-N$ and $t+w$, i.e. $\Omega^{\paren{t-N+1}} =\ldots = \Omega^{\paren{t}} = \ldots = \Omega^{\paren{t+w}}$. Note that the condition is viable when Assumption \ref{Assu4.1} holds. Further, suppose that Assumption \ref{AssuCLT} holds for  $\Omega^{\paren{t}}$; then, asymptotically
\begin{equation*}
\hat{T}_t \cp{d} \cc{N}\paren{0,1},
\end{equation*}
\end{thm}

\begin{figure}
\centering
\scalebox{0.7}{\input{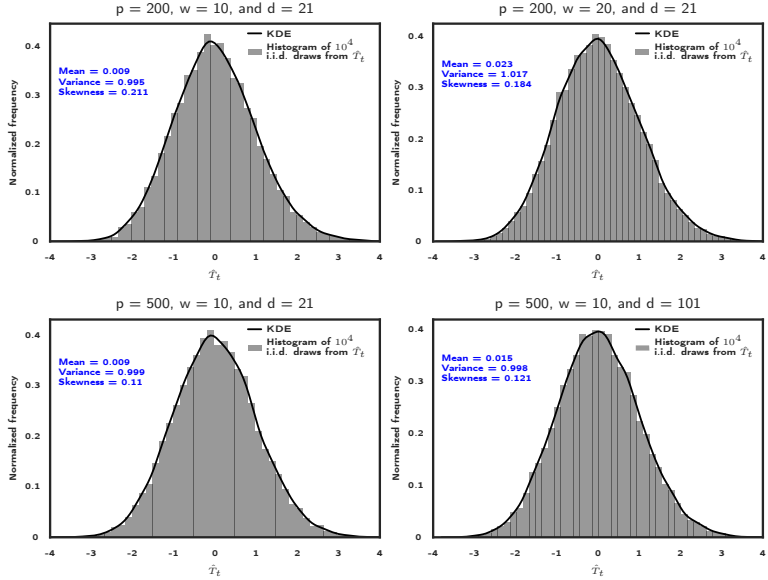}} 
\caption{Histogram and kernel density estimator of $\hat{T}_t$ for different scenarios regarding $\paren{p,d,w}$. In each panel, $\Omega^{\paren{t}}$ is estimated by the graphical LASSO with $N = \lceil pd_{\max}\log p\rceil$. The regularization parameters are chosen as $0.01,0.01,0.01$, and $0.005$ from left to right and top to bottom, respectively.}
\label{Fig:Fig5}
\end{figure}

According to Theorem \ref{CLTforTHat}, if Assumption \ref{Assu4.1} holds, then the null distribution of $\hat{T}_t$ has the same asymptotic behavior as that of $T_t$ (with a fully-known pre-change precision matrix). Roughly speaking, the null distribution of $T_t$ remains almost intact, given large enough separation of consecutive change-point locations. Hence, if the critical value $\zeta_{\pi_0}$ is chosen by $Q_{\pi_0}$, then $\boldsymbol{P}_{\FA}(\hat{\Xi}_t)$ is ensured to be less than $\pi_0$, in the asymptotic regime. We conduct a simple simulation study for verifying Theorem \ref{CLTforTHat}. Figure \ref{Fig:Fig5} depicts the histogram and associated kernel density estimate of $\hat{T}_t$ for four scenarios of $\paren{p,w,d_{\max}}$. In each panel, the histogram summarizes $10^4$ independent replicates based on the same pre-change true and estimated precision matrices. For constructing each histogram, we use the graphical LASSO with $N = \lceil pd_{\max}\log p\rceil$ for estimating the pre-change precision matrix. The \emph{scikit-learn} implementation of the graphical lasso with the coordinate descent optimization algorithm is utilized for estimating $\Omega^{\paren{t}}$. Finally, note that the regularization parameters are chosen as $0.01,0.01,0.01$, and $0.005$ from left to right and top to bottom, respectively.

\begin{rem}
The CLIME estimator requires $\cc{O}\paren{pd_{\max}\log p}$ samples for consistently estimating $\Omega^{\paren{t}}$, in the Frobenius norm (see Theorem $4$ of \cite{cai2011constrained}), which is slightly weaker than the sufficient condition on the interval between two successive change-point in Assumption \ref{Assu4.1}. Moreover, according to Theorem $8$ of Atchad\'{e} et al. \cite{atchade2017scalable}, estimating the location of change point with order $\log p$ error in a $s$-sparse GGM requires $\cc{O}\paren{s\log p}$ samples. Setting $w = \log p$ in Assumption \ref{Assu4.1} leads to a slightly stronger condition on $N$ in the online framework. In particular, we require $N = pd_{\max}\log^3 p$ observations before the change-point.
\end{rem}

\begin{rem}
Notice that $N$ is an increasing function of $w$ in Assumption \ref{Assu4.1}, which may seem counter-intuitive at first glance. This observation can be spelled out by noting that the $\var\paren{T_t}$ decay as order $w^{-2}$, under the null hypothesis. Simply put as $w$ increases, even a slight bias introduced by the CLIME estimator can change the null distribution of $T_t$ (because of its very small variance). In contrast, Theorem \ref{DtctRateT} suggests that increasing $w$ improves the power of our proposed test. Therefore, a suitable value for $w$ is affected by the trade-off between the false alarm and  miss-detection rates of $\hat{\Xi}_t$. Hence, we posit that selecting $w = \cc{O}\paren{\log p}$ is a reasonable choice in practical settings.
\end{rem}

Next, we study the detection rate of $\hat{\Xi}_t$ under Assumption \ref{Assu4.1}. The following result shows that under the same conditions of Theorem \ref{DtctRateT} and if Assumption \ref{Assu4.1} holds, then $\hat{\Xi}_t$ enjoys the same detection power as $\Xi_t$.

\begin{thm}
Suppose that Assumptions \ref{AssuCLT} and \ref{Assum3.2} hold. Further, assume that $\pi_0,\pi_1\leq \frac{1}{2}$. Consider $\hat{\Xi}_t$ with the critical value $\zeta_{\pi_0} = Q_{\pi_0}$. Then, $\hat{\Xi}_t$ satisfies the following condition  asymptotically  
\begin{equation*}
\boldsymbol{P}_{\FA}\paren{\hat{\Xi}_t}\leq \pi_0,\quad\mbox{and}\quad\boldsymbol{P}_{\MD}\paren{\hat{\Xi}_t}\leq \pi_1,
\end{equation*}
provided that Assumption \ref{Assu4.1} is satisfied and
\begin{equation*}
\bar{\Psi}_p\geq \frac{4}{w}\sqrt{\frac{1}{p}\brac{\log\paren{\frac{1}{2\pi_0}} + \log\paren{\frac{1}{2\pi_1}}}}.
\end{equation*} 
\end{thm}

In summary, given a good estimate of $\Omega^{\paren{t}}$, for large networks the false alarm and miss-detection rates of the proposed test are not affected by the estimation error. Namely, our detection algorithm is robust against small estimation error, which is highly desirable in applications. It is worth mentioning that under Assumption \ref{Assu4.1}, the detection power of $\Xi_{t-}$ (recall its formulation from Eq. \eqref{RelaxCond} and Theorem \ref{DtctRateTRelaxed} ) is not also affected by the estimation error of the pre-change precision matrix.

\section{Numerical studies}\label{SimulSec}

The next set of numerical experiments aims to:
\begin{enumerate}[label = (\alph*),leftmargin=*]
\item Corroborate the developed asymptotic theory for $\Xi_t$ and $\hat{\Xi}_t$ in Sections \ref{Section3} and \ref{Section4}.
\item Understand the impact of issues absent in the asymptotic analysis, such as the length of the burn-in period and update window, on $\boldsymbol{P}_{\FA}$ and $\boldsymbol{P}_{\MD}$ of Algorithm \ref{Algo1}.
\item Assess the capabilities of our proposed method on real-world applications through sequential detection of structural changes in the S\&P $500$ over the period $2000-2016$.
\end{enumerate}

In Section \ref{Sec5.1}, we evaluate the detection power of $\Xi_t$ (known pre-change precision matrix) for the three different scenarios introduced in Section \ref{Section3.2}: uniform, rank-$r$ and localized abrupt changes. Section \ref{Sec5.2} gauges the performance of $\hat{\Xi}_t$ in Algorithm \ref{Algo1}. Lastly, Section \ref{Sec5.3} is devoted to experiments with S\&P $500$ data. Throughout this section, the false alarm rate is set to $\pi_0 = 0.01$ and thus the critical value is $\zeta_{\pi_0} = 2.3263$. Further, $\set{G_t \sim \cc{N}\paren{\zero_p, \Omega^{\paren{t}}}:t=1,\ldots,T}$ denotes a time-evolving zero-mean GGM of size $p$, observed on a time horizon of size $T$. The $\Omega^{\paren{t}}$ is always normalized to be a correlation matrix. In other words, $\Omega^{\paren{t}} = R^{\paren{t}}$, where $R^{\paren{t}}$ denotes the associated partial correlation matrix to $\Omega^{\paren{t}}$.  

\subsection{Performance evaluation of $\Xi_t$}\label{Sec5.1}

The time horizon is set to $T = 500$ and a single change point occurs at $t^\star = 250$. We use $\Omega_{\BC}$ and $\Omega_{\AC}$ for referring the before and after-change precision matrix of $G_t$, respectively. Namely
\begin{equation*}
\Omega^{\paren{t}} = \Omega_{\BC} \bbM{1}_{\set{t\leq t^\star}} + \Omega_{\AC} \bbM{1}_{\set{t> t^\star}},\quad\forall\;t \in\set{1,\ldots,T}.
\end{equation*} 

\begin{rem}\label{Rem5.1}
Given $p$ and $d$, we generate a sparse random matrix $U\in\bb{R}^{p\times p}$ with exactly $d$ independent standard Gaussian entries per row. The locations of non-zero elements in each row are selected uniformly at random (without replacement). We then generate $H\in\bb{R}^{p\times p}$ via
\begin{equation*}
H \coloneqq \frac{UU^\top}{\LpNorm{UU^\top}{\infty}}
\end{equation*}
where $H$ is a non-negative definite symmetric matrix whose entries lie in $\brac{-1,1}$. Note that  when $d\ll p$, distinct rows of $U$ have non-overlapping support with high probability. This fact ensures the sparsity of $H$ in our synthetic simulation studies. In practice $H$ is near singular when $\frac{d}{p}\ll 1$. So for controlling the condition number of $\Omega_{\BC}$, we choose
\begin{equation*}
\Omega_{\BC} \longleftarrow H + \lambda_0 I_{p},
\end{equation*}
for some properly chosen $\lambda_0 > 0$.
\end{rem}

The first experiment assesses the performance of $\Xi_t$ for detecting a uniform change across the entire network. In this setting, we select $\lambda_0 = 0.1$ and $\Omega_{\AC} = \paren{1+\beta} \Omega_{\BC}$. Note that $p\beta$ encodes the strength of change-point in the whole network. Our objective is to evaluate the sensitivity of false alarm and miss-detection to $p, \beta, d$, and delay $w$. For any fixed tuple $\paren{p,w,\beta,d}$, $\pi_0$ and $\pi_1$ are estimated based on $100$ replicates of the network. In particular for the $i$-th replication, false alarm and miss-detection rates are respectively estimated by 
\begin{equation}\label{EStimErrors}
\hat{\pi}_{0,i} = \frac{1}{t^\star-w}\sum_{t=1}^{t^\star-w} \Xi^i_t,\quad\mbox{and}\quad \hat{\pi}_{1,i} = \frac{1}{T-w-t^\star}\sum_{t=t^\star}^{T-w} \paren{1-\Xi^i_t},
\end{equation}
where $\set{\Xi^i_t}^T_{i=1}$ represents the sequence of binary decisions in $\set{1,\ldots,T}$. The final estimates $\hat{\pi}_0$ and $\hat{\pi}_1$ are successively obtained by computing the sample mean of $\set{\hat{\pi}_{0,i}}^{100}_{i=1}$ and $\set{\hat{\pi}_{1,i}}^{100}_{i=1}$. 

\begin{figure}
	\centering
	\scalebox{0.68}{\input{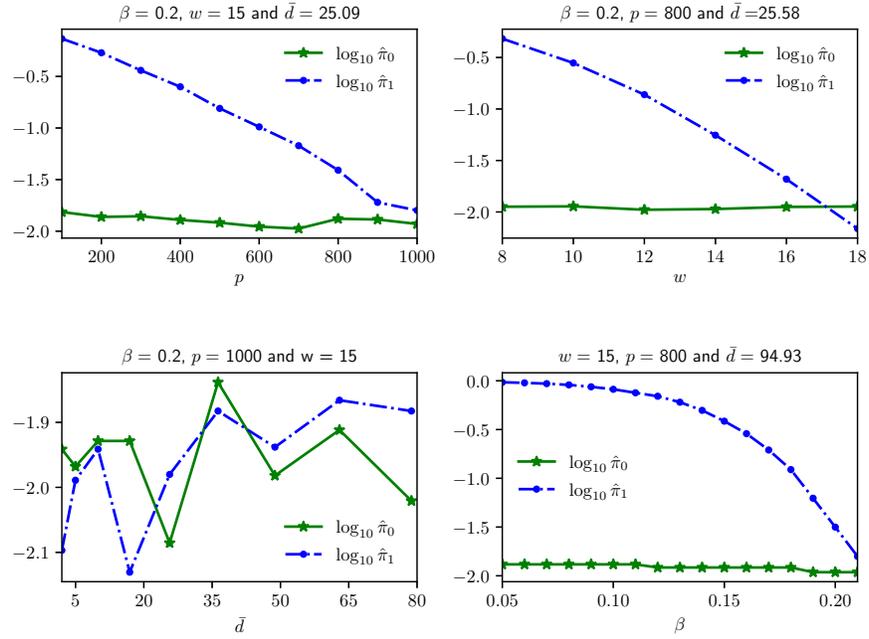}} 
	\caption{$\log_{10} \hat{\pi}_0$ and $\log_{10} \hat{\pi}_1$ for $\Xi_t$ (fully known pre-change attributes) in a uniform change scenario. The four plots present $\hat{\pi}_0$ and $\hat{\pi}_1$ for different $p, w, \bar{d},$ and $\beta$, respectively from left to right and top to bottom.}
	\label{Fig:Fig6}
\end{figure}

Figure \ref{Fig:Fig6} depicts $\log_{10} \hat{\pi}_0$ and $\log_{10} \hat{\pi}_1$ as a function of $p, w, \beta,$ and $\bar{d}$. It is apparent from Figure \ref{Fig:Fig6} that $\hat{\pi}_0$ lies in a small neighborhood of $\pi_0 = 0.01$ in all scenarios. However, $\hat{\pi}_1$ rapidly decays with an increase in signal-to-noise-ratio (SNR), $w$ and network size $p$, which is in accordance with Theorem \ref{DtctRateT}. For instance, as can be seen in the left-down panel in Figure \ref{Fig:Fig6}, when $p = 800$ and $w=15$, the average miss-detection probability is around $1\%$. On the other hand, when the change point uniformly affects all nodes, for fixed values of $\beta,p,$ and $w$, the detection power does not clearly depend on the sparsity structure of the network (which is encoded by $\bar{d}$), which is again an expected finding based on part $\paren{a}$ of Remark \ref{CPScenarios}.

\begin{figure}
	\centering
	\scalebox{0.68}{\input{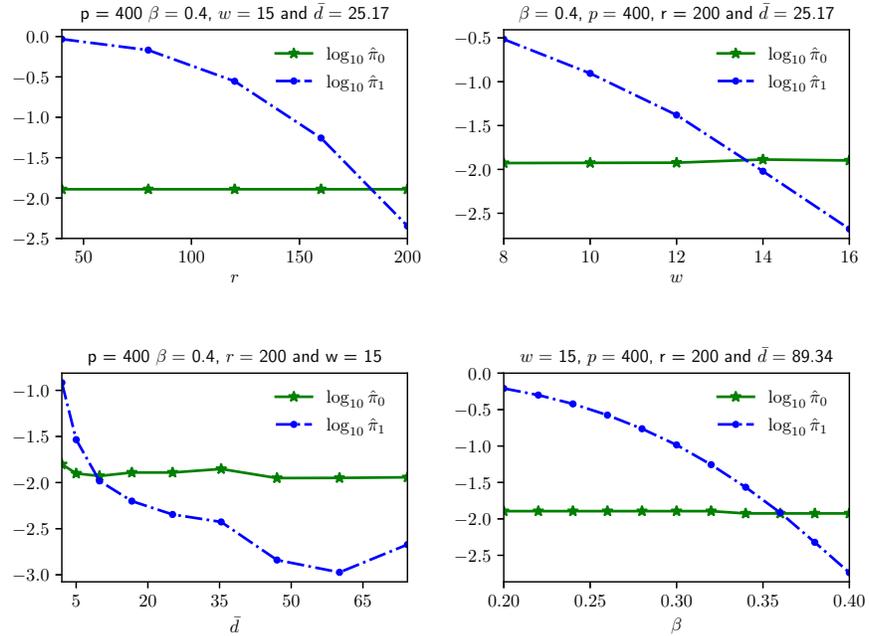}} 
	\caption{$\log_{10} \hat{\pi}_0$ and $\log_{10} \hat{\pi}_1$ for $\Xi_t$ (fully known pre-change attributes) in low-rank change regime. The four plots present $\hat{\pi}_0$ and $\hat{\pi}_1$ to variations in $r, w, \bar{d},$ and $\beta$, respectively from left to right and top to bottom.}
	\label{Fig:Fig7}
\end{figure}

In the next experiment, we focus on a low-rank sudden change in the spectrum of $\set{G_t}^T_{t=1}$. Consider the spectral representation of $\Omega_{\BC}$ given by
\begin{equation*}
\Omega_{\BC} = \sum_{i=1}^p \lambda_{i,\BC} v_iv^\top_i,\quad\mbox{where}\quad \InnerProd{v_{j_1}}{v_{j_2}} = \bbM{1}_{\set{j_1= j_2}},\forall\;j_1,j_2\in\set{1,\ldots,p}.
\end{equation*}
Here we assume that $\set{\lambda_{i,\BC}:\;i=1,\ldots,p}$ has a non-increasing order. In this setting, the top $r$ eigenvalues of $\Omega_{\BC}$ are affected by abrupt change, without any impact on the eigenvectors. Particularly, we choose $\Omega_{\AC}$ in the following way:
\begin{equation*}
\Omega_{\AC} = \sum_{i=1}^r \lambda_{i,\BC}\paren{1+\beta} v_iv^\top_i + \sum_{i=r+1}^p \lambda_{i,\BC}\paren{1+\beta} v_iv^\top_i.
\end{equation*}
The dependency of $\hat{\pi}_0$ and $\hat{\pi}_1$ (over $100$ independent replicates) on $\beta, r, w,$ and $\bar{d}$ has been presented in Figure \ref{Fig:Fig7}. Analogous to the uniform change framework, the miss-detection rate rapidly decays with $\beta, r,$ and $w$. For example, with window size $w = 15$, $\Xi_t$ is capable of detecting $20\%$ change ($\beta = 0.2$) in half of the eigenvalues with $99\%$ percent accuracy ($\hat{\pi}_1 \approx 0.01$). Further, based on the bottom-left panel in Figure \ref{Fig:Fig7}, the proposed algorithm is more powerful for denser graphs (large average degree). 

\begin{figure}
\centering
\scalebox{0.68}{\input{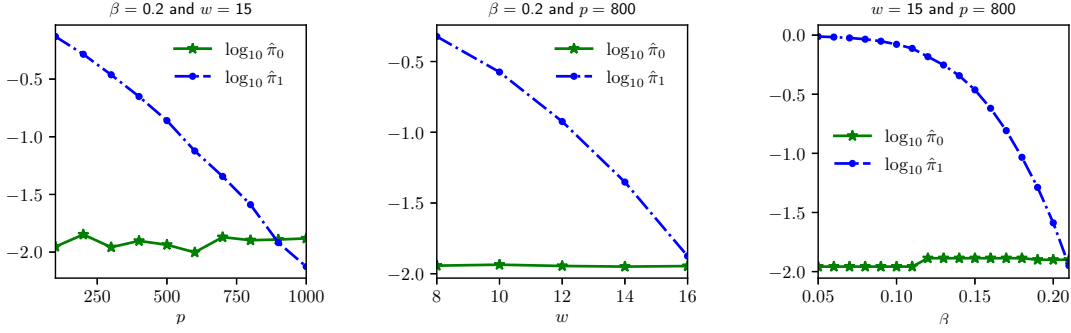}} 
\caption{$\log_{10} \hat{\pi}_0$ and $\log_{10} \hat{\pi}_1$ for $\Xi_t$ (fully known pre-change attributes) for uniform change in star graph. From left to right, $\hat{\pi}_0$ and $\hat{\pi}_1$ are plotted against $p, w,$ and $\beta$, respectively.}
\label{Fig:Fig7Prime}
\end{figure}

In the sequel, we appraise the detection performance of $\Xi_t$ for uniform change in star graphs. Recall that the developed asymptotic theory in Section \ref{Section3} can not be extended to star graphs, as $d_{\max}$ grows faster than $\sqrt{p}$. For modeling star graphs, the following scheme is used for constructing $\Omega_{\BC}$. Let $u = \brac{0, u_1,\ldots,u_{p-1}}^\top\in\bb{R}^p$ be a standard Gaussian vector padded with a zero in its first position and also define $e_1 = \brac{1,0,\ldots,0}^\top\in\bb{R}^p$. The pre-change precision matrix is given by
\begin{equation*}
\Omega_{\BC} = 1.1 I_p + \frac{e_1u^\top + u e^\top_1}{p\LpNorm{u}{\infty}}.
\end{equation*}
Note that $\Omega_{\BC}$ encodes a star network whose root is set as the first node ($s=1$). Despite lack of asymptotic analysis, Figure \ref{Fig:Fig7Prime} shows that the false alarm rate is still around $0.01$, particularly for large graphs. Moreover, the variation of $\hat{\pi}_1$ in terms of $w, p,$ and $\beta$ is analogous to that in preceding simulation studies.

\comment{\begin{table}
\centering
\makebox[0pt][c]{\parbox{\textwidth}{%
\begin{minipage}[b]{\hsize}\centering
\begin{tabular}{|c|c|c|c|c|c|c|c|c|c|c|}
\hline
$p$ & $\boldsymbol{100}$ & $\boldsymbol{200}$ & $\boldsymbol{300}$ & $\boldsymbol{400}$ & $\boldsymbol{500}$ & $\boldsymbol{600}$ & $\boldsymbol{700}$ & $\boldsymbol{800}$ & $\boldsymbol{900}$ & $\boldsymbol{1000}$\\ \hline
$\bb{E}100\hat{\pi}_0$ & $1.59$ & $1.42$ & $1.01$ & $1.27$ & $1.14$ & $1.45$ & $1.05$ & $1.18$ & $1.14$ & $1.34$\\ \hline
$\sd\paren{100\hat{\pi}_0}$ & $1.51$ & $1.77$ & $1.13$ & $1.37$ & $1.31$ & $1.70$ & $1.28$ & $1.52$ & $1.38$ & $1.28$\\ \hline
$\bb{E}100\hat{\pi}_1$ & $74.69$ & $53.56$ & $36.43$ & $24.02$ & $14.06$ & $9.52$ & $4.97$ & $2.92$ & $1.74$ & $0.78$\\ \hline
$\sd\paren{100\hat{\pi}_1}$ & $7.48$ & $9.14$ & $8.69$ & $8.39$ & $5.13$ & $5.18$ & $3.27$ & $2.39$ & $1.92$ & $1.25$\\ \hline
\end{tabular}
\renewcommand{\thetable}{\arabic{table}-(a)}
\caption{Summary of upper-left plot in Figure \ref{Fig:Fig6} }
\label{tab:1a}
\end{minipage}
\vspace{0mm}

\hfill
\begin{minipage}[b]{\hsize}\centering
\begin{tabular}{|c|c|c|c|c|c|c|c|}
\hline
$w$ & $\boldsymbol{8}$ & $\boldsymbol{10}$ & $\boldsymbol{12}$ & $\boldsymbol{14}$ & $\boldsymbol{16}$ & $\boldsymbol{18}$ & $\boldsymbol{20}$ \\ \hline
$\bb{E}100\hat{\pi}_0$ & $1.09$ & $1.14$ & $1.20$ & $1.31$ & $1.29$ & $1.25$ & $1.29$ \\ \hline
$\sd\paren{100\hat{\pi}_0}$ & $0.96$ & $1.00$ & $1.21$ & $1.39$ & $1.49$ & $1.58$ & $1.76$ \\ \hline
$\bb{E}100\hat{\pi}_1$ & $48.45$ & $27.58$ & $13.24$ & $5.70$ & $2.00$ & $0.65$ & $0.13$ \\ \hline
$\sd\paren{100\hat{\pi}_1}$ & $7.63$ & $7.27$ & $5.98$ & $4.03$ & $2.34$ & $1.36$ & $0.49$ \\ \hline
\end{tabular}
\addtocounter{table}{-1}
\renewcommand{\thetable}{\arabic{table}-(b)}
\caption{Summary of upper-right plot in Figure \ref{Fig:Fig6} }
\label{tab:1b}
\end{minipage}
\vspace{0mm}

\hfill
\begin{minipage}[b]{\hsize}\centering
\begin{tabular}{|c|c|c|c|c|c|c|c|c|c|}
\hline
$\bar{d}$ & $\boldsymbol{1.99}$ & $\boldsymbol{4.99}$ & $\boldsymbol{9.95}$ & $\boldsymbol{16.85}$ & $\boldsymbol{25.68}$ & $\boldsymbol{36.35}$ & $\boldsymbol{48.81}$ & $\boldsymbol{62.88}$ & $\boldsymbol{78.73}$ \\ \hline
$\bb{E}100\hat{\pi}_0$ & $1.27$ & $1.31$ & $1.22$ & $1.11$ & $1.20$ & $1.36$ & $1.05$ & $1.11$ & $1.36$ \\ \hline
$\sd\paren{100\hat{\pi}_0}$ & $1.56$ & $1.53$ & $1.20$ & $1.12$ & $1.57$ & $1.53$ & $1.19$ & $1.35$ & $1.42$ \\ \hline
$\bb{E}100\hat{\pi}_1$ & $0.86$ & $1.01$ & $0.97$ & $1.11$ & $0.69$ & $1.07$ & $0.92$ & $1.14$ & $0.82$ \\ \hline
$\sd\paren{100\hat{\pi}_1}$ & $1.25$ & $1.25$ & $1.46$ & $1.54$ & $1.06$ & $1.60$ & $1.64$ & $1.51$ & $1.28$ \\ \hline
\end{tabular}
\addtocounter{table}{-1}
\renewcommand{\thetable}{\arabic{table}-(c)}
\caption{Summary of bottom-left plot in Figure \ref{Fig:Fig6} }
\label{tab:1c}
\end{minipage}
\vspace{0mm}

\hfill
\begin{minipage}[b]{\hsize}\centering
\begin{tabular}{|c|c|c|c|c|c|c|c|c|c|c|c|c|}
\hline
$\beta$ & $\boldsymbol{0.1}$ & $\boldsymbol{0.11}$ & $\boldsymbol{0.12}$ & $\boldsymbol{0.13}$ & $\boldsymbol{0.14}$ & $\boldsymbol{0.15}$ & $\boldsymbol{0.16}$ & $\boldsymbol{0.17}$ & $\boldsymbol{0.18}$ & $0.19$ & $0.2$ & $0.21$ \\ \hline
$\bb{E}100\hat{\pi}_0$ & $1.00$ & $1.00$ & $0.91$ & $0.91$ & $0.91$ & $0.91$ & $0.91$ & $0.91$ & $0.91$ & $1.09$ & $1.09$ & $1.09$ \\ \hline
$\sd\paren{100\hat{\pi}_0}$ & $1.27$ & $1.27$ & $1.17$ & $1.17$ & $1.17$ & $1.17$ & $1.17$ & $1.17$ & $1.17$ & $1.31$ & $1.31$ & $1.31$ \\ \hline
$\bb{E}100\hat{\pi}_1$ & $83.21$ & $76.70$ & $67.69$ & $58.46$ & $48.09$ & $37.08$ & $26.92$ & $17.69$ & $10.94$ & $6.43$ & $3.31$ & $1.82$ \\ \hline
$\sd\paren{100\hat{\pi}_1}$ & $5.84$ & $6.87$ & $8.76$ & $9.51$ & $9.48$ & $9.44$ & $8.45$ & $6.91$ & $5.25$ & $4.61$ & $3.23$ & $2.33$ \\ \hline
\end{tabular}
\addtocounter{table}{-1}
\renewcommand{\thetable}{\arabic{table}-(d)}
\caption{Summary of bottom-right plot in Figure \ref{Fig:Fig6} }
\label{tab:1d}
\end{minipage}%
}}
\end{table}}

\comment{\begin{table}
\centering
\makebox[0pt][c]{\parbox{\textwidth}{%
\begin{minipage}[b]{\hsize}\centering
\begin{tabular}{|c|c|c|c|c|c|}
\hline
$r$ & $\boldsymbol{40}$ & $\boldsymbol{80}$ & $\boldsymbol{120}$ & $\boldsymbol{160}$ & $\boldsymbol{200}$ \\ \hline
$\bb{E}100\hat{\pi}_0$ & $1.23$ & $1.23$ & $1.23$ & $1.23$ & $1.23$ \\ \hline
$\sd\paren{100\hat{\pi}_0}$ & $1.36$ & $1.36$ & $1.36$ & $1.36$ & $1.36$ \\ \hline
$\bb{E}100\hat{\pi}_1$ & $96.00$ & $81.12$ & $46.78$ & $13.35$ & $1.59$ \\ \hline
$\sd\paren{100\hat{\pi}_1}$ & $3.13$ & $6.49$ & $8.47$ & $5.42$ & $1.64$ \\ \hline
\end{tabular}
\renewcommand{\thetable}{\arabic{table}-(a)}
\caption{Summary of upper-left plot in Figure \ref{Fig:Fig7} }
\label{tab:2a}
\end{minipage}
\vspace{0mm}
			
\hfill
\begin{minipage}[b]{\hsize}\centering
\begin{tabular}{|c|c|c|c|c|c|c|}
\hline
$w$ & $\boldsymbol{8}$ & $\boldsymbol{10}$ & $\boldsymbol{12}$ & $\boldsymbol{14}$ & $\boldsymbol{16}$ & $\boldsymbol{18}$ \\ \hline
$\bb{E}100\hat{\pi}_0$ & $1.39$ & $1.44$ & $1.46$ & $1.25$ & $1.27$ & $1.27$ \\ \hline
$\sd\paren{100\hat{\pi}_0}$ & $1.23$ & $1.32$ & $1.39$ & $1.30$ & $1.56$ & $1.58$ \\ \hline
$\bb{E}100\hat{\pi}_1$ & $42.63$ & $22.47$ & $9.19$ & $2.95$ & $0.76$ & $0.15$ \\ \hline
$\sd\paren{100\hat{\pi}_1}$ & $5.92$ & $5.63$ & $4.14$ & $2.35$ & $1.13$ & $0.37$ \\ \hline
\end{tabular}
\addtocounter{table}{-1}
\renewcommand{\thetable}{\arabic{table}-(b)}
\caption{Summary of upper-right plot in Figure \ref{Fig:Fig7} }
\label{tab:2b}
\end{minipage}
\vspace{0mm}
			
\hfill
\begin{minipage}[b]{\hsize}\centering
\begin{tabular}{|c|c|c|c|c|c|c|c|c|c|}
\hline
$\bar{d}$ & $\boldsymbol{2.00}$ & $\boldsymbol{4.97}$ & $\boldsymbol{9.88}$ & $\boldsymbol{16.66}$ & $\boldsymbol{25.13}$ & $\boldsymbol{35.36}$ & $\boldsymbol{46.98}$ & $\boldsymbol{60.10}$ & $\boldsymbol{74.12}$ \\ \hline
$\bb{E}100\hat{\pi}_0$ & $0.99$ & $1.28$ & $1.26$ & $1.25$ & $1.23$ & $1.19$ & $1.07$ & $1.43$ & $1.34$ \\ \hline
$\sd\paren{100\hat{\pi}_0}$ & $1.19$ & $1.40$ & $1.53$ & $1.71$ & $1.36$ & $1.55$ & $1.26$ & $1.49$ & $1.50$ \\ \hline
$\bb{E}100\hat{\pi}_1$ & $16.54$ & $4.26$ & $2.72$ & $1.98$ & $1.59$ & $1.39$ & $1.33$ & $1.06$ & $1.45$ \\ \hline
$\sd\paren{100\hat{\pi}_1}$ & $6.50$ & $3.24$ & $2.43$ & $2.32$ & $1.65$ & $1.65$ & $1.72$ & $1.40$ & $1.73$ \\ \hline
\end{tabular}
\addtocounter{table}{-1}
\renewcommand{\thetable}{\arabic{table}-(c)}
\caption{Summary of bottom-left plot in Figure \ref{Fig:Fig7} }
\label{tab:2c}
\end{minipage}
\vspace{0mm}

\hfill
\begin{minipage}[b]{\hsize}\centering
\begin{tabular}{|c|c|c|c|c|c|c|c|c|c|c|c|}
\hline
$\beta$ & $\boldsymbol{0.2}$ & $\boldsymbol{0.22}$ & $\boldsymbol{0.24}$ & $\boldsymbol{0.26}$ & $\boldsymbol{0.28}$ & $\boldsymbol{0.30}$ & $\boldsymbol{0.32}$ & $\boldsymbol{0.34}$ & $\boldsymbol{0.36}$ & $\boldsymbol{0.38}$ & $\boldsymbol{0.4}$ \\ \hline
$\bb{E}100\hat{\pi}_0$ & $1.31$ & $1.31$ & $1.31$ & $1.31$ & $1.31$ & $1.31$ & $1.31$ & $1.20$ & $1.20$ & $1.20$ & $1.20$ \\ \hline
$\sd\paren{100\hat{\pi}_0}$ & $1.51$ & $1.51$ & $1.51$ & $1.51$ & $1.51$ & $1.51$ & $1.51$ & $1.36$ & $1.36$ & $1.36$ & $1.36$ \\ \hline
$\bb{E}100\hat{\pi}_1$ & $73.54$ & $64.62$ & $54.39$ & $43.32$ & $32.31$ & $22.78$ & $15.18$ & $9.55$ & $5.60$ & $2.97$ & $1.34$ \\ \hline
$\sd\paren{100\hat{\pi}_1}$ & $8.25$ & $9.06$ & $9.85$ & $9.79$ & $9.25$ & $7.93$ & $6.35$ & $4.54$ & $3.24$ & $2.35$ & $1.56$ \\ \hline
\end{tabular}
\addtocounter{table}{-1}
\renewcommand{\thetable}{\arabic{table}-(d)}
\caption{Summary of bottom-right plot in Figure \ref{Fig:Fig7} }
\label{tab:2d}
\end{minipage}%
}}
\end{table}}

\subsection{Sensitivity of $\hat{\Xi}_t$ to the burn-in and update periods}\label{Sec5.2}

Next, we numerically examine the sensitivity of $\hat{\Xi}_t$ to three choice parameters: burn-in period, precision matrix update frequency, and the frequency of updating the regularization parameter (model selection). The burn-in period $n_0$ refers to the number of samples used for computing an initial estimate of the post-change precision matrix (with the premise that the first change-point occurs at $t=0$). We assume that the distance between two consecutive change-points is greater than $n_0$. Since increasing $n_0$ reduces the bias and variance of the estimated precision matrix, it can provide an effective barrier against random fluctuations in $\hat{\Xi}_t$ time series and reduces the false alarm rate. On the other  hand, an unnecessary increase in $n_0$ can affect the applicability of the proposed detection algorithm in real world scenarios.

False alarms are costly as the algorithm needs to wait for a new burn-in window which in turn may lead to missing a true forthcoming sudden change. Strictly speaking, suppose that $n_0 = 500$ and the algorithm raises a false alarm at $t = 1000$, while the true change-point is located at $t = 1250$. Thus, detecting this sudden change with a delay less than $250$ is infeasible. One practical solution that guards against false alarms, which comes at the price of slightly increasing the detection delay, is to declare an abrupt change at time $t$, whenever
\begin{equation*}
\hat{\Xi}_{t+r}\geq \zeta_{\pi_0},\quad \forall\;r=0,\ldots,\iota-1.
\end{equation*}
In other words, we suggest to wait for $\iota$ successive flags before entering a new burn-in period. Our synthetic simulation studies show that $\iota = 5$ is a proper choice and so throughout this section we fix $\iota = 5$.

Another choice parameter, which has direct impact on the computational complexity of the online detection algorithm, is how often to update the estimated pre-change precision matrix. Let $B$ denote a pre-specified block size of recent observations. After the burn-in period and until identifying a new change-point, the precision matrix is updated once every $B$ new samples. In our simulation studies, we exploit the QUIC method (see Algorithm 1 in \cite{hsieh2014quic}) for the estimating pre-change precision matrix because of its smaller error and faster convergence comparing to the graphical lasso. The QUIC algorithm takes advantage of the second order approximation of the Gaussian log-likelihood for solving the following optimization problem.
\begin{equation*}
\hat{\Omega}_{\tau} = \min_{\Omega\in S^{p\times p}_{++}} \set{ -\log\det \Omega + \InnerProd{\Omega}{\frac{1}{n}\sum_{i=1}^{n} Z_iZ^\top_i } + \tau\LpNorm{\Omega}{1}},
\end{equation*}
in which $Z_i$ are i.i.d. draws from a zero mean Gaussian vector with precision matrix $\Omega^\star$. Proper choice of $\tau$ is essential for controlling the sparsity of the solution and avoiding over-fitting. The optimal value of $\tau$ is chosen on a grid of size $20$ to minimize the Bayesian information criterion (BIC) score. In particular, when $\hat{t}_{last}$ stands for the last detected abrupt change before $t$ and $n \coloneqq t-\hat{t}_{last}$, then we optimize the BIC score over the following grid (for finding the best $i\in\set{0,\ldots,19}$).
\begin{equation*}
\cc{G}\coloneqq \set{ 10^{\paren{-1+j/10}}\sqrt{\frac{\log p}{n}}:\;j=0,\ldots,19 }.
\end{equation*} 
Namely, we assume that $\tau = \tau_0 \sqrt{n^{-1}\log p}$ as $n$ increases and the BIC procedure aims to choose best $\tau_0$. Note that the relationship between $\tau$ and $n$ is justified by our asymptotic understanding of the QUIC algorithm. Evaluating the BIC score over $\cc{G}$ is a heavy computational burden for large graphs. On the other hand, when $B$ is considerably smaller than $p$, tuning the optimal $j$ for each update cycle may provide a negligible improvement in the detection accuracy. So for accelerating the whole procedure, we propose to conduct a BIC model selection once every $\kappa$ times of updating the precision matrix. Thus, we update the optimal value of $j$ after getting $\kappa B$ new samples, where $\kappa^{-1}$ refers to the frequency of conducting BIC model selection.

We now have all the required ingredients for describing our numerical experiments. Our objective is to assess the sensitivity of Algorithm \ref{Algo1} to $n_0, B,$ and $\kappa$. In all experiments, we choose $p = 100, \pi_0 = 0.01$ and $w = 20$. Moreover three abrupt changes occur at $\set{3000,6000,9000}$ in $T = 10^4$ samples. For each choice of parameters $\paren{n_0,B, \kappa}$, we employ Algorithm \ref{Algo1} on $50$ replications of the non-stationary GGM generated from the following procedure. In order to have a fair comparison, all experiments are conducted on the same $50$ replications. We denote by $\Omega_0, \Omega_1, \Omega_2,$ and $\Omega_3$ the precision matrix of the network in the following four periods $\set{1,\ldots,2999}$, $\set{3000,\ldots,5999}$, $\set{6000,\ldots,8999}$, and $\set{9000,\ldots,10000}$. In each replicate, $\Omega_0$ is independently generated with the same procedure as in Remark \ref{Rem5.1} with $p = 100, d = 20$ and $\lambda_0 = 0.1$. The first change point uniformly affects all nodes with $\beta = 0.2$, i.e. $\Omega_1 = \paren{1+\beta}\Omega_0$. The second abrupt change only impacts the top $r = 50$ eigenvalues without changing the eigenvectors. Specifically,
\begin{equation*}
\Omega_2 = \sum_{i=1}^r \lambda_i\beta v_iv^\top_i + \Omega_0,\quad \mbox{with}\quad \beta = 0.4,
\end{equation*}
in which $\set{v_i\in\bb{R}^p}^p_{i=1}$ are orthonormal eigenvectors of $\Omega_1$ whose associated eigenvalues are sorted in a non-increasing order. For the last change point, we randomly generate a new precision matrix with the same distribution as $\Omega_0$. In this case possibly all entries, eigenvalues, and eigenvectors are affected by the change-point. Roughly speaking the sudden change signal is more visible compared to the previous ones. So for each replicate, we have access to a multivariate Gaussian time series of length $T = 10^4$ of  $p=100$ vertices.

\begin{figure}
\centering
\scalebox{0.33}{\input{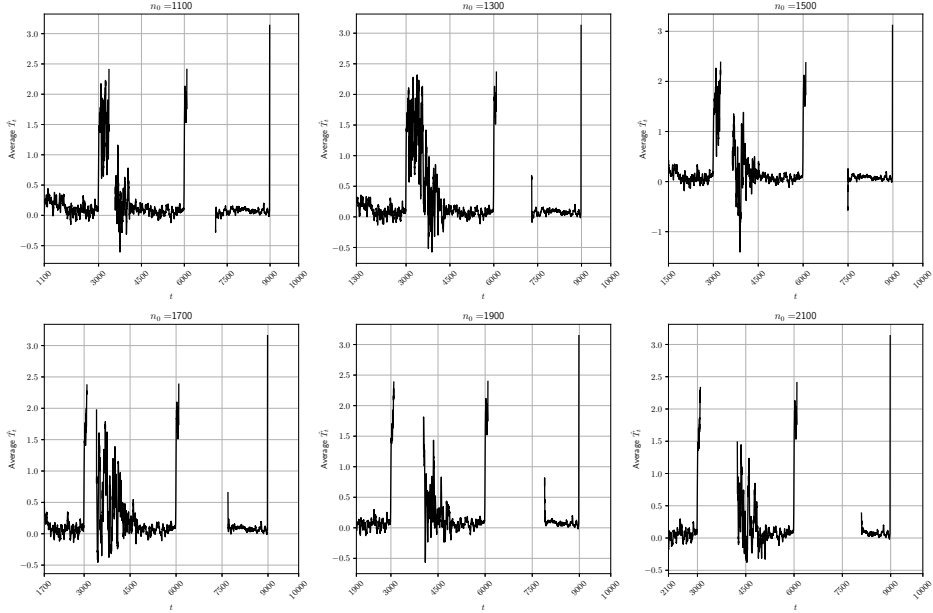}} 
\caption{Average sample path of $\hat{T}_t$ for different values of $n_0$.}
\label{Fig:Fig8}
\end{figure}

The first simulation study aims to assess the sensitivity of Algorithm \ref{Algo1} to the burn-in period $n_0$. We set $B = 50$ and $\kappa = 4$ for all experiments, while varying $n_0$ in the set $\set{1100,1300,1500,1700,1900,2100}$. Note that in all the following experiments, $\hat{T}_t$ is filled with \emph{NA} if $t$ is in the burn-in period. Table \ref{tab:table1} summarizes the median and interquartile range (IQR) of the detection delays for the three aforementioned type of change points, as well as average number of false alarms. According to Table \ref{tab:table1} given enough samples in the burn-in period for estimating pre-change precision matrix, increasing $n_0$ has negligible impact on the detection delay. However, larger values of $n_0$ can guard $\hat{T}_t$ against false alarms. Moreover Figure \ref{Fig:Fig8} exhibits the average sample path of $\hat{T}_t$ time series (over $50$ experiments and after skipping the missing values) for each choice of $n_0$. Figure \ref{Fig:Fig8} shows that the change points randomly affecting all the nodes are easier to detect than uniform or low rank breaks, which corroborates the summary results in Table \ref{tab:table1}.

\vspace{3mm}
\begin{center} 
\begin{adjustbox}{width=0.95\columnwidth}
\begin{tabular}{cc|c|c|c|c|c|c|}
\cline{3-8} 
& & $n_0 = 1100$ & $n_0= 1300$ & $n_0= 1500$ & $n_0= 1700$ & $n_0=1900$ & $n_0=2100$ \\ \cline{1-8}
\multicolumn{1}{|c|}{\multirow{2}{*}{Uniform Change}}  & 
\multicolumn{1}{c|}{Median delay} & $53.5$ & $51$ & $54$ & $54$ & $52.5$ & $51$ \\ \cline{2-8} 
\multicolumn{1}{|c|}{} &
\multicolumn{1}{c|}{IQR of delay} & $48.75$ & $48.75$ & $52.25$ & $51.25$ & $48.75$ & $48.75$  \\ \cline{1-8} 
\multicolumn{1}{|c|}{\multirow{2}{*}{Low rank Change}}  & 
\multicolumn{1}{c|}{Median delay} & $33$ & $32.5$ & $32$ & $32.5$ & $33$ & $33$  \\ \cline{2-8}
\multicolumn{1}{|c|}{} &
\multicolumn{1}{c|}{IQR of delay} & $13.25$ & $14.5$ & $12$ & $11.75$ & $13.25$ & $13.25$  \\ \cline{1-8} 
\multicolumn{1}{|c|}{\multirow{2}{*}{Random Change}}  & 
\multicolumn{1}{c|}{Median delay} & $4$ & $4$ & $4$ & $4$ & $4$ & $4$  \\ \cline{2-8}
\multicolumn{1}{|c|}{} &
\multicolumn{1}{c|}{IQR of delay} & $0.75$ & $0.75$ & $0.75$ & $0.75$ & $0.75$ & $0.75$  \\ 
\cline{1-8} 
\multicolumn{2}{|c|}{Average number of false alarms} & $0.12$ & $0.06$ & $0.08$ & $0.04$ & $0.00$ & $0.00$  \\ 
\cline{1-8} 
\end{tabular}
\end{adjustbox}
\captionof{table}{The median and IQR of detection delay, and average number of false alarams for different values of $n_0$ over $50$ independent experiments.}
\label{tab:table1}
\end{center} 

In the second experiment, we evaluate the robustness of false alarm and miss detection rates to variations of $\kappa$. Recall that for a fixed mini-batch size $B$ and given $n$ samples in the pre-change regime, the precision matrix is estimated using the QUIC algorithm with regularization parameter $\tau_0\sqrt{n^{-1}\log p}$, where $\tau_0$ is updated after getting $\kappa B$ new samples. We fix $B = 50$, and choose $\paren{n_0,\kappa}$ pair from $\set{1500,2000}\times \set{1,2,3,4}$ ($8$ different cases). The median and IQR of detection delay, as well as average number of false alarms (computed from $50$ independent experiment) are given in Table \ref{tab:table2} for each $\paren{n_0,\kappa}$. It can be seen that variations in $\kappa$ barely affects the detection delay and false alarm rate.

\vspace{3mm}
\begin{center} 
\begin{adjustbox}{width=0.95\columnwidth}
\begin{tabular}{cc|c|c|c|c|c|c|c|c|}
\cline{3-10} 
& & $\paren{1500,1}$ & $\paren{1500,2}$ & $\paren{1500,3}$ & $\paren{1500,4}$ & $\paren{2000,1}$ & $\paren{2000,2}$ & $\paren{2000,3}$ & $\paren{2000,4}$ \\ \cline{1-10}
\multicolumn{1}{|c|}{\multirow{2}{*}{Uniform Change}}  & 
\multicolumn{1}{c|}{Median delay} & $54$ & $54$ & $54$ & $54$ & $52.5$ & $52.5$ & $52.5$ & $52.5$\\ \cline{2-10} 
\multicolumn{1}{|c|}{} &
\multicolumn{1}{c|}{IQR of delay} & $52.25$ & $52.25$ & $52.5$ & $52.25$ & $48$ & $48$ & $48.5$ & $48$\\ \cline{1-10} 
\multicolumn{1}{|c|}{\multirow{2}{*}{Low rank Change}}  & 
\multicolumn{1}{c|}{Median delay} & $33$ & $33$ & $33$ & $32$ & $33$ & $33$ & $32$ & $31$ \\ \cline{2-10}
\multicolumn{1}{|c|}{} &
\multicolumn{1}{c|}{IQR of delay} & $13.5$ & $13.25$ & $13.25$ & $12$ & $13.5$ & $13.25$ & $13.25$ & $12$  \\ \cline{1-10} 
\multicolumn{1}{|c|}{\multirow{2}{*}{Random Change}}  & 
\multicolumn{1}{c|}{Median delay} & $4$ & $4$ & $4$ & $4$ & $4$ & $4$ & $4$ & $4$ \\ \cline{2-10}
\multicolumn{1}{|c|}{} &
\multicolumn{1}{c|}{IQR of delay} & $0$ & $0.75$ & $0.75$ & $0.75$ & $0.75$ & $1$  & $1$ & $1$\\ 
\cline{1-10} 
\multicolumn{2}{|c|}{Average number of false alarms} & $0.08$ & $0.08$ & $0.08$ & $0.08$ & $0.00$ & $0.00$  & $0.00$ & $0.00$\\ 
\cline{1-10} 
\end{tabular}
\end{adjustbox}
\captionof{table}{The median and IQR of detection delay, and average number of false alarams for different values of $\paren{n_0, \kappa}$ over $50$ independent experiments.}
\label{tab:table2}
\end{center} 

Lastly, we study the role of increasing the min-batch size $B$ on the performance of $\hat{T}_t$. Note that we use the same data set as in the two previous simulation studies. We fix $\kappa = 4$ for all the experiments and consider $8$ different scenarios for $\paren{n_0, B}$ in  $\set{1100,1500}\times \set{5,10,20,40}$. Table \ref{tab:table3} summarizes the median and IQR of detection delay, and average number of false alarms. Similar to Table \ref{tab:table2}, examining the columns of Table \ref{tab:table3} shows that if provided with an adequate quality initial estimate of the precision matrix in the burn-in period, increasing $B$ exhibits a small impact on the detection performance of $\hat{T}_t$. Roughly speaking, when $B$ is small, the estimated precision matrix (and consequently the detection procedure) slightly changes after an update.

\vspace{3mm}
\begin{center} 
\begin{adjustbox}{width=0.95\columnwidth}
\begin{tabular}{cc|c|c|c|c|c|c|c|c|}
\cline{3-10} 
& & $\paren{1100,5}$ & $\paren{1100,10}$ & $\paren{1100,20}$ & $\paren{1100,40}$ & $\paren{1500,5}$ & $\paren{1500,10}$ & $\paren{1500,20}$ & $\paren{1500,40}$ \\ \cline{1-10}
\multicolumn{1}{|c|}{\multirow{2}{*}{Uniform Change}}  & 
\multicolumn{1}{c|}{Median delay} & $54.5$ & $54.5$ & $53.5$ & $52.5$ & $54.5$ & $54.5$ & $54.5$ & $54.5$\\ \cline{2-10} 
\multicolumn{1}{|c|}{} &
\multicolumn{1}{c|}{IQR of delay} & $50.5$ & $50.5$ & $50.75$ & $48.75$ & $52.75$ & $52.75$ & $52.25$ & $52.25$\\ \cline{1-10} 
\multicolumn{1}{|c|}{\multirow{2}{*}{Low rank Change}}  & 
\multicolumn{1}{c|}{Median delay} & $33$ & $33$ & $32.5$ & $32.5$ & $33$ & $33$ & $32.5$ & $32.5$ \\ \cline{2-10}
\multicolumn{1}{|c|}{} &
\multicolumn{1}{c|}{IQR of delay} & $13.5$ & $16.75$ & $16$ & $13.25$ & $16.75$ & $16.75$ & $16$ & $13.25$  \\ \cline{1-10} 
\multicolumn{1}{|c|}{\multirow{2}{*}{Random Change}}  & 
\multicolumn{1}{c|}{Median delay} & $4$ & $4$ & $4$ & $4$ & $4$ & $4$ & $4$ & $4$ \\ \cline{2-10}
\multicolumn{1}{|c|}{} &
\multicolumn{1}{c|}{IQR of delay} & $0$ & $0$ & $0$ & $0.75$ & $0.75$ & $0$  & $0$ & $0.75$\\ 
\cline{1-10} 
\multicolumn{2}{|c|}{Average number of false alarms} & $0.12$ & $0.12$ & $0.12$ & $0.14$ &  $0.08$ & $0.08$ & $0.08$ & $0.08$\\ 
\cline{1-10} 
\end{tabular}
\end{adjustbox}
\captionof{table}{The median and IQR of detection delay, and average number of false alarams for different values of $\paren{n_0, B}$ over $50$ independent experiments.}
\label{tab:table3}
\end{center} 

\subsection{Real data experiment}\label{Sec5.3}

We assess the performance of $\hat{\Xi}_t$ in a real-world scenario. The objective is to estimate the time location of abrupt changes in the dependency structure of S\&P 500 daily close price data for the period from 2000-01-01 to 2016-03-03 (total of 3814 trading days). Note that the S\&P 500 data do not fit into the formulation of time evolving GGMs with independent observations for two main reasons. First, the list of securities in S\&P 500 pool evolves over time. For example \emph{Alphabet Inc. Class C} (with ticker symbol \emph{GOOG}) entered the list on 2006-04-03. For circumventing the first issue, we follow the cleaning procedure from \cite{atchade2017scalable} and select a fixed list of 436 ticker symbols from 2004-02-06 to 2016-03-03, consisting of 3039 trading days. The second technical challenge is that the daily close price of each ticker typically exhibits strong temporal dependence with non-Gaussian marginal distribution. The geometric Brownian motion (GBM) is a versatile and popular tool for modeling daily stock prices in mathematical finance (see e.g., Chapter 10 of \cite{ross2014introduction}). Note that under the GBM model, the daily log-return time series can be well approximated by a Gaussian random walk. In other words, a high-dimensional GGM with 436 vertices provides a good working model for the daily log-returns of 
S\&P 500 components.

\begin{figure}
\centering
\scalebox{0.35}{\input{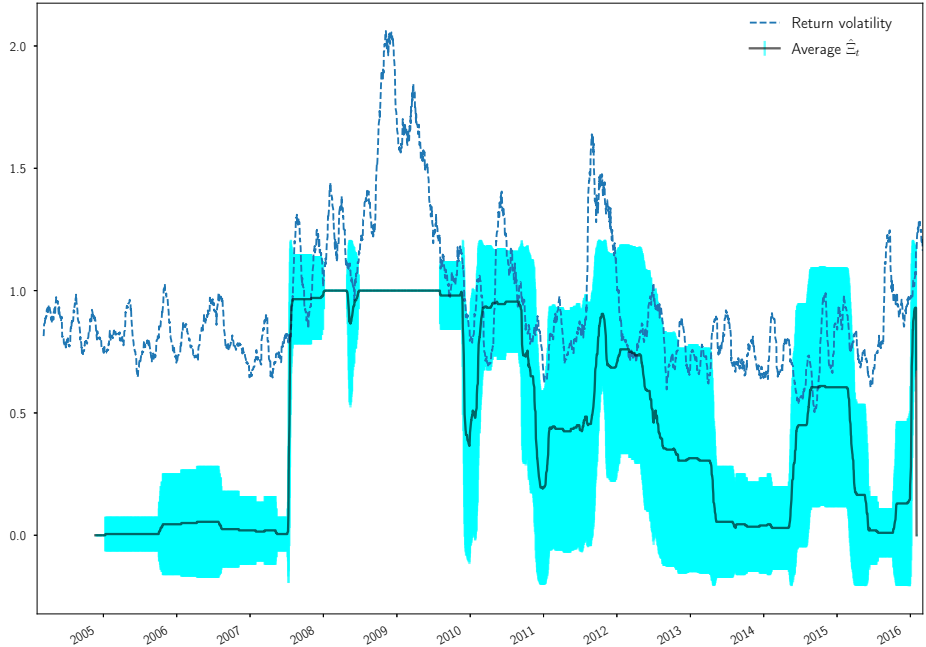}} 
\caption{Confidence bound around $\hat{\Xi}_t$ for $300$ random subsets of S\&P 500 with $100$ stocks, and average return volatility.}
\label{Fig:Fig10}
\end{figure}

We choose $n_0 = 200$, $w = 22$ (corresponding to the number of trading days in a month), $\pi_0 = 0.05$, $\kappa = 2$, and $B = 10$. We also employ the same estimation and model selection approach as in Section \ref{Sec5.2}. For each experiment, we conduct our sequential detection algorithm on a network of 100 randomly chosen tickers, i.e., $p = 100$. We also set $\hat{\Xi}_t = 1$ for $t$ in the burn-in period. The solid black curve and cyan interval in Figure \ref{Fig:Fig10}, respectively exhibits the sample average and standard deviation of $\hat{\Xi}_t$ over $300$ independent experiments. Note that for each $t$, the proximity of the average $\hat{\Xi}_t$ (over $300$ experiments) to one implies that $t$ is close to a sudden change in most of the experiments. So roughly speaking the solid dark curve captures the strength of nearest change-point. 

It becomes apparent from Figure \ref{Fig:Fig10}  that the first cluster of change-points starts in the last few months of $2008$ and lasts for one and half years, which successfully captures the 17-month bear market period from October $2007$ to March $2009$ (when the S\&P 500 index lost approximately 50\% of its value). The beginning of the \emph{Great Recession} in December $2007$ and the bankruptcy filing of \emph{Lehman Brothers Holdings Inc.} on September 15, 2008 are two notable events during this bear market. Another visible change-point in Figure \ref{Fig:Fig10} takes place in late $2010$, which can represent the end of bear market in $2009$. Our detection algorithm find the third change-point starting around September $2011$, in most of $300$ experiments. The new regime likely corresponds to a sharp drop in stock prices during August 2011 affecting the performance of stock exchanges across the United States, Europe, East Asia, and Middle East. The fall of stock market in August $2011$ was due to fears of contagion of the European sovereign debt crisis to Spain and Italy. The proposed algorithm spots the last change-point in Fall $2014$. Note that from September $18$ to October $14$ of $2014$, the S\&P 500 experienced a remarkable decline of $7.4\%$ due to market jitters over the rapid spread of the \emph{Ebola} virus beyond Africa, fears of a global economic slowdown, Hong Kong protests (\emph{umbrella revolution}), and first U.S. strikes against ISIS in Syria.

In the sequel, we introduce a tangible interpretation of the change-points identified in terms of the average volatility of all tickers (nodes in the S\&P 500 graphical model). For each node $s$, let $\set{X_{t,s}}^T_{t=1}$ stand for the daily log-return of $s$. We approximate the volatility of $\set{X_{t,s}}^T_{t=1}$ using
\begin{equation*}
V_{t,s} = \sd\paren{X_{s,t}, \ldots, X_{s,t+w}},\forall\;t=1,\ldots,T-w.
\end{equation*}
The dashed blue line in Figure \ref{Fig:Fig10} represents the sample average of $\set{V_{t,s}}^p_{s=1}$, which is a proxy for the return volatility index in the S\&P 500 pool. Note that based on Figure \ref{Fig:Fig10}, all detected change-points are closely related to periods of relatively high values in the return volatility index.

\section{Future directions}\label{Discussion}

We studied the problem of scalable sequential detection of abrupt changes in the precision matrix of high-dimensional sparse GGMs. The objective is to detect a regime-change with maximum delay of $w$, while keeping the false alarm rate below some pre-specified threshold $\delta$. The proposed test $\hat{T}_t$ (in Algorithm \ref{Algo1}) uses a convex barrier function, which is motivated by the formulation of KL-divergence between two zero mean multivariate Gaussian vectors, for comparing the conditional log-likelihood of all nodes before and after a sudden-change. We also established asymptotic performance guarantees of the algorithm under certain regularity conditions on $p$, $w$, sample size, and  pre- and post-change structure of the graph.

Despite recent progress on scalable offline detection of change-points in the dependence structure of graphical models in the \textquotedblleft{\emph{large $p$, small $n$}}\textquotedblright{} framework, there is still a lot of work to be undertaken in the online setting and this study constitutes a first step in that direction. We conclude this section by addressing several potential extensions of Algorithm \ref{Algo1} as topics of future investigation.

\begin{enumerate}[label = (\alph*),leftmargin=*]
\item Extending the applicability of $\hat{T}_t$ to non-Gaussian random fields having a closed-form conditional log-likelihood, such as \emph{Ising} models, is straightforward. However the asymptotic analysis of the associated $\hat{T}_t$ with Ising models is more technically challenging, since $Y^{\paren{t,w}}_s:\;s=1,\ldots,p$ (in Eq. \eqref{Y^tw_s}) is not a set of $\chi^2_w$ random variables under the null hypothesis.
\item The pre-change precision matrix can be sequentially updated after receiving a new observation by \emph{regularized dual averaging (RDA)} procedure \cite{xiao2010dual} instead of using mini-batch scheme in Algorithm \ref{Algo1}. To the best of our knowledge, the  consistency of RDA is formulated in terms of the regret function, unlike $\ell_2$ or operator norm consistency of the CLIME or QUIC mini-batch updates. It is worth mentioning that RDA-based sequential change-point detection was beyond the scope of this paper, as our core asymptotic focus was on online detection, rather than developing $\ell_2$ or operator norm estimation rate of online update for the pre-change precision matrix. 
\item From a statistical standpoint, investigating optimal choices for the barrier function $f$, as well as minimax lower bound on separability of $\bb{H}_{0,t}$ and $\bb{H}_{1,t}$ constitute interesting subjects of future research.
\item We pointed out in part $\paren{d}$ of Remark \ref{CPScenarios} that $\hat{T}_t$ is not capable of detecting regime-shifts affecting a small sub-graph of the network. When a sudden-change is confined in a subset of nodes $B\subset \set{1,\ldots,p}$, $f\paren{Y^{\paren{t,w}_s}}$ concentrates around zero for all nodes $s$ not directly connected to $B$, i.e. $\Omega^{\paren{t}}_{su} = 0,\;\forall u\in B$. In other words, the SNR is reduced in $\hat{T}_t$ by considering all nodes unaffected by the change-point. In this case scanning through a set of potential candidates for $B$ can improve the detection power. Strictly speaking if $B$ is known to belong a class of clusters $\cc{C}$, then this idea can be formulated as the following.
\begin{equation*}
\hat{T}'_t = \max_{B\in\cc{C}} \frac{ \sum_{s\in B} \brac{f\paren{ \hat{Y}^{\paren{t,w}}_{s} } - g_1\paren{w} } }{ g_2\paren{w} \sqrt{\sum_{s_1,s_2\in B}^{p} h_w\paren{ \hat{R}^{\paren{t}}_{s_1,s_2} } } }.
\end{equation*}
Obtaining null and alternative distributions of $\hat{T}'_t$ is a challenging analytic task and relies on extreme value theory for dependent random variables, which is beyond the scope of this manuscript. 
\end{enumerate}

\section{Proofs}\label{Proofs}

\begin{proof}[Proof of Theorem \ref{CLTforT}]
Recall $T_t$ from Eq. \eqref{ApproxTt}. For simplicity, we only state the proof for $w = 1$. The proof can be easily generalized to the case of $w>1$. We introduce new notation for compactness and transparency of algebraic derivations. For any $s\in\set{1,\ldots,p}$, define
\begin{align}\label{DefEQ}
&\bar{f}\paren{Y^{\paren{t,w}}_s} \coloneqq f\paren{Y^{\paren{t,w}}_s} - g_1\paren{w},\quad
\sigma^2_{p,w} \coloneqq g^2_2\paren{w}\sum_{s_1,s_2=1}^{p} h_w\paren{R^{\paren{t}}_{s_1,s_2}},\nonumber\\
&\xi_s \coloneqq \sum_{l=1}^{p} \bar{f}\paren{Y^{\paren{t,w}}_l} \bbM{1}\paren{ R_{s,l} \ne 0},\quad \xi^c_s \coloneqq \sum_{l=1}^{p} \bar{f}\paren{Y^{\paren{t,w}}_l} \bbM{1}\paren{ R_{s,l} = 0}.
\end{align}
Here $R$ stands for the partial correlation matrix of the network. Notice that with the new notation, we have 
\begin{equation}\label{DefEQ2}
T_t = \frac{1}{ \sigma_{p,w} } \sum_{s=1}^{p} \bar{f}\paren{Y^{\paren{t,w}}_s},\quad\mbox{and}\quad \xi_u + \xi^c_u = \sum_{u=1}^{p} \bar{f}\paren{Y^{\paren{t,w}}_u} = \sigma_{p,w} T_t,\;\;\forall\;u=1,\ldots,p. 
\end{equation}
Furthermore, as the main diagonal of $R$ fills with one, then 
\begin{equation}\label{LowBndsigmap}
\sigma^2_{p,w} \geq g^2_2\paren{w}\sum_{s=1}^{p} h_w\paren{R_{ss}} = g^2_2\paren{w}\sum_{s=1}^{p} h_w\paren{1} = g^2_2\paren{w}p \quad \Rightarrow \sigma_{p,w} \geq g_2\paren{w}\sqrt{p}.
\end{equation}
We adopt the proof of Theorem $3.3.1$ of \cite{guyon1995random}. The proof leverages the well-known Stein's result (Lemma $2$ of \cite{bolthausen1982central}). Any sequence of $\bb{L}^2$ integrable random variables $\eta_n$ converges in distribution to a standard normal random variable as $n\rightarrow\infty$, if 
\begin{equation}\label{Stein}
\bb{E}\brac{ \paren{j\nu - \eta_n} e^{j\nu\eta_n} }\rightarrow 0,\quad\forall\;\nu\in\bb{R}.
\end{equation}
Choose an arbitrary $\nu\in\bb{R}$ and define $A_1$, $A_2$, and $A_3$ in the following way.
\begin{align*}
&A_1 \coloneqq \frac{1}{\sigma_{p,w}} e^{j\nu T_t}\sum_{s=1}^{p}\bar{f}\paren{Y^{\paren{t,w}}_s}\brac{1- \frac{j\nu}{\sigma_{p,w}}\xi_s - \exp\paren{-\frac{j\nu\xi_s}{\sigma_{p,w}}} },\\
&A_2 \coloneqq j\nu e^{j\nu T_t}\brac{1-\sigma^{-2}_{p,w} \sum_{s,l=1}^{p} \bar{f}\paren{Y^{\paren{t,w}}_s}\bar{f}\paren{Y^{\paren{t,w}}_l}\bbM{1}\paren{ R_{s,l} \ne 0} } = j\nu e^{j\nu T_t}\brac{1-\sigma^{-2}_{p,w} \sum_{s=1}^{p} \bar{f}\paren{Y^{\paren{t,w}}_s}\xi_s },\\
&A_3 \coloneqq \sigma^{-1}_{p,w}\sum_{s=1}^{p} \paren{\bar{f}\paren{Y^{\paren{t,w}}_s}e^{j\nu\xi^c_s/\sigma_{p,w}} }.
\end{align*}
Using Eq. \eqref{DefEQ2}, we introduce an equivalent formulation for $A_1 + A_3$.
\begin{eqnarray*}
A_1 + A_3 &=& \frac{e^{j\nu T_t}}{\sigma_{p,w}} \sum_{s=1}^{p}\bar{f}\paren{Y^{\paren{t,w}}_s} - \frac{j\nu e^{j\nu T_t}}{\sigma^2_{p,w}}\sum_{s=1}^{p} \bar{f}\paren{Y^{\paren{t,w}}_s}\xi_s - \frac{1}{\sigma_{p,w}} \sum_{s=1}^{p}\bar{f}\paren{Y^{\paren{t,w}}_s} e^{j\nu \paren{T_t-\frac{\xi_s}{\sigma_{p,w}}}} + A_3\\
&=& e^{j\nu T_t} T_t - \frac{j\nu e^{j\nu T_t}}{\sigma^2_{p,w}}\sum_{s=1}^{p} \bar{f}\paren{Y^{\paren{t,w}}_s}\xi_s - \frac{1}{\sigma_{p,w}} \sum_{s=1}^{p}\bar{f}\paren{Y^{\paren{t,w}}_s} e^{j\nu\xi^c_s/\sigma_{p,w}} + A_3\\
&=& e^{j\nu T_t} T_t - \frac{j\nu e^{j\nu T_t}}{\sigma^2_{p,w}}\sum_{s=1}^{p} \bar{f}\paren{Y^{\paren{t,w}}_s}\xi_s - A_3 + A_3 = e^{j\nu T_t} T_t - \frac{j\nu e^{j\nu T_t}}{\sigma^2_{p,w}}\sum_{s=1}^{p} \bar{f}\paren{Y^{\paren{t,w}}_s}\xi_s.
\end{eqnarray*}
Therefore,
\begin{equation}\label{A1A2A3Iden}
\paren{j\nu - T_t} e^{j\nu T_t} = A_2 - \paren{A_1+A_3} \quad\Longrightarrow\quad \bb{E}\brac{\paren{j\nu - T_t} e^{j\nu T_t}} = \bb{E}A_2 - \bb{E}A_1 - \bb{E}A_3.
\end{equation}
Notice that when $w = 1$, there is a zero mean Gaussian random vector $Z\in\bb{R}^p$ such that
\begin{equation*}
Y^{\paren{t,w}}_s = Z^2_s,\;\;\forall\;s \in\set{1,\ldots,p},\quad \mbox{and}\quad\cov\paren{Z} = R.
\end{equation*}
Thus for any $s$, $Y^{\paren{t,w}}_s$ is independent of $\set{Y^{\paren{t,w}}_v:\; v\ne s, R_{v,s} = 0 }$, implying the independence of $\bar{f}\paren{Y^{\paren{t,w}}_s}$ and $\xi^c_s$. So
\begin{equation*}
\bb{E}A_3 = \sigma^{-1}_{p,w} \sum_{s=1}^{p}\bb{E}\paren{\bar{f}\paren{Y^{\paren{t,w}}_s}e^{j\nu\xi^c_s/\sigma_{p,w}} } = \sigma^{-1}_{p,w} \sum_{s=1}^{p}\cov\paren{\bar{f}\paren{Y^{\paren{t,w}}_s}, e^{j\nu\xi^c_s/\sigma_{p,w}} } = 0.
\end{equation*}
Namely, $\bb{E}A_3 = 0$. So, identity \eqref{A1A2A3Iden} can be rewritten as
\begin{equation*}
\bb{E}\brac{\paren{j\nu - T_t} e^{j\nu T_t}} = \bb{E}A_2 - \bb{E}A_1.
\end{equation*}
We just need to show that $\bb{E}\abs{A_1}$ and $\bb{E}\abs{A_2}$ tend to zero. Let us first focus on $\bb{E}\abs{A_2}$. Due to the Cauchy-Schwartz inequality, we only show that $\norm{A_2}{2}$ goes to zero. It is known from Eq. \ref{LowBndsigmap} that $\sigma_{p,w} \geq g_2\paren{w}\sqrt{p}$. A lower bound on $\sigma_{p,w}$ gives an alternative asymptotic representation for $\norm{A_2}{2}$.
\begin{eqnarray}\label{Norm2A2}
\norm{A_2}{2}^2 &=& \nu^2\bb{E}\brac{\abs{\sum_{s,l=1}^{p} \frac{\bar{f}\paren{Y^{\paren{t,w}}_s}\bar{f}\paren{Y^{\paren{t,w}}_l}}{\sigma^2_{p,w}}\bbM{1}_{\set{R_{s,l} \ne 0}}-1}^2} \nonumber\\
&=& \frac{\nu^2}{\sigma^4_{p,w}}\var\brac{\sum_{s,l=1}^{p} \bar{f}\paren{Y^{\paren{t,w}}_s}\bar{f}\paren{Y^{\paren{t,w}}_l}\bbM{1}_{\set{R_{s,l} \ne 0}}}\nonumber\\
&\leq& \frac{\nu^2}{p^2g^4_2\paren{w}} \sum_{R_{s_1,l_1 \ne 0} } \sum_{R_{s_2,l_2 \ne 0} } \cov\Bigbrac{ \bar{f}\paren{Y^{\paren{t,w}}_{s_1}}\bar{f}\paren{Y^{\paren{t,w}}_{l_1}}, \bar{f}\paren{Y^{\paren{t,w}}_{s_2}}\bar{f}\paren{Y^{\paren{t,w}}_{l_2}} }
\end{eqnarray}
We break the summation in Eq. \eqref{Norm2A2} into two parts. We interchangeably use $B_1$ and $B_2$ instead of $\paren{s_1,l_1}$ and $\paren{s_2,l_2}$, respectively. Define
\begin{align*}
&\cc{C}_1 \coloneqq \set{ B_1, B_2: R_{s_1,l_1}, R_{s_2,l_2} \ne 0,\; \mbox{There is no edge between } B_1 \mbox{ and } B_2 },\\
& \cc{C}_2 \coloneqq \set{ B_1, B_2: R_{s_1,l_1}, R_{s_2,l_2} \ne 0,\; \mbox{There is an edge between } B_1 \mbox{ and } B_2 }.
\end{align*}
Proposition \ref{IndepGauss} states that
\begin{equation*}
\sum_{ B_1,B_2\in \cc{C}_1 } \cov\Bigbrac{ \bar{f}\paren{Y^{\paren{t,w}}_{s_1}}\bar{f}\paren{Y^{\paren{t,w}}_{l_1}}, \bar{f}\paren{Y^{\paren{t,w}}_{s_2}}\bar{f}\paren{Y^{\paren{t,w}}_{l_2}} } = 0.
\end{equation*}
So we can simplify the upper bound on $\norm{A_2}{2}$ in Eq. \eqref{Norm2A2} as follows.
\begin{eqnarray}\label{Thm1Eq1}
\norm{A_2}{2}^2 &\leq&\paren{\frac{\nu}{pg^2_2\paren{w}}}^2 \sum_{ B_1,B_2\in \cc{C}_2 } \cov\Bigbrac{ \bar{f}\paren{Y^{\paren{t,w}}_{s_1}}\bar{f}\paren{Y^{\paren{t,w}}_{l_1}}, \bar{f}\paren{Y^{\paren{t,w}}_{s_2}}\bar{f}\paren{Y^{\paren{t,w}}_{l_2}} }\nonumber\\ &\leq& \paren{\frac{\nu}{pg^2_2\paren{w}}}^2 \abs{\cc{C}_2}\max_{ B_1,B_2 \in \cc{C}_2 } \abs{\cov\Bigbrac{ \bar{f}\paren{Y^{\paren{t,w}}_{s_1}}\bar{f}\paren{Y^{\paren{t,w}}_{l_1}}, \bar{f}\paren{Y^{\paren{t,w}}_{s_2}}\bar{f}\paren{Y^{\paren{t,w}}_{l_2}} }}.
\end{eqnarray} 
We discussed in Section that $wg_2\paren{w}$ is uniformly bounded from below on $\bb{N}$. Namely, $\inf_{w\in\bb{N}} \abs{wg_2\paren{w}} > 0$. So Eq. \eqref{Thm1Eq1} can be simplified as the following.
\begin{equation}\label{Thm1Eq3}
\norm{A_2}{2}^2 \leq \paren{\frac{\nu}{p}}^2\abs{\cc{C}_2} \max_{ B_1,B_2\in \cc{C}_2 } w^4\abs{\cov\Bigbrac{ \bar{f}\paren{Y^{\paren{t,w}}_{s_1}}\bar{f}\paren{Y^{\paren{t,w}}_{l_1}}, \bar{f}\paren{Y^{\paren{t,w}}_{s_2}}\bar{f}\paren{Y^{\paren{t,w}}_{l_2}} }}.
\end{equation}
We now control the covariance between $\bar{f}\paren{Y^{\paren{t,w}}_{s_1}}\bar{f}\paren{Y^{\paren{t,w}}_{l_1}}$, and $\bar{f}\paren{Y^{\paren{t,w}}_{s_2}}\bar{f}\paren{Y^{\paren{t,w}}_{l_2}}$ from above. Choose $B_1,B_2\in \cc{C}_2$ arbitrarily. Observe that 
\begin{eqnarray}\label{Thm1Eq2}
\abs{\cov\Bigbrac{ \bar{f}\paren{Y^{\paren{t,w}}_{s_1}}\bar{f}\paren{Y^{\paren{t,w}}_{l_1}}, \bar{f}\paren{Y^{\paren{t,w}}_{s_2}}\bar{f}\paren{Y^{\paren{t,w}}_{l_2}} }} &\leq& \sd\Bigbrac{ \bar{f}\paren{Y^{\paren{t,w}}_{s_1}}\bar{f}\paren{Y^{\paren{t,w}}_{l_1}} } \sd\Bigbrac{ \bar{f}\paren{Y^{\paren{t,w}}_{s_2}}\bar{f}\paren{Y^{\paren{t,w}}_{l_2}} }\nonumber \\
&\leq& \norm{\bar{f}\paren{Y^{\paren{t,w}}_{s_1}}\bar{f}\paren{Y^{\paren{t,w}}_{l_1}}}{2} \norm{\bar{f}\paren{Y^{\paren{t,w}}_{s_2}}\bar{f}\paren{Y^{\paren{t,w}}_{l_2}} }{2} \nonumber \\
&\leq& \norm{\bar{f}\paren{Y^{\paren{t,w}}_{s_1}}}{4} \norm{\bar{f}\paren{Y^{\paren{t,w}}_{l_1}}}{4} \norm{\bar{f}\paren{Y^{\paren{t,w}}_{s_2}}}{4}  \norm{\bar{f}\paren{Y^{\paren{t,w}}_{l_2}}}{4} \nonumber\\
&\leq& \max_{s=1,\ldots,p}\norm{ \bar{f}\paren{Y^{\paren{t,w}}_{s}} }{4}^4.
\end{eqnarray}
Since $\set{Y^{\paren{t,w}}_s:s=1,\ldots,p}$ form a set of identically distributed random variables, then
\begin{equation*}
\norm{\bar{f}\paren{Y^{\paren{t,w}}_{1}}}{4} = \max_{s=1,\ldots,p}\norm{\bar{f}\paren{Y^{\paren{t,w}}_{s}}}{4} < \infty.
\end{equation*}
Let $M_0\paren{w}$ stands for the forth moment of $w\bar{f}\paren{Y^{\paren{t,w}}_1}$. We substantiated the uniform boundedness of $M_0\paren{w}$ in Lemma \ref{Lem4App}. So 
substituting Eq. \eqref{Thm1Eq2} into Eq. \eqref{Thm1Eq3} leads to
\begin{equation}\label{Norm2A2Uppbnd}
\paren{\bb{E}\abs{A_2}}^2 \leq \norm{A_2}{2}^2 \lesssim \paren{\frac{\abs{\nu}}{p}}^2 \abs{\cc{C}_2}M_0\paren{w} \asymp \paren{\frac{\abs{\nu}}{p}}^2 \abs{\cc{C}_2}.
\end{equation}
\begin{clawithinpf}\label{Claim1}
$\abs{\cc{C}_2}\leq 8sd^2_{\max} = 8p\bar{d}d^2_{\max}$	
\end{clawithinpf}
\begin{proof}[Proof of Claim \ref{Claim1}]
Consider an arbitrary pair $\paren{s_1,l_1}, \paren{s_2,l_2}\in \cc{C}_2$. There is a direct edge between $s_1$ and $l_1$ as $R_{s_1,l_1} \ne 0$. Similarly, there should be an edge connecting $s_2$ and $l_2$. A direct connection between $\paren{s_1,l_1}$, $\paren{s_2,l_2}$, means that four vertices $\paren{s_1,l_1,s_2,l_2}$ form a path of length less $1, 2$, or $3$. Let $m_i$ denotes the number of distinct paths of length $i$ in $G_t$. Then 
\begin{equation*}
\abs{\cc{C}_2}\leq 4\paren{m_1 + m_2 + m_3}.
\end{equation*}
Since each node in network is connected to at most $d_{\max}$ other nodes, one can easily verify that $m_1 = s$ and
\begin{equation*}
m_i = \sum_{s=1}^{p} {d_s \choose 2} d^{i-2}_{\max} \leq \sum_{s=1}^{p} \frac{d^2_s}{2} d^{i-2}_{\max} \leq \frac{1}{2}d^{i-1}_{\max} \sum_{s=1}^{p} d_s = \frac{s}{2} d^{i-1}_{\max}.
\end{equation*}
Thus $\abs{\cc{C}_2}\leq 2s\paren{2+d_{\max}+d^2_{\max} } \leq 8sd^2_{\max} = 8p\bar{d}d^2_{\max}$
\end{proof}

Replacing the upper bound on $\abs{\cc{C}_2}$ in Claim \ref{Claim1} into Eq. \eqref{Norm2A2Uppbnd} shows that 
\begin{equation*}
\bb{E}\abs{A_2} \leq \abs{\nu}\sqrt{\frac{8\bar{d} d^2_{\max}}{p}}.
\end{equation*}
Now we find sufficient condition under which $\bb{E}\abs{A_1}\rightarrow 0$. Observe that
\begin{eqnarray}\label{EQA4}
\abs{A_1} &=& \sigma^{-1}_{p,w}\abs{ \sum_{s=1}^{p} \bar{f}\paren{Y^{\paren{t,w}}_s}\paren{1- \frac{j\nu\xi_s}{\sigma_{p,w}} - e^{-j\nu\xi_s/\sigma_{p,w}}} } \leq \sigma^{-1}_{p,w}\sum_{s=1}^{p} \abs{\bar{f}\paren{Y^{\paren{t,w}}_s}} \abs{ 1- \frac{j\nu\xi_s}{\sigma_{p,w}} - e^{-j\nu\xi_s/\sigma_{p,w}} } \nonumber\\
&\RelNum{\paren{a}}{\leq}& \frac{1}{2\sigma_{p,w}} \sum_{s=1}^{p} \abs{\bar{f}\paren{Y^{\paren{t,w}}_s}}\paren{\frac{\nu \xi_s}{\sigma_{p,w}}}^2 = \frac{\nu^2}{2\sigma^3_p}\sum_{s=1}^{p}\xi^2_s\abs{\bar{f}\paren{Y^{\paren{t,w}}_s}}.
\end{eqnarray}
The inequality $\paren{a}$ is implied form the fact that $\abs{e^{-jy}+jy-1}\leq y^2/2$ for any $y\in\bb{R}$. We can further simplify the upper bound on $\bb{E}\abs{A_1}$ by using Holder's inequality with $p = 3$ and $q = 3/2$.
\begin{equation}\label{Thm1Eq4}
\bb{E}\abs{A_1} \leq \frac{\nu^2}{2\sigma^3_p}\sum_{s=1}^{p} \bb{E}\xi^2_s\abs{\bar{f}\paren{Y^{\paren{t,w}}_s}} \leq \frac{\nu^2}{2\sigma^3_p}\sum_{s=1}^{p} \paren{\bb{E}\abs{\xi_s}^3}^{\frac{2}{3}} \paren{\bb{E}\abs{\bar{f}\paren{Y^{\paren{t,w}}_s}}^3}^{\frac{1}{3}}.
\end{equation}
We previously argued in Eq. \eqref{LowBndsigmap} that $\sigma_{p,w} \geq \sqrt{p}g_2\paren{w}\gtrsim w\sqrt{p}$. This fact is crucial for transforming Eq. \eqref{Thm1Eq4} into a desirable form.
\begin{eqnarray}\label{Thm1Eq5}
\bb{E}\abs{A_1} &\lesssim& \nu^2p^{\frac{3}{2}} \sum_{s=1}^{p} \paren{\bb{E}\abs{w\xi_s}^3}^{\frac{2}{3}} \paren{\bb{E}\abs{w\bar{f}\paren{Y^{\paren{t,w}}_s}}^3}^{\frac{1}{3}} = \nu^2p^{-\frac{3}{2}} \sum_{s=1}^{p} \norm{w\xi_s}{3}^2 \norm{w\bar{f}\paren{Y^{\paren{t,w}}_s}}{3}\nonumber\\ &\RelNum{\paren{b}}{\lesssim}& \nu^2p^{-\frac{3}{2}} \sum_{s=1}^{p} \norm{w\xi_s}{3}^2.
\end{eqnarray}
Here inequality $\paren{b}$ is obvious implication of Lemma \ref{Lem4App}. We finally control $\norm{w\xi_s}{3}$ from above. Again by using Holder's inequality with $p=3, q=3/2$ and Lemma \ref{Lem4App}, we get
\begin{eqnarray}\label{Thm1Eq6}
\norm{w\xi_s}{3}^{3} &=& \bb{E}\brac{ \abs{\sum_{l=1}^{p} w\bar{f}\paren{Y^{\paren{t,w}}_l}\bbM{1}_{\set{R_{sl} \ne 0}} }^3 } \leq d^2_s \sum_{l=1}^{p} \bb{E}\paren{\abs{w\bar{f}\paren{Y^{\paren{t,w}}_l}}^3}\bbM{1}_{\set{R_{sl} \ne 0}} \nonumber\\
&\leq& d^3_s \max_{l=1,\ldots,p} \bb{E}\paren{\abs{w\bar{f}\paren{Y^{\paren{t,w}}_l}}^3} \lesssim d^3_s.
\end{eqnarray}
We finally substitute Eq. \eqref{Thm1Eq6} into Eq. \eqref{Thm1Eq6}.
\begin{equation*}
\bb{E}\abs{A_1}\lesssim \nu^2p^{-\frac{3}{2}} \sum_{s=1}^{p} d^2_s \leq \nu^2p^{-\frac{3}{2}} \sum_{s=1}^{p} d_s d_{\max} = \nu^2p^{-\frac{3}{2}} p\bar{d} d_{\max} = \nu^2 \frac{\bar{d} d_{\max}}{\sqrt{p}}.
\end{equation*}
Therefore $\bb{E}\abs{A_1}$ and $\bb{E}\abs{A_2}$ simultaneously tend to zero (for any fixed $\nu\in\bb{R}$) whenever
\begin{equation*}
\sqrt{\frac{\bar{d}d^2_{\max}}{p}} \vee \frac{\bar{d} d_{\max}}{\sqrt{p}}  = \frac{\bar{d} d_{\max}}{\sqrt{p}} \rightarrow 0,
\end{equation*}
which exactly coincides with Assumption \ref{AssuCLT}.
\end{proof}

\begin{proof}[Proof of Theorem \ref{DtctRateT}]
Let $Q_{\eta}$ denotes $\paren{1-\eta}$ quantile of $\cc{N}\paren{0,1}$ distribution, for any $\eta\in\paren{0,1}$. We prove that 
\begin{equation*}
\bb{P}\paren{T_t\geq Q_{\pi_0} \mid \bb{H}_{1,t} }\geq 1-\pi_1.
\end{equation*}
Let $\mu_{p,w}$ and $\sigma^2_{p,w}$ respectively stand for the mean and variance of $T_t$ under alternative hypothesis. Obviously
\begin{equation}\label{Eq1Thm2}
\bb{P}\paren{T_t\geq Q_{\pi_0} \mid \bb{H}_{1,t} } = \bb{P}\paren{ \frac{T_t-\mu_{p,w}}{\sigma_{p,w}} \geq \frac{Q_{\pi_0}-\mu_{p,w}}{\sigma_{p,w}} \mid \bb{H}_{1,t} }.
\end{equation}
Since smallest eigenvalue of $\Omega^{\paren{t+1}}$ is greater than $\alpha_{\min}$ and its $\ell_1$-operator norm is bounded above by $M$, we can show that $\paren{T_t-\mu_{p,w}}/\sigma_{p,w}$ has asymptotically a standard Gaussian distribution by employing exact same techniques as in the proof of Theorem \ref{CLTforT}. So the desirable condition in Eq. \eqref{Eq1Thm2} is satisfied if we can introduce a sufficient condition under which
\begin{equation}\label{Eq2Thm2}
\frac{Q_{\pi_0}-\mu_{p,w}}{\sigma_{p,w}} \leq -Q_{\pi_1},\quad\mbox{or equivalently}\quad \frac{\mu_{p,w}}{\sigma_{p,w}}\paren{1-\frac{Q_{\pi_0}}{\mu_{p,w}}} \geq Q_{\pi_1}.
\end{equation}
The inequality \eqref{Eq2Thm2} trivially holds if we can prove that
\begin{equation}\label{SuffCondThm2}
\mu_{p,w}\geq 2Q_{\pi_0},\quad\mbox{and}\quad\mu_{p,w}\geq 2\sigma_{p,w}Q_{\pi_1}.
\end{equation}
Recall $\Delta$ from Eq. \eqref{Delta} and define
\begin{equation*}
\bar{\Psi}_p \coloneqq \frac{1}{p}\sum_{s=1}^p f\paren{1+\Delta_s},\quad \bar{\Psi}'_p \coloneqq \frac{1}{p}\sum_{s=1}^p \Delta^2_s.
\end{equation*}
We first note that all terms $\Delta_s,\;s=1,\ldots,p$ are uniformly bounded, because for any $s\in\set{1,\ldots,p}$
\begin{equation*}
\Delta_s = \frac{1}{\Omega^{\paren{t}}_{ss} } \Omega^{\paren{t}}_{s,:}\paren{\Omega^{\paren{t+1}}}^{-1} \Omega^{\paren{t}}_{:,s} -1\leq \frac{\LpNorm{\Omega^{\paren{t}}_{:,s}}{2}^2}{\lambda_{\min}\paren{\Omega^{\paren{t+1}}}\Omega^{\paren{t}}_{ss}}-1\leq \frac{\LpNorm{\Omega^{\paren{t}}_{:,s}}{2}^2}{\lambda_{\min}\paren{\Omega^{\paren{t+1}}}\lambda_{\min}\paren{\Omega^{\paren{t}}}}-1\leq \frac{M^2}{\alpha^2_{\min}}-1.
\end{equation*}
So we can find a bounded scalar $C$ (depending on $\frac{M^2}{\alpha^2_{\min}}-1$) such that 
\begin{equation*}
\Delta^2_s \leq Cf\paren{1+\Delta_s},\;\;\forall\;s=1,\ldots,p\quad \Longrightarrow \quad \bar{\Psi}'_p\leq C\bar{\Psi}_p.
\end{equation*}
We now have required tools for introducing a sufficient condition that guarantees two inequalities in Eq. \eqref{SuffCondThm2}. We begin by obtaining a sharp lower bound on $\mu_{p,w}$. Recall $T_t$ from Eq. \eqref{ApproxTt}. Using the first part of Lemma \ref{Lemma4App} yields
\begin{equation}\label{mu_{p,w}}
\mu_{p,w} = \bb{E}\paren{T_t\mid \bb{H}_{1,t}} = \frac{\sum_{s=1}^p f\paren{1+\Delta_s} }{g_2\paren{w}\sqrt{\sum_{s_1,s_2=1}^p h_w\paren{R_{s_1,s_2}} } } \RelNum{\paren{a}}{\asymp} \frac{pw\bar{\Psi}_p}{\sqrt{\sum_{s_1,s_2=1}^p h_w\paren{R_{s_1,s_2}} } }.
\end{equation}
Notice that $\paren{a}$ in above equation holds since $g_2\paren{w}\geq c_0w$ for some universal constant $c_0>0$. Moreover, it is easy to verify the existence of a bounded constant $C_1\paren{M,\alpha_{\min}}$ for which
\begin{equation*}
\sum_{s_1,s_2=1}^p h_w\paren{R_{s_1,s_2}} \leq C_1\paren{M,\alpha_{\min}}p.
\end{equation*}
Thus $\mu_{p,w}$ in Eq. \eqref{mu_{p,w}} satisfies $\mu_{p,w}\gtrsim w\sqrt{p}\bar{\Psi}_p$. So the first desired condition in Eq. \eqref{SuffCondThm2} holds if
\begin{equation}\label{BarPsi_pLowBnd1}
\bar{\Psi}_p\geq C_2\paren{M,\alpha_{\min}}\frac{Q_{\pi_0}}{w\sqrt{p}},
\end{equation} 
for some bounded constant $C_2\paren{M,\alpha_{\min}}$. We now focus on finding a sharp lower bound on $\sigma_{p,w}$.
\setcounter{clawithinpf}{0}
\begin{clawithinpf}\label{Claim1Thm2}
$\sigma^2_{p,w}\leq C_3\paren{M,\alpha_{\min}}\paren{1 \vee \frac{w}{\sqrt{p}}\bar{\Psi}_p }$ for a bounded scalar $C_3\paren{M,\alpha_{\min}}$.	
\end{clawithinpf}
\begin{proof}[Proof of Claim \ref{Claim1Thm2}]
	
\end{proof}

If Claim \ref{Claim1Thm2} holds true, then the second condition in Eq. \eqref{SuffCondThm2} is satisfied when
\begin{equation}\label{BarPsi_pLowBnd2}
w\sqrt{p}\bar{\Psi}_p\geq C_4\paren{M,\alpha_{\min}}Q_{\pi_1}\paren{1 \vee \sqrt{\frac{w}{\sqrt{p}}\bar{\Psi}_p} },
\end{equation}
for some $C_4\paren{M,\alpha_{\min}}<\infty$. It is easy to show that condition \eqref{BarPsi_pLowBnd2} is equivalent to the following relationship. 
\begin{equation}\label{BarPsi_pLowBnd3}
\bar{\Psi}\geq C_5\paren{M,\alpha_{\min}}\frac{Q_{\pi_1}}{w\sqrt{p}}\paren{1+\frac{Q_{\pi_1}}{p}}.
\end{equation}
Finally notice that if $\pi_1 \leq p^{-\xi}$ for some fixed positive scalar $\xi$, then 
\begin{equation*}
\frac{Q_{\pi_1}}{p} = \cc{O}\paren{\frac{\log p}{p}}\rightarrow 0.
\end{equation*}
So combining Eq. \eqref{BarPsi_pLowBnd1} and \eqref{BarPsi_pLowBnd3} terminates the proof.
\end{proof}

\begin{proof}[Proof of Theorem \ref{CLTforTHat}]
Under Assumption \ref{AssuCLT} and, $n_t, p$, and possibly $w,\bar{d}, d_{\max}$ asymptotically grow in such a way that
\begin{equation}\label{Conds}
\frac{\bar{d} d_{\max}}{\sqrt{p}}\rightarrow 0,\quad\mbox{and}\quad \frac{pd_{\max}\log^2 p}{n} \paren{w\vee \log p}\rightarrow 0.
\end{equation}
All the statements regarding probabilistic convergence results in the proof are restricted to large $p$ and $n_t$ framework described in Eq. \eqref{Conds}. According to Theorem \ref{CLTforT}, $T_t$ converges in distribution to a standard Gaussian random variable. Therefore, we only require to show that $\abs{\hat{T}_t-T_t}$ converges to zero, in probability. 

Recall $\sigma_{p,w}$ and $\bar{f}\paren{\cdot}$ from Eq. \eqref{DefEQ} and define $\hat{\sigma}_{p,w}$ (stands for estimated $\sigma_{p,w}$ ) by
\begin{equation*}
\hat{\sigma}^2_{p,w} \coloneqq g^2_2\paren{w}\sum_{s_1,s_2=1}^{p} h_w\paren{\hat{R}^{\paren{t}}_{s_1,s_2}}.
\end{equation*}
Using triangle inequality leads to
\begin{eqnarray}\label{THatMinusT}
\abs{\hat{T}_t-T_t} &=& \abs{\frac{1}{\hat{\sigma}_{p,w}}\sum_{s=1}^p \bar{f}\paren{\hat{Y}^{\paren{t,w}}_s} - \frac{1}{\sigma_{p,w}}\sum_{s=1}^p \bar{f}\paren{Y^{\paren{t,w}}_s}}\nonumber\\
&\leq& \frac{1}{\sigma_{p,w}}\abs{\sum_{s=1}^{p} \bar{f}\paren{\hat{Y}^{\paren{t,w}}_s} - \bar{f}\paren{Y^{\paren{t,w}}_s}} + \abs{\frac{1}{\hat{\sigma}_{p,w}}-\frac{1}{\sigma_{p,w}}} \abs{\sum_{s=1}^{p} \bar{f}\paren{Y^{\paren{t,w}}_s}}\nonumber\\
&=&\frac{1}{\sigma_{p,w}}\abs{\sum_{s=1}^{p} \bar{f}\paren{\hat{Y}^{\paren{t,w}}_s} - \bar{f}\paren{Y^{\paren{t,w}}_s}} + \abs{\paren{\frac{\sigma_{p,w}}{\hat{\sigma}_{p,w}}-1}T_t}.
\end{eqnarray} 
The following facts provides a simpler upper bound on $\abs{\hat{T}_t-T_t}$.
\begin{itemize}
\item $\sigma_{p,w}$ is (obviously) greater than $g_2\paren{w}\sqrt{p}$.
\item Based upon  Theorem \ref{CLTforT}, $T_t$ converges in distribution to a standard Gaussian random variable. Therefore, in the asymptotic regime
\begin{equation*}
\bb{P}\paren{\abs{T_t} \geq 4\sqrt{\log p}}\leq p^{-1}.
\end{equation*}
\end{itemize}
Applying these facts on inequality \eqref{THatMinusT} yields
\begin{equation}\label{THatMinusT2}
\bb{P}\paren{ \frac{\abs{\hat{T}_t-T_t}}{4} \leq \abs{\sum_{s=1}^{p} \frac{\bar{f}\paren{\hat{Y}^{\paren{t,w}}_s} - \bar{f}\paren{Y^{\paren{t,w}}_s}}{g_2\paren{w}\sqrt{p}}} + \abs{\frac{\sigma_{p,w}}{\hat{\sigma}_{p,w}}-1}\sqrt{\log p} }\geq 1-\frac{1}{p}.
\end{equation}
For further simplification of the upper bound on $\abs{\hat{T}_t-T_t}$, we need to control $\abs{\hat{\sigma}_{p,w}-\sigma_{p,w}}$ from above.

\setcounter{clawithinpf}{0}
\begin{clawithinpf}\label{Claim1Thm3}
There exists a bounded constant $C\paren{\alpha_{\min}, M}$ such that
\begin{equation*}
\bb{P}\paren{\abs{\frac{\hat{\sigma}_{p,w}}{\sigma_{p,w}}-1} \geq C\paren{\alpha_{\min}, M}\varepsilon_{n_t,p,d_{\max}} }\leq \frac{2}{p},\quad\mbox{where}\quad \varepsilon_{n_t,p,d_{\max}} \coloneqq d^{\frac{1}{4}}_{\max}\sqrt{\frac{\log p}{n_t}}.
\end{equation*}
\end{clawithinpf}

Before proving Claim \ref{Claim1Thm3}, we use it for simplifying Eq. \eqref{THatMinusT2}. Under conditions in Eq. \eqref{Conds}, $\varepsilon_{n_t,p,d_{\max}}\sqrt{\log p}$ converges to zero, as
\begin{equation*}
\varepsilon_{n_t,p,d_{\max}}\sqrt{\log p}  = \frac{d^{\frac{1}{4}}_{max}\log p}{\sqrt{n_t}} \lesssim \frac{d^{\frac{1}{4}}_{max}\log p}{\sqrt{pd_{\max}\log^2 p}} = \paren{p\sqrt{d_{\max}}}^{-\frac{1}{2}}\rightarrow 0.
\end{equation*}
Therefore for large $p$ and $n$, $2\hat{\sigma}_{p,w}\geq\sigma_{p,w}$ with probability at least $1-p^{-1}$, and
\begin{eqnarray}\label{SigmaHatMinusSigma}
\abs{\frac{\sigma_{p,w}}{\hat{\sigma}_{p,w}}-1}\sqrt{\log p} &=& \frac{\abs{\hat{\sigma}_{p,w}-\sigma_{p,w}}}{\hat{\sigma}_{p,w}}\sqrt{\log p} \leq \frac{2\abs{\hat{\sigma}_{p,w}-\sigma_{p,w}}}{\sigma_{p,w}}\sqrt{\log p}\leq 2C\paren{\alpha_{\min}, M} \varepsilon_{n_t,p,d_{\max}}\sqrt{\log p} \nonumber\\
&\asymp& \varepsilon_{n_t,p,d_{\max}}\sqrt{\log p}\rightarrow 0.
\end{eqnarray}
Replacing Eq. \eqref{SigmaHatMinusSigma} into Eq. \eqref{THatMinusT2} ensures that
\begin{equation*}
\abs{\hat{T}_t-T_t}  = \cc{O}_{\bb{P}}\paren{ \abs{\sum_{s=1}^{p} \frac{\bar{f}\paren{\hat{Y}^{\paren{t,w}}_s} - \bar{f}\paren{Y^{\paren{t,w}}_s}}{g_2\paren{w}\sqrt{p}}} \vee \varepsilon_{n_t,p,d_{\max}}\sqrt{\log p} }.
\end{equation*}
So $\abs{\hat{T}_t-T_t}\rightarrow 0$ (in probability), which terminates the proof, if we prove that 
\begin{equation*}
\sum_{s=1}^{p}\abs{\bar{f}\paren{\hat{Y}^{\paren{t,w}}_s} - \bar{f}\paren{Y^{\paren{t,w}}_s}} = o_{\bb{P}}\paren{g_2\paren{w}\sqrt{p}}.
\end{equation*}
Since $wg_2\paren{w}$ is strictly less than some bounded universal constant $C'_0$, we just need to show that
\begin{equation}\label{FinalClaim}
\frac{w}{\sqrt{p}}\sum_{s=1}^{p}\abs{\bar{f}\paren{\hat{Y}^{\paren{t,w}}_s} - \bar{f}\paren{Y^{\paren{t,w}}_s}} = o_{\bb{P}}\paren{1}.
\end{equation}
Define
\begin{equation}\label{eta_s}
\eta_s \coloneqq \frac{ \hat{\Omega}^{\paren{t}}_{s,:} \paren{\Omega^{\paren{t}}}^{-1} \hat{\Omega}^{\paren{t}}_{:,s} }{\hat{\Omega}^{\paren{t}}_{ss}},\quad \forall\;s=1,\ldots,p.
\end{equation}
$\eta_s$ is an important quantity in our analysis as it captures the conditional expected value of $\hat{Y}^{\paren{t,w}}_s$ given $X_1,\ldots,X_t$. Particularly, it is easy to show that
\begin{equation*}
\bb{E}\brac{Y^{\paren{t,w}}_s\mid X_1,\ldots,X_t} = 1,\quad\mbox{and}\quad \bb{E}\brac{\hat{Y}^{\paren{t,w}}_s \mid X_1,\ldots,X_t} = \eta_s.
\end{equation*}
A vigilant reader notices that despite random nature of $\eta_s$, it does not depend on $X_{t+1},\ldots, X_{t+w}$. So using Corollary \ref{SensAnalfCor} yields
\begin{equation*}
\bb{P}\brac{\abs{\bar{f}\paren{\hat{Y}^{\paren{t,w}}_s} - \bar{f}\paren{Y^{\paren{t,w}}_s}}\geq f\paren{\eta_s} + \abs{\eta_s-1}\sqrt{\frac{8\log p}{w}}\paren{1\vee \sqrt{\frac{8\log p}{w}} } } \leq p^{-2},\quad \forall\;s=1\ldots,p.
\end{equation*}
Using union bound technique, we get
\begin{equation*}
\bb{P}\brac{\abs{\bar{f}\paren{\hat{Y}^{\paren{t,w}}_s} - \bar{f}\paren{Y^{\paren{t,w}}_s}}\geq f\paren{\eta_s} + \abs{\eta_s-1}\sqrt{\frac{8\log p}{w}}\paren{1\vee \sqrt{\frac{8\log p}{w}} },\quad \forall\;s=1\ldots,p } \leq p^{-1}.
\end{equation*}
In summary we showed that 
\begin{equation}\label{fSensAnalEq1}
\frac{w}{\sqrt{p}}\sum_{s=1}^{p}\abs{\bar{f}\paren{\hat{Y}^{\paren{t,w}}_s} - \bar{f}\paren{Y^{\paren{t,w}}_s}} = \cc{O}_{\bb{P}}\paren{\frac{w\sum_{s=1}^p f\paren{\eta_s}}{\sqrt{p}} + \frac{\log p \vee \sqrt{w\log p}}{\sqrt{p}} \sum_{s=1}^p \abs{\eta_s-1} }.
\end{equation}
Using Cauchy-Schwartz inequality, we can rewrite Eq. \eqref{fSensAnalEq1} as the following.
\begin{equation}\label{fSensAnalEq2}
\frac{w}{\sqrt{p}}\sum_{s=1}^{p}\abs{\bar{f}\paren{\hat{Y}^{\paren{t,w}}_s} - \bar{f}\paren{Y^{\paren{t,w}}_s}} = \cc{O}_{\bb{P}}\paren{\frac{w\sum_{s=1}^p f\paren{\eta_s}}{\sqrt{p}} + \paren{\log p \vee \sqrt{w\log p}} \paren{\sum_{s=1}^p \abs{\eta_s-1}^2}^{\frac{1}{2}} }.
\end{equation}
Let us investigate the behaviour of $f\paren{\eta_s}$ for $s=1,\ldots,p$. We assumed that $\lambda_{\min}\paren{\Omega^{\paren{t}}}$ is greater than $\alpha_{\min}$. It is also known that (see Theorem $1$ of \cite{cai2011constrained}) with probability at least $1-1/p$
\begin{equation*}
\OpNorm{\hat{\Omega}^{\paren{t}}-\Omega^{\paren{t}}}{2}{2} \leq CM^2d_{\max}\sqrt{\frac{\log p}{n}} = \cc{O}_{\bb{P}}\paren{d_{\max}\sqrt{\frac{\log p}{n}}}\rightarrow 0.
\end{equation*}
for some bounded universal constant $C$. Thus
\begin{equation}\label{delta_sUppBnd}
\paren{1-\frac{CM^2d_{\max}}{\alpha_{\min}}\sqrt{\frac{\log p}{n}}}\hat{\Omega}^{\paren{t}}\preccurlyeq\Omega^{\paren{t}}\preccurlyeq \paren{1+\frac{CM^2d_{\max}}{\alpha_{\min}}\sqrt{\frac{\log p}{n}}}\hat{\Omega}^{\paren{t}}.
\end{equation}
The inequality \eqref{delta_sUppBnd} is essential for studying the behaviour of $\eta_s$, as
\begin{equation*}
\abs{\eta_s-1}\leq \paren{1+\frac{CM^2d_{\max}}{\alpha_{\min}}\sqrt{\frac{\log p}{n}}}\frac{ \hat{\Omega}^{\paren{t}}_{s,:} \paren{\hat{\Omega}^{\paren{t}}}^{-1} \hat{\Omega}^{\paren{t}}_{:,s} }{\hat{\Omega}_{ss}} - 1 = \frac{CM^2d_{\max}}{\alpha_{\min}}\sqrt{\frac{\log p}{n}}\rightarrow 0,\quad\forall\;s=1,\ldots,p.
\end{equation*}
So all $\eta_s$ are in a small neighbourhood around one, which means that for large enough $p$ and $n_t$
\begin{equation*}
f\paren{\eta_s} \leq 2\paren{\eta_s-1}^2. 
\end{equation*}
From Proposition \ref{CondProbProp} we know that if we replace the last inequality into Eq. \eqref{fSensAnalEq1}, then we get
\begin{equation}\label{fSensAnalEq3}
\frac{w}{\sqrt{p}}\sum_{s=1}^{p}\abs{\bar{f}\paren{\hat{Y}^{\paren{t,w}}_s} - \bar{f}\paren{Y^{\paren{t,w}}_s}} = \cc{O}_{\bb{P}}\paren{\frac{\sum_{s=1}^p \paren{\eta_s-1}^2}{w^{-1}\sqrt{p}} + \paren{\log p \vee \sqrt{w\log p}} \paren{\sum_{s=1}^p \abs{\eta_s-1}^2}^{\frac{1}{2}} }.
\end{equation}
\begin{clawithinpf}\label{Claim2Thm3}
$\sum_{s=1}^p \paren{\eta_s-1}^2 = \cc{O}_{\bb{P}}\paren{\frac{pd_{\max}\log p}{n_t}}$. The constant involved in $\cc{O}_{\bb{P}}$ depends on $M$ and $\alpha_{\min}$.
\end{clawithinpf}
Given Claim $2$, by applying Proposition \ref{CondProbProp} we can rewrite Eq. \eqref{fSensAnalEq3} in the following way.
\begin{eqnarray*}
\frac{w}{\sqrt{p}}\sum_{s=1}^{p}\abs{\bar{f}\paren{\hat{Y}^{\paren{t,w}}_s} - \bar{f}\paren{Y^{\paren{t,w}}_s}} &=& \cc{O}_{\bb{P}}\paren{ \frac{pwd_{\max}\log p}{\sqrt{p}} + \sqrt{\frac{pd_{\max}\log^2 p \paren{w\vee \log p}}{n_t}} } \\
&=& \cc{O}_{\bb{P}}\paren{\sqrt{\frac{pd_{\max}\log^2 p \paren{w\vee \log p}}{n_t}}} \RelNum{\paren{a}}{=} o_{\bb{P}}\paren{1}.
\end{eqnarray*}
Here identity $\paren{a}$ is implied form assumptions in Eq. \eqref{Conds}. Notice that the last equation is identical as Eq. \eqref{FinalClaim}, which concludes the proof. Finally, we establish Claim \ref{Claim1Thm3} and \ref{Claim2Thm3}.
\begin{proof}[Proof of Claim \ref{Claim1Thm3}]
Lemma \ref{LemmaB4} states that as long as $\alpha_{\min} > 0$, the following result holds with probability at least $1-p^{-1}$, and for some bounded $C'\paren{\alpha_{\min}}$.
\begin{eqnarray}\label{Claim1Thm3Ineq}
\abs{\frac{\hat{\sigma}_{p,w}}{\sigma_{p,w}}-1} &\leq& C'\paren{\alpha_{\min}}\paren{\LpNorm{R}{4}^{-1}\sqrt{\LpNorm{\hat{\Omega}^{\paren{t}}-\Omega^{\paren{t}}}{\infty} \LpNorm{\hat{\Omega}^{\paren{t}}-\Omega^{\paren{t}}}{2}} \vee \LpNorm{\hat{\Omega}^{\paren{t}}-\Omega^{\paren{t}}}{\infty}} \nonumber\\
&\RelNum{\paren{a}}{\lesssim}& p^{-\frac{1}{4}} \sqrt{\LpNorm{\hat{\Omega}^{\paren{t}}-\Omega^{\paren{t}}}{\infty} \LpNorm{\hat{\Omega}^{\paren{t}}-\Omega^{\paren{t}}}{2}} \vee \LpNorm{\hat{\Omega}^{\paren{t}}-\Omega^{\paren{t}}}{\infty}.
\end{eqnarray}
The inequality $\paren{a}$ is valid as all diagonal entries of $R$ are equal to one. It is also known that (see Theorem $4$ of \cite{cai2011constrained}) with probability at least $1-p^{-1}$
\begin{equation}\label{CLIMECons}
\LpNorm{\hat{\Omega}^{\paren{t}}-\Omega^{\paren{t}}}{\infty} \leq C_0M^2\sqrt{\frac{\log p}{n_t}},\quad\mbox{and}\quad \LpNorm{\hat{\Omega}^{\paren{t}}-\Omega^{\paren{t}}}{2} \leq C_0M^2\sqrt{\frac{pd_{\max}\log p}{n_t}},
\end{equation}
for some bounded universal constant $C_0$. Combining Eq. \eqref{CLIMECons} and Eq. \eqref{Claim1Thm3Ineq} ends the proof.
\end{proof}

\begin{proof}[Proof of Claim \ref{Claim2Thm3}]
Recall $\eta_s$ from Eq. \eqref{eta_s}. Observe that
\begin{equation*}
\eta_s = \frac{ \hat{\Omega}^{\paren{t}}_{s,:} \paren{\Omega^{\paren{t}}}^{-1} \hat{\Omega}^{\paren{t}}_{:,s} }{\hat{\Omega}^{\paren{t}}_{ss}} = \frac{ \LpNorm{\paren{\Omega^{\paren{t}}}^{-1/2} \hat{\Omega}^{\paren{t}}_{:,s}}{2}^2 }{\hat{\Omega}^{\paren{t}}_{ss}} = \paren{\frac{ \LpNorm{\paren{\Omega^{\paren{t}}}^{-1/2} \hat{\Omega}^{\paren{t}}_{:,s}}{2} }{\sqrt{\hat{\Omega}^{\paren{t}}_{ss}}}}^{2}
\end{equation*}
Using Eq. \eqref{delta_sUppBnd}, we showed that all $\delta_s,\;s=1,\ldots,p$ are in a small neighborhood of one, with probability at least $1-1/p$. Thus, $\abs{\eta_s-1}\leq 3\abs{\sqrt{\eta_s}-1}$ in the asymptotic framework and hence
\begin{eqnarray}\label{eta_sUppBnd}
\abs{\sqrt{\eta_s} - 1 } &=& \abs{\frac{ \LpNorm{\paren{\Omega^{\paren{t}}}^{-1/2} \hat{\Omega}^{\paren{t}}_{:,s}}{2} }{\sqrt{\hat{\Omega}^{\paren{t}}_{ss}}} - 1} \RelNum{\paren{b}}{=} \abs{\frac{ \LpNorm{\paren{\Omega^{\paren{t}}}^{-1/2} \hat{\Omega}^{\paren{t}}_{:,s}}{2} }{\sqrt{\hat{\Omega}^{\paren{t}}_{ss}}} - \frac{ \LpNorm{\paren{\Omega^{\paren{t}}}^{-1/2} \Omega^{\paren{t}}_{:,s}}{2} }{\sqrt{\Omega^{\paren{t}}_{ss}}}}\nonumber\\
&\leq&\LpNorm{ \paren{\Omega^{\paren{t}}}^{-1/2} \paren{ \frac{\hat{\Omega}^{\paren{t}}_{:,s}}{\sqrt{\hat{\Omega}_{ss}}} - \frac{\Omega^{\paren{t}}_{:,s}}{\sqrt{\Omega^{\paren{t}}_{ss}}}}}{2}\leq \sqrt{\frac{1}{ \lambda_{\min}\paren{\Omega^{\paren{t}}} }}\LpNorm{  \frac{\hat{\Omega}^{\paren{t}}_{:,s}}{\sqrt{\hat{\Omega}_{ss}}} - \frac{\Omega^{\paren{t}}_{:,s}}{\sqrt{\Omega^{\paren{t}}_{ss}}}}{2}\nonumber\\
&\leq&\alpha^{-\frac{1}{2}}_{\min}\LpNorm{  \frac{\hat{\Omega}^{\paren{t}}_{:,s}}{\sqrt{\hat{\Omega}^{\paren{t}}_{ss}}} - \frac{\Omega^{\paren{t}}_{:,s}}{\sqrt{\Omega^{\paren{t}}_{ss}}}}{2}.
\end{eqnarray}
Note that identity $\paren{b}$ is an immediate consequence of the fact that 
\begin{equation*}
\Omega^{\paren{t}}_{ss} = \brac{ \Omega^{\paren{t}} \paren{\Omega^{\paren{t}}}^{-1} \Omega^{\paren{t}} } = \Omega^{\paren{t}}_{:,s} \paren{\Omega^{\paren{t}}}^{-1} \Omega^{\paren{t}}_{:,s} = \LpNorm{\paren{\Omega^{\paren{t}}}^{-\frac{1}{2}} \Omega^{\paren{t}}_{:,s} }{2}^2.
\end{equation*}
For simplicity define the following two matrices
\begin{equation*}
L^{\paren{t}} \coloneqq \brac{ \frac{ \Omega^{\paren{t}}_{s_1,s_2} }{ \sqrt{\Omega^{\paren{t}}_{s_2,s_2}} } }_{s_1,s_2=1}^p,\quad \hat{L}^{\paren{t}} \coloneqq \brac{ \frac{ \hat{\Omega}^{\paren{t}}_{s_1,s_2} }{ \sqrt{\hat{\Omega}^{\paren{t}}_{s_2,s_2}} } }_{s_1,s_2=1}^p.
\end{equation*}
Next, we find an upper bound on $\sum_{s=1}^p \paren{\eta_s-1}^2$ by using Eq. \eqref{eta_sUppBnd}.
\begin{equation*}
\sum_{s=1}^p \paren{\eta_s-1}^2 \leq 9\sum_{s=1}^p \paren{\sqrt{\eta_s}-1}^2 \leq \frac{9}{\alpha_{\min}} \sum_{s=1}^p \LpNorm{ \hat{L}^{\paren{t}}_{:,s}-L^{\paren{t}}_{:,s} }{2}^2 = \frac{9}{\alpha_{\min}} \LpNorm{ \hat{L}^{\paren{t}}-L^{\paren{t}} }{2}^2.
\end{equation*}
Lemma \ref{SensAnalL} provides an upper bound on $\LpNorm{ \hat{L}^{\paren{t}}-L^{\paren{t}} }{2}$. There is $C_1<\infty$ (depending on $\alpha_{\min}$) such that
\begin{eqnarray*}
\LpNorm{ \hat{L}^{\paren{t}}-L^{\paren{t}} }{2}^2&\leq& C_1\paren{ \LpNorm{\Omega^{\paren{t}}}{2}^2 \LpNorm{\hat{\Omega}^{\paren{t}}-\Omega^{\paren{t}}}{\infty}^2 + \LpNorm{\hat{\Omega}^{\paren{t}}-\Omega^{\paren{t}}}{2}^2 }\\
&\RelNum{\paren{c}}{=}&\cc{O}_{\bb{P}}\paren{ M^2\frac{\log p}{n_t}\LpNorm{\Omega^{\paren{t}}}{2}^2 + M^2\frac{pd_{\max}\log p}{n_t} }.
\end{eqnarray*}
Note that identity $\paren{c}$ is implied by the rates obtained in Theorem $4$ of \cite{cai2011constrained}. Recall that $\ell_1$ operator norm of $\Omega^{\paren{t}}$ is less than $M$. Therefore
\begin{equation*}
\LpNorm{\Omega^{\paren{t}}}{2}^2 \leq p\OpNorm{\Omega^{\paren{t}}}{2}{2}^2 \leq p\OpNorm{\Omega^{\paren{t}}}{2}{2}^2\leq pM^2.
\end{equation*}
We conclude the proof by combining the last two inequalities.
\begin{equation*}
\LpNorm{ \hat{L}^{\paren{t}}-L^{\paren{t}} }{2}^2 = \cc{O}_{\bb{P}}\paren{ M^4\frac{p\log p}{n_t} + M^2\frac{pd_{\max}\log p}{n_t} } = \cc{O}_{\bb{P}}\paren{ \frac{pd_{\max}\log p}{n_t} }.
\end{equation*}
\end{proof}

\end{proof}

\appendix
\appendixpage
\makeatletter
\def\@seccntformat#1{\csname Pref@#1\endcsname \csname the#1\endcsname\quad}
\def\Pref@section{~}
\makeatother

\counterwithin{thm}{section}
\counterwithin{assu}{section}
\counterwithin{lem}{section}
\counterwithin{cor}{section}
\counterwithin{prop}{section}

\section{Auxiliary results on moments of a Chi-square random variable}\label{AppendixA}

We used $Y^{\paren{t,w}}_s$ (recall its definition from Eq. \eqref{Y^tw_s}) to denote the sample mean of conditional log-likelihood of node $s$ given other nodes of $G_t$, over a window of size $w$. We argued in Section \ref{DtctAlgo} that if GGM does not experience a sudden change at $t+1$, then $wY^{\paren{t,w}}_s$ is distributed as a Chi-square random variable with $w$ degrees of freedom. Also recall $f:\brac{0,\infty}\mapsto\bb{R}$ from Eq. \eqref{f}. In this section we formulate central moments of $f\paren{Y^{\paren{t,w}}_s},\;s=1,\ldots,p$ under $\bb{H}_{0,t}$ (no change at $t+1$) and $\bb{H}_{1,t}$ (abrupt change at $t+1$). The two following results, Lemma \ref{Lemma0App} and \ref{Lemma3App}, are crucial in the proof of principal results in this section.

\begin{lem}\label{Lemma0App}
Let $X$ and $Y$ be two real-valued independent zero-mean random variables. For arbitrary functions $f_1,f_2:\bb{R}\mapsto\bb{R}$, we have
\begin{equation*}
\cov\Bigparen{ f_1\paren{X}, Yf_2\paren{X}} = 0.
\end{equation*}
\end{lem}

\begin{proof}
Independence of $X$ and $Y$ means that 
\begin{eqnarray*}
\cov\Bigparen{ f_1\paren{X}, Yf_2\paren{X}} &=& \bb{E} f_1\paren{X}f_2\paren{X}Y - \bb{E}f_1\paren{X} \bb{E}f_2\paren{X}Y =  \bb{E} f_1\paren{X}f_2\paren{X} \bb{E} Y -  \bb{E} f_1\paren{X} \bb{E}f_2\paren{X}\bb{E}Y \\
&=& \bb{E}Y \cov\Bigparen{f_1\paren{X}, f_2\paren{X}}.
\end{eqnarray*}
Thus $f_1\paren{X}$ and $Yf_2\paren{X}$ are obviously uncorrelated, since $\bb{E}Y = 0$.
\end{proof}

\begin{lem}\label{Lemma3App}
Let $\paren{X,X'}$ be a standard bi-variate Gaussian random vector with $\corr\paren{X, X'} = r$. Let $\set{\paren{X_k, X'_k}}^w_{k=1}$ be i.i.d. draws from $\paren{X,X'}$ and define
\begin{equation*}
Z_w \coloneqq \sum_{k=1}^{w} X^2_k,\quad Z'_w \coloneqq \sum_{k=1}^{w} X'^2_k.
\end{equation*}
Then,
\begin{align*}
&p\paren{r,w}\coloneqq \cov\brac{ \paren{Z_w - w}^2, \paren{Z'_w - w}^2 } = 8r^2\brac{4w+r^2w^2+2r^2}, \\
&q\paren{r,w}\coloneqq\cov\brac{ \paren{Z_w - w}^2, \paren{Z'_w - w}^3 } = 72r^4\paren{w^2+w+2} + 240r^2w.
\end{align*}
\end{lem}

\begin{proof}
$Z_w$ and $Z'_w$ are identically distributed chi-square random variables with $w$ degrees of freedom. We aim to find a recursive formula for $p\paren{r,w}$ in terms of $p\paren{r,w-1}$ and $p\paren{r,1}$. $Z_w$ and $Z'_w$ can be formulated in a recursive way.
\begin{equation*}
Z_w - w = \Bigparen{Z_{w-1} - \paren{w-1}} + X^2_w-1, \quad Z'_w - w = \Bigparen{Z'_{w-1} -\paren{w-1} } + \paren{X'^2_w-1}.
\end{equation*}
For simplicity, define $\bar{Z}_w = Z_w-w$ and $\bar{Z}'_w =Z'_w-w$. Notice that the pair of random variables $\paren{\bar{Z}_{w-1}, \bar{Z}'_{w-1}}$ are independent from $X'_w$ and $X_w$. Thus,
\begin{equation}\label{Lem5.2Eq3}
\cov\Bigbrac{ \bar{Z}^2_{w-1}, \paren{X'^2_w-1}^2 } = 0,\quad \cov\Bigbrac{ \bar{Z}'^2_{w-1}, \paren{X^2_w-1}^2 } = 0.
\end{equation}
Furthermore, Lemma \ref{Lemma0App} implies that
\begin{eqnarray}\label{Lem5.2Eq4}
&\cov\Bigbrac{ \paren{X^2_w-1}^2, \bar{Z}'_{w-1}\paren{X'^2_w-1} } = 0,
\quad\cov\Bigbrac{ \paren{X'^2_w-1}^2, \bar{Z}_{w-1}\paren{X^2_w-1} } = 0,\nonumber\\
&\cov\Bigbrac{ \bar{Z}^2_{w-1}, \bar{Z}'_{w-1}\paren{X'^2_w-1} } = 0,
\quad\cov\Bigbrac{ \bar{Z}'^2_{w-1}, \bar{Z}_{w-1}\paren{X^2_w-1} } = 0.
\end{eqnarray}
The identities \eqref{Lem5.2Eq3} and \eqref{Lem5.2Eq4} are crucial for finding a simple alternative way of expressing $p\paren{r,w}$. 
\begin{eqnarray*}
p\paren{r,w} &=& \cov\Bigbrac{ \paren{X^2_w-1}^2 + \bar{Z}^2_{w-1} + 2\bar{Z}_{w-1}\paren{X^2_w-1}, \paren{X'^2_w-1}^2 + \bar{Z}'^2_{w-1} + 2\bar{Z}'_{w-1}\paren{X'^2_w-1} }\\
&=& \cov\brac{ \paren{X^2_w-1}^2, \paren{X'^2_w-1}^2 } + \cov\brac{ \bar{Z}^2_{w-1}, \bar{Z}^2_{w-1} } + 4\cov\brac{ \bar{Z}_{w-1}\paren{X^2_w-1}, \bar{Z}'_{w-1}\paren{X'^2_w-1} }\\
&=& p\paren{r,1} + p\paren{r,w-1} + 4\cov\brac{ \bar{Z}_{w-1}\paren{X^2_w-1}, \bar{Z}'_{w-1}\paren{X'^2_w-1} }.
\end{eqnarray*}
Next we evaluate $\cov\brac{ \bar{Z}_{w-1}\paren{X^2_w-1}, \bar{Z}'_{w-1}\paren{X'^2_w-1} }$ in terms of $w$ and $r$. Since $X^2_w-1$ is a zero mean random variable, so
\begin{equation*}
\bb{E} \bar{Z}_{w-1}\paren{X^2_w-1} = \bb{E} \bar{Z}_{w-1} \bb{E}\paren{X^2_w-1} = 0.
\end{equation*}
Therefore,
\begin{eqnarray*}
\cov\brac{ \bar{Z}_{w-1}\paren{X^2_w-1}, \bar{Z}'_{w-1}\paren{X'^2_w-1} } &=& \bb{E}\Bigbrac{\bar{Z}_{w-1}\paren{X^2_w-1}\bar{Z}'_{w-1}\paren{X'^2_w-1}} \\
&=& \bb{E}\paren{\bar{Z}_{w-1}\bar{Z}'_{w-1}} \bb{E}\Bigbrac{\paren{X'^2_w-1}\paren{X^2_w-1}} \\
&=& \cov\paren{\bar{Z}_{w-1},\bar{Z}'_{w-1}}\cov\paren{X^2_w, X'^2_w }\\
&=& \sum_{i,j=1}^{w-1}\cov\paren{X^2_i, X'^2_j }\cov\paren{X^2_w, X'^2_w } \RelNum{\paren{a}}{=} \paren{w-1}\brac{\cov\paren{X^2_1, X'^2_1 }}^2\\
&\RelNum{\paren{b}}{=}& 4r^4\paren{w-1}
\end{eqnarray*}
Identity $\paren{a}$ is an obvious consequence of the independence of $X_i$ and $X'_j$, for any distinct pair $\paren{i,j}$. Finally equality $\paren{b}$ is implied from the fact that $\cov\paren{X^2_1, X'^2_1 } = 2r^2$. Thus,
\begin{equation}\label{recursive}
p\paren{r,w} = p\paren{r,w-1} + p\paren{r,1} + 4r^4\paren{w-1},\quad\forall\;w>1.
\end{equation}
Eq. \eqref{recursive} can be easily reformulated as the following way
\begin{equation}\label{prw}
p\paren{r,w} = p\paren{r,1} + \paren{w-1}p\paren{r,1} + 4r^4\sum_{k=1}^{w} \paren{k-1} = wp\paren{r,1} + 8r^4\paren{w-1}\paren{w-2}.
\end{equation}

In the last step, we calculate $p\paren{r,1}$. The cumbersome algebraic details (which is mainly based on using moment identities for standard Gaussian distribution) are omitted due to space constraint. The fact that $Z_1$ and $Z'_1$ has the same distribution yields
\begin{eqnarray}\label{Lem5.2Eq0}
p\paren{r,1} &=& \bb{E}\brac{ \paren{Z_1 - 1}^2 \paren{Z'_1 - 1}^2 } - \var\paren{Z_1} \var\paren{Z'_1} = \bb{E}\brac{ \paren{Z_1 - 1}^2 \paren{Z'_1 - 1}^2 } - 4 \nonumber\\
&=& \bb{E}Z^2_1Z'^2_1 - 4\bb{E}Z^2_1Z'_1 + 2\bb{E}Z^2_1 + 1 - 4\bb{E}Z_1 + 4\bb{E}Z_1Z'_1 - 4 \nonumber \\
&=& \bb{E}Z^2_1Z'^2_1 - 4\bb{E}Z^2_1Z'_1 + 4\bb{E}Z_1Z'_1 - 1.
\end{eqnarray}
Let $Y_1$ be a standard Gaussian random variable independent of $X_1$. Straightforward algebra leads to the following equality (in distribution)
\begin{equation}\label{Eqd}
\paren{Z_1,Z'_1} = \paren{X^2_1,X'^2_1} \eqd \paren{X^2_1, r^2X^2_1 + \paren{1-r^2} Y^2_1 + 2r\sqrt{1-r^2} X_1Y_1}.
\end{equation}
Eq. \eqref{Eqd} is critical for finding the conditional expected value of $Z'_w$ and $Z'^2_w$ given $Z_w$. Independence of $X_1$ and $Y_1$ implies that
\begin{equation*}
\bb{E}\brac{ Z'_1 \mid Z_1 } = \bb{E}\brac{r^2X^2_1 + \paren{1-r^2} Y^2_1 + 2r\sqrt{1-r^2} X_1Y_1  \mid X_1} = r^2X^2_1 + 1 - r^2 = r^2Z_1 + 1 - r^2.
\end{equation*}
Therefore, 
\begin{align}\label{Lem5.2Eq1}
&\bb{E}Z_1Z'_1 = \bb{E}\brac{Z_1 \bb{E}\brac{ Z'_1 \mid Z_1 } } = \bb{E}\paren{r^2Z^2_1 + \paren{1 - r^2}Z_1} = 3r^2 + 1 - r^2 = 1+2r^2, \nonumber\\
&\bb{E}Z_1Z'_1 = \bb{E}\brac{Z^2_1 \bb{E}\brac{ Z'_1 \mid Z_1 } } = \bb{E}\paren{r^2Z^3_1 + \paren{1 - r^2}Z^2_1} = 15r^2+3\paren{1-r^2} = 3+12r^2.
\end{align}
In addition, using analogous techniques, we get
\begin{eqnarray*}
\bb{E}\brac{ Z'^2_1 \mid Z_1 } &=& \bb{E}\brac{ \paren{r^2X^2_1 + \paren{1-r^2} Y^2_1 + 2r\sqrt{1-r^2} X_1Y_1}^2 \mid X_1 } \\
&=& r^2X^4_1 + \paren{1-r^2}^2 \bb{E}Y^4_1 + 4r^2\paren{1-r^2} X^2_1 \bb{E}Y^2_1 + 2r^2\paren{1-r^2}X^2_1\bb{E}Y^2_1 \\
&+& 4r\paren{1-r^2}^{3/2}X_1\bb{E}Y^3_1 + 4r^3\paren{1-r^2}^{1/2}X^3_1\bb{E}Y_1\\
&=& r^4X^4_1 + 3\paren{1-r^2}^2 + 6r^2\paren{1-r^2}X^2_1 = r^4Z^2_1 + 3\paren{1-r^2}^2 + 6r^2\paren{1-r^2}Z_1.
\end{eqnarray*}
Thus,
\begin{eqnarray}\label{Lem5.2Eq2}
\bb{E}Z^2_1Z'^2_1 &=& \bb{E}\brac{Z^2_1 \bb{E}\brac{ Z'^2_1 \mid Z_1 } } = \bb{E}\brac{Z^2_1\paren{r^4Z^2_1 + 3\paren{1-r^2}^2 + 6r^2\paren{1-r^2}Z_1}} \nonumber\\
&=& 105r^4+9\paren{1-r^2}^2 + 90r^2\paren{1-r^2} = 24r^4 + 72r^2 + 9.
\end{eqnarray}
Substituting Eq. \eqref{Lem5.2Eq1} and \eqref{Lem5.2Eq2} into \eqref{Lem5.2Eq0} yields
\begin{equation}\label{pr1}
p\paren{r,1} = \bb{E}Z^2_1Z'^2_1 - 9 -40r^2 = 24r^4 + 72r^2 + 9 - 9 -40r^2 = 8r^2\paren{4+3r^2}.
\end{equation}
We terminate the proof by replacing Eq. \eqref{pr1} into Eq. \eqref{prw}. As the second claim in the statement of Lemma \ref{Lemma3App} can be substantiated by similar techniques, we skip its proof for avoiding repetitions.
\end{proof}

The succeeding Lemma comes in handy at Section \ref{DtctAlgo} for standardizing $T_t$ under $\bb{H}_{0,t}$ regime. 

\begin{lem}\label{Lemma1App}
Let $Z_w$ be a chi-square random variable with $w$ degrees of freedom. Then the mean and standard deviation of $f\paren{Z_w/w}$ is given by
\begin{equation*}
g_1\paren{w} \coloneqq \bb{E}f\paren{\frac{Z_w}{w}} = \log\paren{\frac{w}{2}} - \psi^{\paren{0}}\paren{\frac{w}{2}}, \quad g_2\paren{w} \coloneqq \sd f\paren{\frac{Z_w}{w}} = \sqrt{\psi^{\paren{1}}\paren{\frac{w}{2}} - \frac{2}{w}}.
\end{equation*}
in which $\psi^{\paren{r}}$ stands for the \emph{poly-gamma function} of order $r$.
\end{lem}

\begin{proof}
It is known that the expected value and variance of $\log Z_w$ are respectively given by $\log 2 + \psi^{\paren{0}}\paren{w/2}$ and $\psi^{\paren{1}}\paren{w/2}$ \cite{pav2015moments}. Therefore,
\begin{equation*}
\bb{E}f\paren{\frac{Z_w}{w}} = \bb{E}\brac{ \frac{Z_w}{w}-1-\log\paren{\frac{Z_w}{w}}  } = 1-1+\log w - \bb{E} \log Z_{w} = \log\paren{\frac{w}{2}} - \psi\paren{\frac{w}{2}}.
\end{equation*}	
We now focus on calculating the variance of $f\paren{\frac{Z_w}{w}}$. Observe that
\begin{eqnarray}\label{varfZ}
\var f\paren{\frac{Z_w}{w}} &=& \var\paren{\frac{Z_w}{w}-\log Z_w} = \frac{\var Z_w}{w^2} + \var \log Z_w - \frac{2\cov\paren{Z_w, \log Z_w}}{w} \nonumber\\
&=& \frac{2}{w} + \psi^{\paren{1}}\paren{\frac{w}{2}} - \frac{2}{w}\cov\paren{Z_w, \log Z_w} \nonumber\\
&=& \frac{2}{w} + \psi^{\paren{1}}\paren{\frac{w}{2}} -\frac{2}{w}\brac{\bb{E}\paren{Z_w\log Z_w}-w\bb{E} \log Z_w }.
\end{eqnarray}
So we just need to evaluate the expected value of $Z_w\log Z_w$. Using the fact that $\Gamma\paren{x+1} = x\Gamma\paren{x}$ for any $x>0$ leads to
\begin{eqnarray*}
\bb{E}\paren{Z_w\log Z_w} &=& \int_{0}^{\infty} x\log x \frac{x^{w/2-1}\exp\paren{x/2}}{2^{w/2}\Gamma\paren{w/2}} dx = \int_{0}^{\infty} \log x \frac{x^{\paren{w+2}/2-1}\exp\paren{x/2}}{2^{w/2}\Gamma\paren{w/2}} dx \\
&=& w \int_{0}^{\infty} \log x \frac{x^{\paren{w+2}/2-1}\exp\paren{x/2}}{2^{\paren{w+2}/2}\Gamma\paren{\paren{w+2}/2}} dx =  w\bb{E} \log Z_{w+2}.
\end{eqnarray*}
Thus the covariance of $Z_w$ and $\log Z_w$ can be rewritten as
\begin{eqnarray}\label{covZlogZ}
\cov\paren{Z_w, \log Z_w} &=& w\paren{ \bb{E} \log Z_{w+2} - \bb{E} \log Z_{w}  } = w\brac{ \psi^{\paren{0}}\paren{\frac{w+2}{2}} + \log 2 - \psi^{\paren{0}}\paren{\frac{w}{2}} - \log 2 } \nonumber\\
&=& w\brac{ \psi^{\paren{0}}\paren{1+\frac{w}{2}} - \psi^{\paren{0}}\paren{\frac{w}{2}} } \RelNum{\paren{a}}{=} w\frac{2}{w} = 2.
\end{eqnarray}
In which the identity $\paren{a}$ is a direct implication of the fact that $\psi^{\paren{0}}\paren{x+1} - \psi^{\paren{0}}\paren{x} = 1/x$. Replacing Eq. \eqref{covZlogZ} into Eq. \eqref{varfZ} yields 
\begin{equation*}
\var f\paren{\frac{Z_w}{w}} = \frac{2}{w} + \psi^{\paren{1}}\paren{\frac{w}{2}} - \frac{4}{w} = \psi^{\paren{1}}\paren{\frac{w}{2}} - \frac{2}{w},
\end{equation*}
which concludes the proof.
\end{proof}

The following result, which is needed for proving Theorem \ref{CLTforT}, shows that all moments of $wf\paren{\frac{Z_w}{w}}$ are bounded

\begin{lem}\label{Lem4App}
There exist universal constant $C_1,C_2>0$ such that 
\begin{align*}
&\norm{wf\paren{\frac{Z_w}{w}}}{k}  = w\abs{\bb{E} f^k\paren{\frac{Z_w}{w}}}^{1/k}\leq C_1k,\\
&\norm{w\brac{f\paren{\frac{Z_w}{w}} - g_1\paren{w}}}{k} \leq C_2k, \quad \forall\; k\in\bb{N}.
\end{align*}
\end{lem}

\begin{proof}
Let $G_w\paren{t}$ denotes the moments generating function of $f\paren{Z_w/w}$. Due to non-negativity of $f$, if $G_w$ is well defined for a strictly positive $t$, then all moments of $f\paren{Z_w/w}$ can be controlled from above by $G_w\paren{t}$. We first show that $G_w$ is well defined at $t = w/4$. Observe that
\begin{eqnarray*}
G_w\paren{\frac{w}{4}} &=& \bb{E}\brac{\exp\paren{\frac{w}{4}f\paren{\frac{Z_w}{w}} }} = \bb{E}\brac{ \exp\paren{ \frac{Z_w}{4} - \frac{w}{4} - \frac{w}{4}\log\paren{\frac{Z_w}{w}} }} = e^{-\frac{w}{4}} w^{\frac{w}{4}}\bb{E}\brac{ Z^{-\frac{w}{4}}_w \exp\paren{\frac{Z_w}{4}} }\\
&=& e^{-\frac{w}{4}} w^{\frac{w}{4}} \int_{0}^{\infty} x^{-\frac{w}{4}} e^{\frac{x}{4}} \frac{x^{\frac{w}{2}-1} e^{-\frac{x}{2}} }{2^{\frac{w}{2}}\Gamma\paren{\frac{w}{2}} } dx = e^{-\frac{w}{4}} w^{\frac{w}{4}} \int_{0}^{\infty}  \frac{x^{\frac{w}{4}-1} e^{-\frac{x}{4}} }{2^{\frac{w}{2}}\Gamma\paren{\frac{w}{2}} } dx.
\end{eqnarray*}
The last integral can be simplified by introducing $u\coloneqq \frac{x}{4}$.
\begin{equation}\label{G_w}
G_w\paren{\frac{w}{4}} = e^{-\frac{w}{4}} w^{\frac{w}{4}} \int_{0}^{\infty}  \frac{u^{\frac{w}{4}-1} e^{-u} }{\Gamma\paren{\frac{w}{2}} } du 
= \exp\brac{ -\frac{w}{4} + \frac{w}{4}\log w + \log \Gamma\paren{\frac{w}{4}} - \log \Gamma\paren{\frac{w}{2}} }.
\end{equation}
This identity ensures that $G_w\paren{\frac{w}{4}}$ is well defined as long as $w$ does not grow. Strictly speaking we can find bounded $w_0\in\bb{N}$ nd $C_0>0$ (depending on $w_0$) so that $G_w\paren{w/4}$ is smaller than $C_0$ for all $w\leq w_0$. Next we study the behaviour of $G_w\paren{\frac{w}{4}}$ for large $w$. It is known that \footnote{See \url{http://functions.wolfram.com/GammaBetaErf/LogGamma/introductions/Gammas/ShowAll.html}} 
\begin{equation*}
\lim\limits_{x\rightarrow\infty}\log \Gamma\paren{x} - \paren{x-\frac{1}{2}}\log x + x = \frac{\log \paren{2\pi}}{2}.
\end{equation*}
Applying the asymptotic identity of $\log\Gamma$ function on Eq. \eqref{G_w} implies that
\begin{equation*}
G_w\paren{\frac{w}{4}} = \exp\paren{\frac{1}{2} \log 2} = \sqrt{2}.
\end{equation*}
In summary we guaranteed the existence of a bounded universal constant $C_1\geq 1$ for which $G_w\paren{\frac{w}{4}} \leq C_1$, for all $w\in\bb{N}$. Elementary properties of moment generating function implies that
\begin{equation*}
\frac{1}{k!}\paren{\frac{w}{4}}^k \norm{f\paren{\frac{Z_w}{w}}}{k}^k \leq G_w\paren{\frac{w}{4}},\quad\forall\; k\in\bb{N}\quad \Longrightarrow \quad \norm{\frac{w}{4}f\paren{\frac{Z_w}{w}}}{k} \leq \sqrt[k]{k! G_w\paren{\frac{w}{4}}} \leq \sqrt[k]{k!}C^{\frac{1}{k}}_1,\quad\forall\; k\in\bb{N}.
\end{equation*}
We conclude the proof by mentioning that $\sqrt[k]{k!} \leq k$ for any $k\in\bb{N}$. Now we turn to the proof of second claim. Using Minkowski's inequality, we get
\begin{equation*}
\norm{w\brac{f\paren{\frac{Z_w}{w}} - g_1\paren{w}}}{k} \leq \norm{wf\paren{\frac{Z_w}{w}}}{k} + \abs{wg_1\paren{w}}\leq Ck + \abs{wg_1\paren{w}}.
\end{equation*}
So the desired result obviously holds, since there is a bounded universal scalar $C'>0$ such that $\abs{wg_1\paren{w}}\leq C'$ for all $w\in\bb{N}$ \footnote{We refer the reader to $5.11.2$ of \url{https://dlmf.nist.gov/5.11}}. 
\end{proof}

We now find a polynomial approximation for $h_w$ function defined in Eq. \eqref{h_w}.

\begin{lem}\label{Lemma2App}
Under the same notation and conditions as Lemma \ref{Lemma3App}, define $h_w\brac{-1,1}\mapsto \brac{-1,1}$ by
\begin{equation*}
h_w\paren{r} \coloneqq \corr\brac{ f\paren{\frac{Z_w}{w}}, f\paren{\frac{Z'_w}{w}} }
\end{equation*}
Then 
\begin{enumerate}
\item $h_w$ is an even function, i.e. $h_w\paren{x} = h_w\paren{-x}$ for any $x\in\brac{0,1}$,  with $h_w\paren{0} = 0$ and $h_w\paren{1} = 1$.
\item There exist a bounded universal scalar $C > 0$and $w_0\in\bb{N}$ such that for any $w\geq w_0$
\begin{equation*}
\max_{r\in\brac{-1,1}} \abs{ h_w\paren{r} - r^4 }\leq \frac{C}{w}.
\end{equation*}
\end{enumerate}
\end{lem}

\begin{proof}
The first claim in obvious. So we focus on finding a uniform upper bound on the difference between $h_w\paren{r}$ and $r^4$, for $r\in\brac{-1,1}$. Choose an arbitrary $r\in\brac{-1,1}$. Define random variables $\bar{Z}_w$ and $\bar{Z}'_w$ by
\begin{equation}
\bar{Z}_w\coloneqq \paren{Z_w-w},\quad \bar{Z}'_w\coloneqq \paren{Z'_w-w}.
\end{equation}
Taylor expansion of $f$ around $1$ gives us
\begin{equation*}
f\paren{\frac{Z_w}{w}} = f\paren{1+ \frac{\bar{Z}_w}{w} } = \sum_{k=2}^{\infty} \frac{\paren{-1}^k \bar{Z}^k_w}{kw^k} .
\end{equation*}
Therefore, 
\begin{eqnarray}\label{TaylorExp}
\cov\brac{f\paren{\frac{Z_w}{w}}, f\paren{\frac{Z'_w}{w}}} &=& \cov\Bigbrac{ \sum_{k=2}^{\infty} \frac{\paren{-1}^k \bar{Z}^k_w}{kw^k}, \sum_{k=2}^{\infty} \frac{\paren{-1}^k \bar{Z}'^k_w}{kw^k} } = \sum_{j,k=2}^{\infty}\frac{\paren{-1}^{j+k} }{jk w^{j+k}} \cov\paren{\bar{Z}^j_w, \bar{Z}'^k_w} \nonumber\\
&=& \frac{\cov\paren{\bar{Z}^2_w, \bar{Z}'^2_w}}{4w^4} - \frac{\cov\paren{\bar{Z}^3_w, \bar{Z}'^2_w}}{3w^5} + \sum_{j+k\geq 6}^{\infty}\frac{\paren{-1}^{j+k} }{jk w^{j+k}} \cov\paren{\bar{Z}^j_w, \bar{Z}'^k_w}.
\end{eqnarray}
Lemma \ref{Lemma3App} gives an equivalent representation for the first two terms in the second line of Eq. \eqref{TaylorExp}. One line of algebra guarantees the existence of a large enough constant $C_1$ such that
\begin{equation}\label{Ineq1}
\abs{ \frac{\cov\paren{\bar{Z}^2_w, \bar{Z}'^2_w}}{4w^4} - \frac{\cov\paren{\bar{Z}^3_w, \bar{Z}'^2_w}}{3w^5} - \frac{2r^4}{w^2}} \leq \frac{C_1}{w^3},\quad \forall\;w\geq 1. 
\end{equation}
We also use $\xi$ to denote the third term in the second line of Eq. \eqref{TaylorExp}. Observe that
\begin{equation*}
\abs{\xi} = w^{-3} \abs{\sum_{j+k\geq 6}^{\infty}\frac{\paren{-1}^{j+k}}{jk} \paren{\frac{2}{w}}^{\paren{j+k-6}/2} \cov\brac{ \paren{\frac{\bar{Z}_w}{\sqrt{2w}}}^j, \paren{\frac{\bar{Z}'_w}{\sqrt{2w}}}^k }}
\end{equation*}
We know from central limit Theorem that as $w$ tends to infinity, then 
\begin{equation*}
\paren{\frac{\bar{Z}_w}{\sqrt{2w}}, \frac{\bar{Z}'_w}{\sqrt{2w}}} \cp{d} \cc{N}\paren{\brac{\begin{matrix} 0\\ 0 \end{matrix}}, \brac{\begin{matrix} 1&r^2\\ r^2&1 \end{matrix}} }.
\end{equation*}
Thus, there exist a large enough $w_0\in\bb{N}$ and finite positive scalar $C_2$ such that for any $w\geq w'_0$
\begin{equation}\label{Ineq2}
\abs{\xi}\leq \frac{C_2}{w^3}.
\end{equation} 
Combining Eq. \eqref{TaylorExp}, \eqref{Ineq1} and \eqref{Ineq2} gives us
\begin{equation*}
\abs{\cov\brac{f\paren{\frac{Z_w}{w}}, f\paren{\frac{Z'_w}{w}}} - \frac{2r^4}{w^2}} \leq \frac{C_1+C_2}{w^3},\quad \forall\; w\geq w'_0.
\end{equation*}
According to Lemma \ref{Lemma1App}, $\var\brac{f\paren{\frac{Z_w}{w}}} = \var\brac{f\paren{\frac{Z'_w}{w}}} = \psi^{\paren{1}}\paren{\frac{w}{2}} - \frac{2}{w}$. So
\begin{equation}\label{Ineq3}
\abs{h_w\paren{r} - \frac{2r^4}{w^2}\paren{\psi^{\paren{1}}\paren{\frac{w}{2}} - \frac{2}{w}}^{-1} }\leq \frac{C_1+C_2}{w^3} \paren{\psi^{\paren{1}}\paren{\frac{w}{2}} - \frac{2}{w}}^{-1}.
\end{equation}
It is known that\footnote{See 5.15.8 at \url{https://dlmf.nist.gov/5.15}} there are two finite scalars $C'_1$ and $w''_0$ such that 
\begin{equation*}
\abs{\psi^{\paren{1}}\paren{\frac{w}{2}} - \frac{2}{w} - \frac{2}{w^2}} \leq \frac{C'_1}{w^3},\quad \forall\;w\geq w''_0.
\end{equation*}
So inequality \eqref{Ineq3} can be simplified as
\begin{equation*}
\abs{ h_w\paren{r} - r^4 } \leq \frac{C_1+C_2}{2w-C'_1} + \frac{2C'_1r^4}{w} \leq \frac{C}{w}.
\end{equation*}
Here $C$ is a finite scalar depending on $C_1,C_2, C'_1, w'_0$ and $w''_0$. In summary we showed that for any $r\in\brac{-1,1}$,
\begin{equation*}
\abs{ h_w\paren{r} - r^4 } \leq \frac{C}{w},\quad \forall \;w\geq w_0\coloneqq w'_0 \vee w''_0.
\end{equation*}
\end{proof}

\begin{lem}\label{Lem5App}
There exists a bounded constant $C_w > 1$ such that $\sup_{r\in\brac{-1,1}}\abs{\frac{h_w\paren{r}}{r^4}-1} \leq C_w$.
\end{lem}

\begin{proof}
Since $h_w$ is a continuous function on a compact space, we only need to show that $\limsup\limits_{r\rightarrow 0} \frac{h_w\paren{r}}{r^4} \leq C'_w \coloneqq 1+C_w$. According to Lemma \ref{Lem4App}, $h_w\paren{r} \asymp r^4$ as $w\rightarrow \infty$, so we only present the proof for $w=1$.
\end{proof}

The following result is beneficiary for finding the expected value and standard deviation of $T_t$, under alternative hypothesis $\bb{H}_{1,t}$.

\begin{lem}\label{Lemma4App}
Let $\paren{X,X'}$ be a standard bi-variate Gaussian random vector with $\corr\paren{X, X'} = r$. Let $\set{\paren{X_k, X'_k}}^w_{k=1}$ be i.i.d. draws from $\paren{X,X'}$ and define
\begin{equation*}
Z_w \coloneqq \sum_{k=1}^{w} X^2_k,\quad Z'_w \coloneqq \sum_{k=1}^{w} X'^2_k.
\end{equation*}
Recall $g_1\paren{\cdot}$ and $g_2\paren{\cdot}$ from Lemma \ref{Lemma1App}. For two arbitrary positive scalars $\alpha,\alpha'$,
\begin{enumerate}
\item $\bb{E}f\paren{\frac{\alpha Z_w}{w}} = g_1\paren{w} + f\paren{\alpha}$.
\item $\cov\brac{ f\paren{\frac{\alpha Z_w}{w}}, f\paren{\frac{\alpha'Z'_w}{w}} } = \frac{2r^2}{w}\paren{\alpha-1}\paren{\alpha'-1} + h_w\paren{r}g^2_2\paren{w}$.
\end{enumerate}
\end{lem}

\begin{proof}
We employ a similar approach as the proof of Lemma \ref{Lemma1App} and \ref{Lemma2App}. Notice that 
\begin{equation}\label{Ident1}
f\paren{\frac{\alpha Z_w}{w}} = f\paren{\frac{Z_w}{w}} +\paren{\alpha-1}\frac{Z_w}{w} - \log \alpha.
\end{equation}
taking expected value from both sides substantiates the first claim. 
\begin{equation*}
\bb{E}f\paren{\frac{\alpha Z_w}{w}} = \bb{E}f\paren{\frac{Z_w}{w}} + \alpha-1-\log \alpha = \bb{E}f\paren{\frac{Z_w}{w}} + f\paren{\alpha} = g_1\paren{w} + f\paren{\alpha}.
\end{equation*}
We now aim to prove the second claim. $h_w$ is defined to satisfy the following identity.
\begin{equation*}
\cov\brac{ f\paren{\frac{Z_w}{w}}, f\paren{\frac{Z'_w}{w}} } = h_w\paren{r}g^2_2\paren{w}.
\end{equation*}
Since covariance operator is bilinear, using Eq. \eqref{Ident1} one can show that
\begin{eqnarray}\label{Ident2}
\cov\brac{ f\paren{\frac{\alpha Z_w}{w}}, f\paren{\frac{\alpha'Z'_w}{w}} } - h_w\paren{r}g^2_2\paren{w}&=&  \paren{\alpha\alpha'-1}\cov\paren{\frac{Z_w}{w},\frac{Z'_w}{w}} - \paren{\alpha-1}\cov\paren{\frac{Z_w}{w},\log \frac{Z'_w}{w}} \nonumber\\
&-& \paren{\alpha'-1}\cov\paren{\frac{Z'_w}{w},\log \frac{Z_w}{w}}.
\end{eqnarray}
Let $\set{Y_k}^w_{k=1}$ be a set of i.i.d. standard Gaussian random variables. Obviously
\begin{equation*}
\paren{Z_w,Z'_w} = \paren{Z_w, r^2Z_w, \paren{1-r^2}\sum_{k=1}^{w}Y^2_k + 2r\sqrt{1-r^2}\sum_{k=1}^{w}X_kY_k}.
\end{equation*}
This distributional identity alongside with Lemma \ref{Lemma0App} shows that
\begin{align*}
&\cov\paren{Z_w, Z'_w} = r^2 \var Z_w = 2r^2w,\\
& \cov\paren{\frac{Z_w}{w},\frac{Z'_w}{w}} = \cov\paren{\frac{Z'_w}{w},\log \frac{Z_w}{w}} = r^2\cov\paren{\frac{Z_w}{w},\log\frac{Z_w}{w}}.
\end{align*}
So we can rewrite Eq. \eqref{Ident2} as the following.
\begin{equation*}
\cov\brac{ f\paren{\frac{\alpha Z_w}{w}}, f\paren{\frac{\alpha'Z'_w}{w}} } - h_w\paren{r}g^2_2\paren{w} = \frac{2r^2}{w}\paren{\alpha\alpha'-1} - \frac{r^2\paren{\alpha+\alpha'-2}}{w}\cov\paren{Z_w,\log Z_w}.
\end{equation*}
We conclude the proof by applying Eq. \eqref{covZlogZ}.
\begin{equation*}
\cov\brac{ f\paren{\frac{\alpha Z_w}{w}}, f\paren{\frac{\alpha'Z'_w}{w}} } - h_w\paren{r}g^2_2\paren{w}=\frac{2r^2}{w}\paren{\alpha\alpha'-1} - \frac{2r^2}{w}\paren{\alpha+\alpha'-2} = \frac{2r^2}{w}\paren{\alpha-1}\paren{\alpha'-1}.
\end{equation*}
\end{proof}

\section{Some technical probabilistic results}\label{AppendixB}

This section contains auxiliary results that are useful for proving main theoretical contributions of this manuscript in Section \ref{Proofs}. The following Proposition comes in handy for substantiating Theorems \ref{CLTforT} and \ref{CLTforTHat}.

\begin{prop}\label{IndepGauss}
Let $Z\in\bb{R}^p$ be a zero mean GGM with inverse covariance matrix $\Omega$. Define another (zero mean) GGM $Y\in\bb{R}^p$ by
\begin{equation*}
Y_s = \frac{\InnerProd{\Omega_{s,:}}{Z}}{\sqrt{\Omega_{ss}}},\quad\forall\;s\in\set{1,\ldots,p}.
\end{equation*}
Choose four distinct points $s_1,t_1,s_2,t_2\in\set{1,\ldots,p}$ such that there is no edge between $B_1 = \set{s_1,t_1}$ and $B_2 = \set{s_2,t_2}$. Then $\brac{Y_{s_1},Y_{t_1}}$ is independent of $\brac{Y_{s_2},Y_{t_2}}$.
\end{prop}
\begin{proof}
Let $R$ be the partial correlation matrix of $Z$. It is easy to show that $\cov\paren{Y} = R$. Thus $\brac{Y_{s_1},Y_{t_1}}$ is independent of $\brac{Y_{s_2},Y_{t_2}}$ if and only if
\begin{equation*}
\cov\paren{\brac{Y_{s_1},Y_{t_1}}, \brac{Y_{s_2},Y_{t_2}} } = R_{B_1,B_2} = \zero_{2\times 2}.
\end{equation*}
Since there is no edge between $\set{s_1,t_1}$ and $\set{s_2,t_2}$, then $\Omega_{B_1,B_2} = \zero_{2\times 2}$. We conclude the proof by reminding the reader that $R$ has the same sparsity pattern as $\Omega$, i.e. $\supp\paren{R} = \supp\paren{\Omega}$. 
\end{proof}

Next we conduct a probabilistic sensitivity analysis for $f$ with a stochastic argument (Chi-square random variable with $w$-degrees of freedom). Such result is crucial in the proof of Theorem \ref{CLTforTHat}.

\comment{\begin{prop}\label{ChiSqConIneq2}
For an arbitrary $p>1$, let $Z_s,\;s=1,\ldots,p$ be a set of identically distributed chi-square random variables with $w$ degrees of freedom. Then
\begin{equation}\label{EqChiSqIneq2}
\bb{P}\brac{ \max_{s=1,\ldots,p} \frac{Z_s}{w} \geq 1-2\paren{\frac{t+\log p}{w} + \sqrt{\frac{t+\log p}{w}}} } \leq e^{-t}.
\end{equation}
Furthermore, if $w\geq 24\log p$, then
\begin{equation*}
\bb{P}\paren{ \max_{s=1,\ldots,p} \frac{Z_s}{w} \geq \frac{1}{4} } \leq \frac{1}{p}.
\end{equation*}
\end{prop}

\begin{proof}
The Eq. \eqref{EqChiSqIneq2} is an immediate consequence of Remark $2.11$ of \cite{boucheron2013concentration}. For the second part, just replace $t = \log p$ into Eq. \eqref{EqChiSqIneq}.
\end{proof}}

\begin{lem}\label{SensAnalfLemma}
Let $\alpha$ be a bounded strictly positive scalar and let $Z_w$ be a chi-square random variables with $w$ degrees of freedom. Then for any $\xi >0$,
\begin{equation*}
\bb{P}\brac{\abs{ f\paren{\frac{\alpha Z_w}{w}} - f\paren{\frac{Z_w}{w}} -f\paren{\alpha} } \geq \abs{\alpha-1}\sqrt{\frac{8\xi\log p}{w}}\paren{ 1 \vee \sqrt{\frac{8\xi\log p}{w}} } } \leq p^{-2\xi}.
\end{equation*}
\end{lem}

\begin{proof}
Without loss of generality, assume that $\alpha\ne 1$. The following identity holds for any $x>0$.
\begin{equation*}
f\paren{\alpha x} - f\paren{x} - f\paren{\alpha} = \brac{\alpha x - 1 -\log\paren{\alpha x}} - \paren{x - 1 -\log x} - \paren{\alpha - 1 - \log \alpha } = \paren{\alpha-1}\paren{x-1}.
\end{equation*}
Thus
\begin{equation*}
\frac{\sqrt{\frac{w}{2}}}{\abs{\alpha-1}} \abs{ f\paren{\frac{\alpha Z_w}{w}} - f\paren{\frac{Z_w}{w}} -f\paren{\alpha} } = \frac{\abs{Z_w - w}}{\sqrt{2w}}.
\end{equation*}
Therefore we just need a concentration inequality for a standardized version of $Z_w$. We know form Remark $2.11$ of \cite{boucheron2013concentration} that for any $t>0$
\begin{equation}\label{EqChiSqIneq}
\bb{P}\paren{ \frac{\abs{Z_w - w}}{\sqrt{2w}} \geq \sqrt{2t} + \sqrt{\frac{2}{w}}t }\leq e^{-t}.
\end{equation}
Replacing $t = 2\xi\log p$ in Eq. \eqref{EqChiSqIneq} concludes the proof.
\end{proof}

One can extend Lemma \ref{SensAnalfLemma} to random $\alpha$, as long as it is independent from $Z_w$. We skip the proof due to its simplicity.
\begin{cor}\label{SensAnalfCor}
Lemma \ref{SensAnalfLemma} is satisfied for a strictly positive random variable $\alpha$, independent of $Z_w$.
\end{cor}

\comment{\begin{prop}\label{HansonWrightIneq}
Let $\set{Z_s}^w_{s=1}$ be a set of independent $p-$dimensional standard Gaussian vectors. The following inequality holds for any deterministic matrix $A\in\bb{R}^{p\times p}$ and any $\xi\in\paren{0,\infty}$.
\begin{equation*}
\bb{P}\brac{ \abs{\frac{1}{w}\sum_{r=1}^w Z^\top_r A Z_r - \tr\paren{A}} \geq \max\paren{ \frac{8\OpNorm{A}{2}{2}\xi\log p}{w}, \LpNorm{A}{2}\sqrt{\frac{8\xi\log p}{w}} } } \leq 2p^{-\xi}.
\end{equation*}
\end{prop}

\begin{proof}
Consider the following two events:
\begin{align*}
& E_1 \coloneqq \brac{ \frac{1}{w}\sum_{r=1}^w Z^\top_r A Z_r - \tr\paren{A} \geq \max\paren{ \frac{8\OpNorm{A}{2}{2}\xi\log p}{w}, \LpNorm{A}{2}\sqrt{\frac{8\xi\log p}{w}} } },\\
& E_2 \coloneqq \brac{ \frac{1}{w}\sum_{r=1}^w Z^\top_r A Z_r - \tr\paren{A} \leq -\max\paren{ \frac{8\OpNorm{A}{2}{2}\xi\log p}{w}, \LpNorm{A}{2}\sqrt{\frac{8\xi\log p}{w}} } }.
\end{align*} 
$E_2$ is obviously same probable as $E_1$ (substitute $A$ by $-A$ in $E_1$). So, it suffices to show that the occurrence probability of $E_1$ is bounded above by $p^{\xi}$. Our proof is built upon well known \emph{Chernoff bound}. Choose strictly positive $\zeta$ arbitrarily. Furthermore define $\cc{C} \coloneqq \brac{0,\frac{w}{4\OpNorm{A}{2}{2}}}$. It is easy to verify that for any $\gamma\in\cc{C}$,
\begin{equation*}
\log \bb{E}\exp\paren{\frac{\gamma}{w}Z^\top_r A Z_r } = -\frac{1}{2}\log\det \paren{I_p-\frac{2\gamma}{w}A} < \infty,\quad\forall\;r=1,\ldots,w.
\end{equation*}
Observe that
\begin{eqnarray*}
\log\bb{P}\paren{\frac{1}{w}\sum_{r=1}^w Z^\top_r A Z_r \geq \zeta + \tr\paren{A}} &\leq& \min_{\gamma\in\cc{C}} \brac{-\gamma\paren{\tr\paren{A} +\zeta} + \log \bb{E}\exp\paren{ \frac{\gamma}{w}\sum_{r=1}^w Z^\top_r A Z_r }}\\
&=& \min_{\gamma\in\cc{C}} \brac{-\gamma\paren{\tr\paren{A} +\zeta} + \sum_{r=1}^{w}\log \bb{E}\exp\paren{ \frac{\gamma}{w}w Z^\top_r A Z_r }}\\
&=&\min_{\gamma\in\cc{C}} \brac{-\gamma\paren{\tr\paren{A} +\zeta} -\frac{w}{2}\log\det \paren{I_p-\frac{2\gamma}{w}A}}\\
&=& \min_{\gamma\in\cc{C}} \set{-\gamma\zeta - \gamma\sum_{i=1}^{p} \brac{\lambda_i\paren{A} + \frac{w}{2\gamma}\log\paren{1-\frac{2\gamma\lambda_i\paren{A}}{w} }} }\\
&\RelNum{\paren{a}}{\leq}& \min_{\gamma\in\cc{C}} \brac{ -\gamma\zeta + \frac{2\gamma^2}{w}\sum_{i=1}^{p} \lambda^2_i\paren{A} } = \min_{\gamma\in\cc{C}} \paren{ -\gamma\zeta + \frac{2\gamma^2}{w}\LpNorm{A}{2}^2 }.
\end{eqnarray*}
The inequality $\paren{a}$ is implied form the following result ((its proof is left to the reader).
\begin{equation*}
x + a\log\paren{1-\frac{x}{a}} \geq -\frac{x^2}{a},\quad \forall\; a>0,\mbox{and}\;x<\frac{a}{2}.
\end{equation*}
So we showed that
\begin{equation}\label{Ineq1PropB3}
\log\bb{P}\paren{\frac{1}{w}\sum_{r=1}^w Z^\top_r A Z_r \geq \zeta + \tr\paren{A}} \leq  \min_{\gamma\in\cc{C}} \paren{ -\gamma\zeta + \frac{2\gamma^2}{w}\LpNorm{A}{2}^2 }.
\end{equation}
Define $H:\brapar{0,\infty}\mapsto\bb{R}$ by
\begin{equation*}
H\paren{\eta} = -\gamma\zeta + \frac{2\gamma^2}{w}\LpNorm{A}{2}^2.
\end{equation*}
In order to find the optimal upper bound on $\bb{P}E_1$, we need to find the global minimizer of $H$ on $\cc{C}$. For doing so, we consider two mutually exclusive scenarios regarding $\zeta$. 
\begin{enumerate}[label = (\alph*),leftmargin=*]
\item $\zeta \leq \LpNorm{A}{2}^2/\OpNorm{A}{2}{2}$.
\item $\zeta > \LpNorm{A}{2}^2/\OpNorm{A}{2}{2}$.
\end{enumerate}
In case $\paren{a}$, the single root of $H'$ ($H'$ denotes the first derivative of $H$ with respect to $\zeta$) belongs in $\cc{C}$. So we can prove in one line of algebra that
\begin{equation*}
\bb{P}\paren{\frac{1}{w}\sum_{r=1}^w Z^\top_r A Z_r \geq \zeta + \tr\paren{A}} \leq \exp\paren{-\frac{w\zeta^2}{8\LpNorm{A}{2}^2}}.
\end{equation*}
If the condition $\paren{b}$ holds, then $H$ attains its global minimizer at $\hat{\zeta} = \frac{w}{4\OpNorm{A}{2}{2}}$ (right boundary point of $\cc{C}$). Therefore, one can easily show that
\begin{equation*}
\bb{P}\paren{\frac{1}{w}\sum_{r=1}^w Z^\top_r A Z_r \geq \zeta + \tr\paren{A}} \leq \exp\brac{-\frac{w\zeta}{8\OpNorm{A}{2}{2}}\paren{2- \frac{\LpNorm{A}{2}^2}{\zeta \OpNorm{A}{2}{2}}}} \leq \exp\paren{-\frac{w\zeta}{8\OpNorm{A}{2}{2}}}.
\end{equation*}
Combining the last two inequalities yields 
\begin{equation*}
\bb{P}\paren{\frac{1}{w}\sum_{r=1}^w Z^\top_r A Z_r -\tr\paren{A} \geq \zeta }\leq \exp\brac{-\frac{w}{8} \min\paren{ \frac{\zeta}{\OpNorm{A}{2}{2}}, \frac{\zeta^2}{\LpNorm{A}{2}^2} } }.
\end{equation*}
Choosing $\zeta$ such that $\frac{w}{8} \min\paren{ \frac{\zeta}{\OpNorm{A}{2}{2}}, \frac{\zeta^2}{\LpNorm{A}{2}^2} } = \xi\log p$ leads to the desired upper bound on $\bb{P}E_1$.
\end{proof}}

\begin{prop}\label{CondProbProp}
Let $X$ and $Y$ be random variables jointly distributed as $\bb{P}$. Let $B$ be a measurable set with $\bb{P}\paren{Y\in B} \geq 1-\epsilon_1$, for some $\epsilon_1\in\paren{0,1}$. If there exist a measurable set $A$ and $\epsilon_2\in\paren{0,1}$ such that
\begin{equation}\label{CondPABound}
\bb{P}\paren{A\mid Y=y}\leq \epsilon_2,\quad\forall\;y\in B,
\end{equation}
then $\bb{P}\paren{A}\leq \epsilon_1+\epsilon_2$
\end{prop}

\begin{proof}
Let $\bb{P}_Y$ stands for the marginal distribution of $Y$. Observe that
\begin{eqnarray*}
\bb{P}\paren{A} &=& \bb{P}\paren{A\cap \brac{Y\notin B}} + \bb{P}\paren{A\cap \brac{Y\in B}} \leq \bb{P}\paren{Y\notin B} + \bb{P}\paren{A\cap \brac{Y\in B}}\leq \epsilon_1 + \bb{P}\paren{A\cap \brac{Y\in B}}\\
&=&\epsilon_1 + \int_{B} \bb{P}\paren{A\mid Y = y} d\bb{P}_Y\paren{y} \leq \epsilon_1 + \sup_{y\in B} \bb{P}\paren{A\mid Y = y}.
\end{eqnarray*}
Finally condition \eqref{CondPABound} trivially substantiates the desired upper bound on $\bb{P}\paren{A}$.
\end{proof}

\comment{\begin{lem}\label{Lem1AppB}
Let $a,b\in\bb{R}^p$ be two column vectors such that $\LpNorm{a}{2}, \LpNorm{b}{2} \leq M$ for some $M<\infty$. Then,
\begin{equation*}
\OpNorm{bb^\top - aa^\top}{2}{2} \leq 2M\paren{\LpNorm{b-a}{2} \wedge \LpNorm{b+a}{2}}.
\end{equation*}
\end{lem}

\begin{proof}
It is easy to verify that
\begin{eqnarray*}
\OpNorm{bb^\top - aa^\top}{2}{2}  &=& \max_{ \LpNorm{v}{2} = 1 } v^\top \paren{bb^\top - aa^\top} v = \max_{ \LpNorm{v}{2} = 1 } \InnerProd{v}{b}^2 - \InnerProd{v}{a}^2 = \max_{ \LpNorm{v}{2} = 1 } \InnerProd{v}{b-a} \InnerProd{v}{b+a}\\
&\leq& \LpNorm{b-a}{2} \LpNorm{b+a}{2}.
\end{eqnarray*}
According to the triangle inequality, both terms $\LpNorm{b-a}{2}$ and $\LpNorm{b+a}{2}$ are bounded above by $2M$. Thus,
\begin{equation*}
\LpNorm{b-a}{2} \LpNorm{b+a}{2} \leq 2M\LpNorm{b-a}{2},\quad\mbox{and}\quad \LpNorm{b-a}{2} \LpNorm{b+a}{2} \leq 2M\LpNorm{b+a}{2},
\end{equation*}
which concludes the proof
\end{proof}

\begin{cor}
Let $\set{Z_s}^w_{s=1}$ be independent $p-$dimensional standard Gaussian vectors. Let $a,b\in\bb{R}^p$ be two random column vectors independent of $\set{Z_s}^w_{s=1}$ such that
\begin{equation*}
\bb{P}\paren{ \LpNorm{a}{2} \vee \LpNorm{b}{2} \geq M_{\varepsilon} } \leq \varepsilon,
\end{equation*}
for some $\varepsilon\in\paren{0,1}$ and bounded strictly positive $M_{\varepsilon}$. Suppose that $p\geq 2$ and $w\geq 8\log p$. Then
\begin{equation*}
\bb{P}\brac{ \frac{1}{w}\abs{ \sum_{r=1}^{w} \InnerProd{Z_r}{b}^2 - \InnerProd{Z_r}{a}^2 } \geq 6M_{\varepsilon}\LpNorm{b-a}{2} } \leq \frac{2}{p^2} + \varepsilon.
\end{equation*}
\end{cor}

\begin{proof}
For simplicity define random variable $\psi_w\paren{a,b} $ and event $B$ as the following:
\begin{align*}
& \psi_w\paren{a,b} \coloneqq w^{-1}\sum_{r=1}^{w}\paren{\InnerProd{Z_r}{b}^2 - \InnerProd{Z_r}{a}^2},\\
& B \coloneqq \brac{\paren{\omega_1, \omega_2}:\; \LpNorm{a\paren{\omega_1}}{2}, \LpNorm{b\paren{\omega_2}}{2} \leq M_{\varepsilon}}.
\end{align*}
Notice that $B$ occurs with probability at least $1-\varepsilon$. Using Lemma \ref{Lem1AppB}, we can show that for all $\paren{\omega_1, \omega_2}\in B$,
\begin{align*}
&\OpNorm{b\paren{\omega_2}b^\top\paren{\omega_2}-a\paren{\omega_1}a^\top\paren{\omega_1}}{2}{2}\leq 2M_{\varepsilon}\LpNorm{b\paren{\omega_2}-a\paren{\omega_1}}{2},\\ &\LpNorm{b\paren{\omega_2}b^\top\paren{\omega_2}-a\paren{\omega_1}a^\top\paren{\omega_1}}{2}\leq M_{\varepsilon}\sqrt{8}\LpNorm{b\paren{\omega_2}-a\paren{\omega_1}}{2}.
\end{align*} 
Let us introduce an alternative formulation for $\psi_w\paren{a,b}$, which is more suitable for rest of the proof. It is easy to verify that
\begin{equation*}
\psi_w\paren{a,b} = \frac{1}{w} \sum_{r=1}^{w} Z^\top_r \paren{bb^\top-aa^\top} Z_r.
\end{equation*}
Given $a$ and $b$, the expected value of $\psi_w\paren{a,b}$ satisfies the following inequality.
\begin{eqnarray*}
\abs{\bb{E}\brac{\psi_w\paren{a,b} \mid a,b}} &=& \frac{1}{w}\abs{\sum_{r=1}^{w} \bb{E}\InnerProd{Z_rZ^\top_r}{bb^\top-aa^\top} } = \abs{ \tr\paren{bb^\top-aa^\top}  } = \abs{ \LpNorm{b}{2}^2-\LpNorm{a}{2}^2 }\\
&\leq& 2\LpNorm{b-a}{2}\paren{\LpNorm{a}{2} \vee \LpNorm{b}{2}}.
\end{eqnarray*}
So for any $\paren{\omega_1, \omega_2}\in B$
\begin{equation*}
\bb{P}\paren{ \Bigl| \bb{E}\brac{\psi_w\paren{a,b} \mid a = a\paren{\omega_1},b = b\paren{\omega_2}} \Bigr| \leq 2M_{\varepsilon}\LpNorm{b\paren{\omega_2}-a\paren{\omega_1}}{2} } \geq 1-\varepsilon.
\end{equation*}
We now have all the required ingredients for using Proposition \ref{HansonWrightIneq}. Applying straightforward algebra, we can show that  for all $\paren{\omega_1, \omega_2}\in B$
\begin{equation*}
\bb{P}\brac{\abs{\psi_w\Bigparen{ a\paren{\omega_1}, b\paren{\omega_2} }} \geq 2M_{\varepsilon}\LpNorm{b\paren{\omega_2}-a\paren{\omega_1}}{2} + 2M_{\varepsilon}\LpNorm{b\paren{\omega_2}-a\paren{\omega_1}}{2}\max\paren{\frac{16\log p}{w}, \sqrt{\frac{32\log p}{w}}} } \leq \frac{2}{p^2}.
\end{equation*}
Note that for any $w\geq \log p$, we have $\max\paren{\frac{16\log p}{w}, \sqrt{\frac{32\log p}{w}}} \leq 2$. Thus
\begin{equation*}
\bb{P}\brac{\abs{\psi_w\Bigparen{ a\paren{\omega_1}, b\paren{\omega_2} }} \geq 6M_{\varepsilon}\LpNorm{b\paren{\omega_2}-a\paren{\omega_1}}{2} } \leq \frac{2}{p^2}.
\end{equation*}
In summary, we obtained a conditional concentration inequality for $\psi_w\paren{a,b}$. We conclude the proof by directly applying Proposition \ref{CondProbProp}.
\end{proof}}

\begin{lem}\label{LemmaB3}
Let $\Omega,\Omega'\in S^{p\times p}_{++}$ be two inverse covariance matrices such that
\begin{equation}\label{NormBound}
\lambda_{\min}\paren{\Omega} \wedge \lambda_{\min}\paren{\Omega'} \geq \alpha_{\min},
\end{equation}
for a strictly positive scalar $\alpha_{\min}$. We use $R$ and $R'$ to denote the partial correlation matrices associated to $\Omega$ and $\Omega'$, respectively. If $\LpNorm{\Omega'-\Omega}{\infty} \leq \alpha_{\min}$, then there exists a bounded universal constant $C$ such that
\begin{equation*}
\frac{\LpNorm{R'-R}{4}}{\LpNorm{R}{4}}\leq \frac{C}{\alpha^2_{\min}}\brac{ \frac{\sqrt{\LpNorm{\Omega'-\Omega}{\infty} \LpNorm{\Omega'-\Omega}{2}} }{\LpNorm{R}{4}} +  \LpNorm{\Omega'-\Omega}{\infty} }
\end{equation*}
\end{lem}

\begin{proof}
For simplicity define $\Delta\coloneqq R'-R$. Our objective is control $\abs{\Delta_{st}},\; s,t=1,\ldots,p$ from above. Using triangle inequality yields
\begin{eqnarray*}
\abs{\Delta_{st}} &=& \abs{\frac{\Omega'_{st}}{\sqrt{\Omega'_{ss}\Omega'_{tt}}} - \frac{\Omega_{st}}{\sqrt{\Omega_{ss}\Omega_{tt}}}}\leq \frac{\abs{\Omega'_{st}-\Omega_{st}}}{\sqrt{\Omega'_{ss}\Omega'_{tt}}} + \abs{\Omega_{st}} \abs{\frac{1}{\sqrt{\Omega'_{ss}\Omega'_{tt}}} - \frac{1}{\sqrt{\Omega_{ss}\Omega_{tt}}}}\\
&\leq& \frac{\abs{\Omega'_{st}-\Omega_{st}}}{\sqrt{\Omega_{ss}\Omega_{tt}}} \sqrt{\frac{\Omega_{tt} }{\Omega'_{tt}}}\sqrt{\frac{\Omega_{ss} }{\Omega'_{ss}}}+  \abs{R_{st}}\abs{\sqrt{\frac{\Omega_{tt} }{\Omega'_{tt}}}\sqrt{\frac{\Omega_{ss} }{\Omega'_{ss}}}-1}
\end{eqnarray*}
Let us simplify the right hand side term in preceding inequality. The right hand side condition in Eq. \eqref{NormBound} guarantees that all diagonal entries of $\Omega$ are greater than $\alpha_{\min}$. Thus for all $s\in\set{1,\ldots,p}$,
\begin{equation}\label{AuxIneq}
\sqrt{\frac{\Omega_{ss} }{\Omega'_{ss}}} = \paren{1+ \frac{\Omega_{ss}-\Omega'_{ss}}{\Omega'_{ss}}}^{1/2} \leq 1+\frac{\abs{\Omega'_{ss}-\Omega_{ss}}}{2\Omega'_{ss}}\leq 1+\frac{\LpNorm{\Omega'-\Omega}{\infty}}{2\Omega'_{ss}} \leq 1+\frac{\LpNorm{\Omega'-\Omega}{\infty}}{2\alpha_{\min}}.
\end{equation} 
Moreover we know that $\LpNorm{\Omega'-\Omega}{\infty} \leq \alpha_{\min}$. So, the upper bound on $\abs{\Delta_{st}}$ can be rewritten as
\begin{eqnarray*}
\abs{\Delta_{st}} &\leq& \frac{\abs{\Omega'_{st}-\Omega_{st}}}{\alpha_{\min}} \paren{1+\frac{\LpNorm{\Omega'-\Omega}{\infty}}{2\alpha_{\min}}}^2+  \abs{R_{st}}\abs{\paren{1+\frac{\LpNorm{\Omega'-\Omega}{\infty}}{2\alpha_{\min}}}^2-1} \\
&\leq& \frac{9\abs{\Omega'_{st}-\Omega_{st}}}{4\alpha_{\min}} +  \frac{5\abs{R_{st}}}{4\alpha_{\min}}\LpNorm{\Omega'-\Omega}{\infty} \leq \frac{9\sqrt{\abs{\Omega'_{st}-\Omega_{st}}}}{4\alpha_{\min}} \sqrt{\LpNorm{\Omega'-\Omega}{\infty}}+  \frac{5\abs{R_{st}}}{4\alpha_{\min}}\LpNorm{\Omega'-\Omega}{\infty}.
\end{eqnarray*}
We now apply the following result to find an upper bound on $\abs{\Delta_{st}}^4$. 
\begin{equation*}
\paren{x+y}^4\leq 8\paren{x^4 + y^4},\quad \forall\;x,y\geq 0.
\end{equation*}
The proof of preceding result is left to the reader. So,
\begin{equation*}
\abs{\Delta_{st}}^4 \leq \frac{206}{\alpha^4_{\min}}\abs{\Omega'_{st}-\Omega_{st}}^2 \LpNorm{\Omega'-\Omega}{\infty}^2 \vee \frac{20}{\alpha^4_{\min}}\LpNorm{\Omega'-\Omega}{\infty}^4 R^4_{st}.
\end{equation*}
Now summing up over all entries of $\Delta$ (omitting the straightforward algebra), we get
\begin{equation}\label{AuxilIneq1}
\LpNorm{\Delta}{4}^4 \leq \frac{C}{\alpha^4_{\min}}\brac{ \LpNorm{\Omega'-\Omega}{\infty}^2 \LpNorm{\Omega'-\Omega}{2}^2 + \LpNorm{\Omega'-\Omega}{\infty}^4 \LpNorm{R}{4}^4 },
\end{equation}
for some bounded universal constant $C$. Taking forth root from both sides of Eq. \eqref{AuxilIneq1} ends the proof.
\end{proof}

The following result is an easy consequence of Lemma \ref{Lem5App} and Lemma \ref{LemmaB3}.

\begin{lem}\label{LemmaB4}
The following inequality holds under the same conditions of Lemma \ref{LemmaB3}.
\begin{equation*}
\abs{\frac{\sum_{s_1,s_2=1}^{p} h_w\paren{R'_{s_1,s_2}}}{\sum_{s_1,s_2=1}^{p} h_w\paren{R_{s_1,s_2}} }-1}\leq \frac{C}{\alpha^2_{\min}}\brac{ \frac{\sqrt{\LpNorm{\Omega'-\Omega}{\infty} \LpNorm{\Omega'-\Omega}{2}} }{\LpNorm{R}{4}} +  \LpNorm{\Omega'-\Omega}{\infty} }.
\end{equation*}
\end{lem}

\begin{proof}
Lemma \ref{Lem5App} ensures the existence of two bounded strictly positive constants $C_w, C'_w$ such that
\begin{equation*}
h_w\paren{r}\geq C'_wr^4,\;\; \mbox{and}\;\; h_w\paren{r}\leq C_wr^4,\quad \forall\; r\in\brac{-1,1}.
\end{equation*}
Therefore,
\begin{eqnarray}\label{Eq1}
\paren{\sum_{s_1,s_2=1}^{p} h_w\paren{R'_{s_1,s_2}}}^{\frac{1}{4}} &\leq& C^{\frac{1}{4}}_w\LpNorm{R'}{4}\leq C^{\frac{1}{4}}_w \LpNorm{R}{4} \paren{1+\frac{\LpNorm{R'-R}{4}}{\LpNorm{R}{4}}}\nonumber\\
&\leq& \paren{\frac{C_w}{C'_w}}^{\frac{1}{4}} \paren{\sum_{s_1,s_2=1}^{p} h_w\paren{R_{s_1,s_2}}}^{\frac{1}{4}}\paren{1-\frac{\LpNorm{R'-R}{4}}{\LpNorm{R}{4}}}.
\end{eqnarray}
Applying triangle inequality on Eq. \eqref{Eq1} leads to
\begin{equation*}
\frac{\sum_{s_1,s_2=1}^{p} h_w\paren{R'_{s_1,s_2}}}{\sum_{s_1,s_2=1}^{p} h_w\paren{R_{s_1,s_2}} }-1\lesssim \paren{1-\frac{\LpNorm{R'-R}{4}}{\LpNorm{R}{4}}}^4 - 1\leq \abs{1-\frac{\LpNorm{R'-R}{4}}{\LpNorm{R}{4}}} - 1 \leq \frac{\LpNorm{R'-R}{4}}{\LpNorm{R}{4}}.
\end{equation*}
Employing similar techniques, one can show that $1- \frac{\sum_{s_1,s_2=1}^{p} h_w\paren{R'_{s_1,s_2}}}{\sum_{s_1,s_2=1}^{p} h_w\paren{R_{s_1,s_2}} } \leq \frac{\LpNorm{R'-R}{4}}{\LpNorm{R}{4}}$. Thus,
\begin{equation}\label{Eq2}
\abs{\frac{\sum_{s_1,s_2=1}^{p} h_w\paren{R'_{s_1,s_2}}}{\sum_{s_1,s_2=1}^{p} h_w\paren{R_{s_1,s_2}} }-1}\lesssim \frac{\LpNorm{R'-R}{4}}{\LpNorm{R}{4}}.
\end{equation}
We conclude the proof by applying Lemma \ref{LemmaB3} to Eq. \eqref{Eq2}.
\end{proof}

\begin{lem}\label{SensAnalL}
Let $\Omega,\Omega'\in S^{p\times p}_{++}$ be two precision matrices such that
\begin{equation*}
\lambda_{\min}\paren{\Omega} \wedge \lambda_{\min}\paren{\Omega'} \geq \alpha_{\min},
\end{equation*}
for a strictly positive scalar $\alpha_{\min}$. Define $L_1,L_2\in\bb{R}^{p\times p}$ by
\begin{equation*}
L \coloneqq \brac{\frac{\Omega_{st}}{\sqrt{\Omega_{tt}}}}^p_{s,t=1},\quad\mbox{and}\quad L' \coloneqq \brac{\frac{\Omega'_{st}}{\sqrt{\Omega'_{tt}}}}^p_{s,t=1}.
\end{equation*}
If $\LpNorm{\Omega'-\Omega}{\infty} \leq \alpha_{\min}$, then there exists a bounded universal constant $C$ such that
\begin{equation*}
\LpNorm{L'-L}{2}\leq C\paren{ \frac{\LpNorm{\Omega'-\Omega}{2}}{\sqrt{\alpha_{\min}}} + \frac{\LpNorm{\Omega}{2}\LpNorm{\Omega'-\Omega}{\infty}}{\sqrt{\alpha^3_{\min}}} }.
\end{equation*}
\end{lem}

\begin{proof}
We apply similar techniques as the proof of Lemma \ref{LemmaB3}. For simplicity define $\Delta \coloneqq L'-L$. Triangle inequality leads to
\begin{equation*}
\abs{\Delta_{st}}\leq \frac{\abs{\Omega'_{st}-\Omega_{st}}}{\sqrt{\Omega'_{tt}}} + \frac{\abs{\Omega_{st}}}{\sqrt{\Omega_{tt}}} \abs{\sqrt{\frac{\Omega_{tt} }{\Omega'_{tt}}} -1},\quad\forall\;s,t\in\set{1,\ldots,p}.
\end{equation*}
Using Eq. \eqref{AuxIneq} and the fact that $\LpNorm{\Omega'-\Omega}{\infty} \leq \alpha_{\min}$ are critical for simplifying the obtained upper bound on $\abs{\Delta_{st}}$. We skip the algebraic details as it is exactly the same in Eq. \eqref{AuxIneq}.
\begin{equation}\label{AuxIneq2}
\abs{\Delta_{st}}^2 \leq \brac{\frac{\abs{\Omega'_{st}-\Omega_{st}}}{\sqrt{\alpha_{\min}}} + \frac{\abs{\Omega_{st}}\LpNorm{\Omega'-\Omega}{\infty}}{2\sqrt{\alpha^3_{\min}}}}^2 \leq 2\paren{\frac{\paren{\Omega'_{st}-\Omega_{st}}^2}{\alpha_{\min}} + \frac{\Omega^2_{st}\LpNorm{\Omega'-\Omega}{\infty}^2}{4\alpha^3_{\min}}}, ,\quad\forall\;s,t\in\set{1,\ldots,p}.
\end{equation}
Finally by combining inequality \eqref{AuxIneq2} for all pairs $\paren{s,t}$ in $\set{1,\ldots,p} \times \set{1,\ldots,p}$, we get
\begin{equation*}
\LpNorm{\Delta}{2}^2 = \sum_{s,t=1}^{p} \abs{\Delta_{st}}^2 \leq \frac{2}{\alpha_{\min}}\LpNorm{\Omega'-\Omega}{2}^2 + \frac{1}{2\alpha^3_{\min}}\LpNorm{\Omega}{2}^2 \LpNorm{\Omega'-\Omega}{\infty}^2.
\end{equation*}
This inequality can be easily reformulated as the desirable upper bound on $\LpNorm{\Delta}{2}$.
\end{proof}
\bibliographystyle{alpha}
\bibliography{./Manuscript-bib}

\newcommand{\etalchar}[1]{$^{#1}$}
\begin{thebibliography}{WZWD14}

\bibitem[AB17]{atchade2017scalable}
Yves Atchade and Leland Bybee.
\newblock A scalable algorithm for gaussian graphical models with
  change-points.
\newblock {\em arXiv preprint arXiv:1707.04306}, 2017.

\bibitem[AHH{\etalchar{+}}09]{aue2009break}
Alexander Aue, Siegfried H{\"o}rmann, Lajos Horv{\'a}th, Matthew Reimherr,
  et~al.
\newblock Break detection in the covariance structure of multivariate time
  series models.
\newblock {\em The Annals of Statistics}, 37(6B):4046--4087, 2009.

\bibitem[BLM13]{boucheron2013concentration}
St{\'e}phane Boucheron, G{\'a}bor Lugosi, and Pascal Massart.
\newblock {\em Concentration inequalities: A nonasymptotic theory of
  independence}.
\newblock Oxford university press, 2013.

\bibitem[BN{\etalchar{+}}93]{basseville1993detection}
Mich{\`e}le Basseville, Igor~V Nikiforov, et~al.
\newblock {\em Detection of abrupt changes: theory and application}, volume
  104.
\newblock Prentice Hall Englewood Cliffs, 1993.

\bibitem[BO16]{barnett2016change}
Ian Barnett and Jukka-Pekka Onnela.
\newblock Change point detection in correlation networks.
\newblock {\em Scientific reports}, 6:18893, 2016.

\bibitem[Bol82]{bolthausen1982central}
Erwin Bolthausen.
\newblock On the central limit theorem for stationary mixing random fields.
\newblock {\em The Annals of Probability}, pages 1047--1050, 1982.

\bibitem[BP98]{bai1998estimating}
Jushan Bai and Pierre Perron.
\newblock Estimating and testing linear models with multiple structural
  changes.
\newblock {\em Econometrica}, pages 47--78, 1998.

\bibitem[BP03]{bai2003computation}
Jushan Bai and Pierre Perron.
\newblock Computation and analysis of multiple structural change models.
\newblock {\em Journal of applied econometrics}, 18(1):1--22, 2003.

\bibitem[BVDG11]{buhlmann2011statistics}
Peter B{\"u}hlmann and Sara Van De~Geer.
\newblock {\em Statistics for high-dimensional data: methods, theory and
  applications}.
\newblock Springer Science \& Business Media, 2011.

\bibitem[BY03]{bessler2003structure}
David~A Bessler and Jian Yang.
\newblock The structure of interdependence in international stock markets.
\newblock {\em Journal of international money and finance}, 22(2):261--287,
  2003.

\bibitem[CLL11]{cai2011constrained}
Tony Cai, Weidong Liu, and Xi~Luo.
\newblock A constrained $\ell_1$ minimization approach to sparse precision
  matrix estimation.
\newblock {\em Journal of the American Statistical Association},
  106(494):594--607, 2011.

\bibitem[CLX13]{cai2013two}
Tony Cai, Weidong Liu, and Yin Xia.
\newblock Two-sample covariance matrix testing and support recovery in
  high-dimensional and sparse settings.
\newblock {\em Journal of the American Statistical Association},
  108(501):265--277, 2013.

\bibitem[CZ{\etalchar{+}}15]{chen2015graph}
Hao Chen, Nancy Zhang, et~al.
\newblock Graph-based change-point detection.
\newblock {\em The Annals of Statistics}, 43(1):139--176, 2015.

\bibitem[GR17]{gibberd2017multiple}
Alex~J Gibberd and Sandipan Roy.
\newblock Multiple changepoint estimation in high-dimensional gaussian
  graphical models.
\newblock {\em arXiv preprint arXiv:1712.05786}, 2017.

\bibitem[Guy95]{guyon1995random}
Xavier Guyon.
\newblock {\em Random fields on a network: modeling, statistics, and
  applications}.
\newblock Springer Science \& Business Media, 1995.

\bibitem[HAM{\etalchar{+}}16]{hindriks2016can}
Rikkert Hindriks, Mohit~H Adhikari, Yusuke Murayama, Marco Ganzetti, Dante
  Mantini, Nikos~K Logothetis, and Gustavo Deco.
\newblock Can sliding-window correlations reveal dynamic functional
  connectivity in resting-state fmri?
\newblock {\em Neuroimage}, 127:242--256, 2016.

\bibitem[HR14]{horvath2014extensions}
Lajos Horv{\'a}th and Gregory Rice.
\newblock Extensions of some classical methods in change point analysis.
\newblock {\em Test}, 23(2):219--255, 2014.

\bibitem[HSDR14]{hsieh2014quic}
Cho-Jui Hsieh, M{\'a}ty{\'a}s~A Sustik, Inderjit~S Dhillon, and Pradeep
  Ravikumar.
\newblock Quic: quadratic approximation for sparse inverse covariance
  estimation.
\newblock {\em Journal of Machine Learning Research}, 15(1):2911--2947, 2014.

\bibitem[HWG{\etalchar{+}}13]{hutchison2013resting}
R~Matthew Hutchison, Thilo Womelsdorf, Joseph~S Gati, Stefan Everling, and
  Ravi~S Menon.
\newblock Resting-state networks show dynamic functional connectivity in awake
  humans and anesthetized macaques.
\newblock {\em Human brain mapping}, 34(9):2154--2177, 2013.

\bibitem[JW05]{jones2005covariance}
Beatrix Jones and Mike West.
\newblock Covariance decomposition in undirected gaussian graphical models.
\newblock {\em Biometrika}, 92(4):779--786, 2005.

\bibitem[KBSM16]{kallitsis2016adaptive}
Michael~G Kallitsis, Shrijita Bhattacharya, Stilian Stoev, and George
  Michailidis.
\newblock Adaptive statistical detection of false data injection attacks in
  smart grids.
\newblock In {\em Signal and Information Processing (GlobalSIP), 2016 IEEE
  Global Conference on}, pages 826--830. IEEE, 2016.

\bibitem[KSAX10]{kolar2010estimating}
Mladen Kolar, Le~Song, Amr Ahmed, and Eric~P Xing.
\newblock Estimating time-varying networks.
\newblock {\em The Annals of Applied Statistics}, pages 94--123, 2010.

\bibitem[KSN17]{keshavarz2017optimal}
Hossein Keshavarz, Clayton Scott, and XuanLong Nguyen.
\newblock Optimal change point detection in gaussian processes.
\newblock {\em Journal of Statistical Planning and Inference}, 2017.

\bibitem[KX12]{kolar2012estimating}
Mladen Kolar and Eric~P Xing.
\newblock Estimating networks with jumps.
\newblock {\em Electronic journal of statistics}, 6:2069, 2012.

\bibitem[LC{\etalchar{+}}12]{li2012two}
Jun Li, Song~Xi Chen, et~al.
\newblock Two sample tests for high-dimensional covariance matrices.
\newblock {\em The Annals of Statistics}, 40(2):908--940, 2012.

\bibitem[Pav15]{pav2015moments}
Steven~E Pav.
\newblock Moments of the log non-central chi-square distribution.
\newblock {\em arXiv preprint arXiv:1503.06266}, 2015.

\bibitem[RAM16]{roy2016change}
Sandipan Roy, Yves Atchad{\'e}, and George Michailidis.
\newblock Change point estimation in high dimensional markov random-field
  models.
\newblock {\em Journal of the Royal Statistical Society: Series B (Statistical
  Methodology)}, 2016.

\bibitem[Ros14]{ross2014introduction}
Sheldon~M Ross.
\newblock {\em Introduction to probability models}.
\newblock Academic press, 2014.

\bibitem[WJ{\etalchar{+}}08]{wainwright2008graphical}
Martin~J Wainwright, Michael~I Jordan, et~al.
\newblock Graphical models, exponential families, and variational inference.
\newblock {\em Foundations and Trends{\textregistered} in Machine Learning},
  1(1--2):1--305, 2008.

\bibitem[WZWD14]{wu2014data}
Xindong Wu, Xingquan Zhu, Gong-Qing Wu, and Wei Ding.
\newblock Data mining with big data.
\newblock {\em IEEE transactions on knowledge and data engineering},
  26(1):97--107, 2014.

\bibitem[Xia10]{xiao2010dual}
Lin Xiao.
\newblock Dual averaging methods for regularized stochastic learning and online
  optimization.
\newblock {\em Journal of Machine Learning Research}, 11(Oct):2543--2596, 2010.

\end{thebibliography}

\end{document}